\documentclass[fleqn]{SCYE}
\usepackage{relsize}
\usepackage{bm}
\usepackage{amsmath}
\usepackage{amsthm}
\usepackage{algorithm,algorithmic}
\usepackage{amsfonts}
\usepackage{import}
\usepackage{multirow}
\usepackage{subfigure}
\usepackage{graphicx}
\usepackage{cite}

\theoremstyle{plain}

\begin{document}

%%%%%%%%%%%%%%%%%%%%%%%%%%%%%%%%%%%%%%%%%%%%%%%%%%%%%%%
%%% Authors do not modify the information below
%%% ????????????????
%%% ??????????, ????????????{}, ???????????????????
%Letter to the Editor??Article%??????
\ArticleType{Letter to the Editor}%??Article
%\SpecialTopic{SPECIAL TOPIC: }%???????
\Year{2019}
\Month{January}
\Vol{62}
\No{1}
\DOI{}
\BeginPage{1} % ÆðÒ³Âë
\EndPage{}
% \ReceiveDate{January 11, 2017}
% \AcceptDate{April 6, 2017}
% \OnlineDate{January 1, 2017}
%%%%%%%%%%%%%%%%%%%%%%%%%%%%%%%%%%%%%%%%%%%%%%%%%%%%%%%

%%% title: ????
%%%   \title{title}{title for citation}
\title{Latent Matrices for Tensor Network Decomposition and to Tensor Completion}
{Latent Matrices for Tensor Network Decomposition and to Tensor Completion}

%%% Corresponding author: ???????
%%%   \author[number]{Full name}{{email@xxx.com}}
%%% General author: ???????
%%%   \author[number]{Full name}{}
\author[1,2]{Peilin Yang}{2112104344@mail2.gdut.edu.cn}%
\author[1,2]{Weijun Sun}{gdutswj@gdut.edu.cn}
\author[1,3]{Qibin Zhao}{}%\protect\\?§Þ??§Ø????
\author[1,4]{Guoxu Zhou}{}%
%\author[1]{Nanjian WU}{}

%%% Author information for page head. ?¨¹?§Ö????????
%%% ??????????????, ??????????author???
\AuthorMark{Yang P L}%\authorcr????????

%%% Authors for citation. ????????§Ö????????
%%% ??????????????, ??????????author???
\AuthorCitation{Yang P L, Sun W J, Zhao Q B, Zhou G X}

%%% Address.
%%%   \address[number]{Address, City {\rm Postcode}, Country}
\address[1]{School of Automation, Guangdong University of
Technology, Guangzhou {\rm 510006}, China}
\address[2]{Guangdong Key Laboratory of IoT Information Technology, Guangzhou {\rm 510006}, China}
\address[3]{Center for Advanced Intelligence Project (AIP), RIKEN, Tokyo {\rm 103-0027}, Japan}
\address[2]{Key Laboratory of Intelligent Detection and The Internet of Things in Manufacturing, Ministry of Education, Guangzhou {\rm 510006}, China}

\contributions{These authors contributed equally to the work.}%

%%% Abstract.
\abstract{The prevalent fully-connected tensor network (FCTN) has achieved excellent success to compress data. 
  However, the FCTN decomposition suffers from slow computational speed when facing higher-order and large-scale data. 
  Naturally, there arises an interesting question: can a new model be proposed that decomposes the tensor into smaller ones and speeds up the computation of the algorithm? 
  This work gives a positive answer by formulating a novel higher-order tensor decomposition model that utilizes latent matrices based on the tensor network structure, 
  which can decompose a tensor into smaller-scale data than the FCTN decomposition, hence we named it Latent Matrices for Tensor Network Decomposition (LMTN). 
  Furthermore, three optimization algorithms, LMTN-PAM, LMTN-SVD and LMTN-AR, have been developed and applied to the tensor-completion task. 
  In addition, we provide proofs of theoretical convergence and complexity analysis for these algorithms.
  Experimental results show that our algorithm has the effectiveness in both deep learning dataset compression 
  and higher-order tensor completion, and that our LMTN-SVD algorithm is 3-6 times faster than the FCTN-PAM algorithm and only a 1.8 points accuracy drop. }                  

%%% Keywords.

\maketitle

%\tableofcontents%?????

%%%%%%%%%%%%%%%%%%%%%%%%%%%%%%%%%%%%%%%%%%%%%%%%%%%%%%%
%%% The main text. ???????
%\cite{3,4,5,6}  Á´ÓÃ [1£­3]
% ¹«Ê½ÒýÓà        ×Ô¶¯¼ÓÀ¨ºÅ
%\eqref{eq1}   (1)
%\cref{eq1}     eq.(1)
%\Cref{eq1}    Eq.(1)
%\cref{fig1}   Figure 1
%\cref{tab1}   Table  1
%\href{Á´½ÓÍøÖ·}{ÏÔʾÍøÖ·}%%\href{https://mc03. manuscriptcentral.com/scpma}{https://mc03.manuscriptcentral.com/scpma}

%\twocolumn\onecolumn
%%%%%%%%%%%%%%%%%%%%%%%%%%%%%%%%%%%%%%%%%%%%%%%%%%%%%%%
\begin{multicols}{2}

\section{Introduction}\label{section1}
Tensor, refer to as multi-way data, can be seen as the higher-order generalization of matrices and
has attracted considerable attentions in a variety fields including machine learning, 
signal processing, quantum physics, chemo metrics and brain science\cite{1,2,18}. 
However, storing a higher-order tensor is a huge drain on computer memory usually. 
To tackle this issue, researchers have proposed an essential technique for higher-order data dimensionality reduction - tensor decomposition. 
Tensor decompositions aim to represent a higher-order tensor data by multilinear operations over lant factors. 
The success of the tensor decomposition algebra is \Authorfootnote closely related to the effective way of dealing with the curse of tensor dimensionality. 
By designing the structure of different latent factors and different linear operations, many different tensor decomposition 
algorithms have emerged. Specifically, CANDECOMP/PARAFAC (CP) decomposition and Tucker decomposition 
are two of the most classical decomposition algorithms and have achieved considerable success in tensor decompositions\cite{3,4,5,6,7,8}. 
% \Authorfootnote%Ê×Ò³EmailµØÖ·

Recently, a growing number of tensor decomposition algorithms have been proposed, and they show superior 
performance on higher-order tensor decompositions\cite{31,32,33,34,35,36,37}. Two of the most popular are the tensor train (TT) and tensor ring (TR) decompositions, 
which require $\mathcal{O}((N-2)IR^2+2IR)$ and $\mathcal{O}(NIR^2)$ parameters respectively to represent a higher dimensional tensor\cite{9,10,38,39,40}. 
Furthermore, they have been employed in many applications, such as 
signal recovery, compression, image (video) recovery and noise removal\cite{11,12,13}. Then, the 
recently proposed FCTN decomposition captures the intrinsic correlation between any 
two modes of tensors adequately and improves the data compression performance\cite{14}. Subsequently, a 
number of algorithms have been proposed based on the FCTN decomposition. For example, Yang et al. developed a weighting model for the FCTN decomposition, which was optimised using the gradient descent method\cite{30}. 
Zheng et al. proposed a novel nonlocal patch-based FCTN (NL-FCTN) decomposition for remote sensing image (RSI) inpainting, which 
increases tensor order by stacking similar small-sized patches to nonlocal self-similarity (NSS) groups, cleverly leverages the 
remarkable ability of FCTN decomposition to deal with higher-order tensors\cite{16}. Besides, Liu et.al proposed a 
FCTN-based robust convex optimization model (RC-FCTN) for the robust tensor completion (RTC) problem\cite{17}. 
And Nie et al. proposed a data-adaptive tensor decomposition model establishs on a generalized tensor 
rank, which constructs an optimal topological structure for tensor decomposition according to the intrinsic properties of the data\cite{15}. 

However, FCTN decomposition requires $\mathcal{O}(NIR^{n-1})$ parameters to approximate an observed higher-order tensor. This means that as the order of the 
tensor increases, the required parameters will grow exponentially and the calculation speed of the computer will be slow. As a 
sample, the required parameters is $\mathcal{O}(5IR^{4})$ when the observed tensor's order is 5, which is 
much larger than the $\mathcal{O}(5IR^2)$ parameters required from TR decomposition. To tackle this problem, 
Nie et al. proposed an adaptive tensor network (ATN) decomposition\cite{15}, which can construct the structure of factors to reduce 
the number of parameters. Furthermore, Sedighin et al. proposed an approach for selecting ranks adaptively to reduce 
the storage cost considerably\cite{19}. But none of them attempted to learn a new structure to explore the latent information of the FCTN decomposition factors based on dimensionality reduction.

Therefore, we are keen to design a decomposition model that retains the ability of the FCTN 
decomposition to characterise intrinsic correlations, while being able to mine the latent information of the factors and reduce the model parameters of the higher-order tensor. 
Inspired on top of the deep non-negative matrix decomposition model by Trigeorgis G et al.\cite{29}, 
we propose a latent matrices for tensor network decomposition model (LMTN). 
This model decomposes the observed tensor into several factors which are interconnected with each other and 
each factor is also interconnected with an orthogonal matrix. The proposed LMTN decomposition model can capture the latent information of the factors 
and  has a great dimensionality reduction when approximating the higher-order tensor.

In summary, contributions of this work are as follows:
\begin{itemize}
  \item{Based on the FCTN decomposition, we propose an LMTN decomposition that can compress the parameters of the FCTN decomposition very well.}
  \item{We propose different optimization algorithms for the LMTN decomposition and the experiment shows that the proposed algorithms are 3-6 times faster than FCTN-PAM algorithm and only 1.8 points accuracy drop.}
  \item{We experimentally demonstrate that the proposed algorithms can globally converge to a local minima and has good recovery results.}
\end{itemize}

The proposed approach is quite general and can serve a wide variety of purposes. For its validation, we consider 
case studies ranging from synthetic data, real data (image and video) and traffic data. We have used different metrics 
to analyse and compare different algorithms, ultimately demonstrating the effectiveness and stability of our algorithms. 

\section{Preliminariess}\label{section2}
In this paper, we use boldface calligraphic letters to denote tensors, e.g., $\mathbf{\mathcal{X}}$; boldface capital letters 
to denote matrices, e.g., $\mathbf{X}$. The boldface lowercase letters and lowercase letters respectively denote vectors and 
scalars, e.g., $\mathbf{x}$ and $x$. For an $N$th-order tensor $\mathcal{X} \in \mathbb{R}^{I_1\times I_2\times \cdots \times I_N}$, 
we employ $\mathcal{X}(i_1,i_2, \cdots ,i_N)$ to denote its $(i_1,i_2,\cdots, i_N)$th element. $\Vert \mathcal{X}\Vert_F$ denotes 
the Frobenius norm of $\mathcal{X}$, calculated by $(\sum_{i_1,i_2,\cdots,i_N}^{I_1,I2,\cdots,I_N}\mathcal{X}^2)^{1/2}$. In addition, 
the transpose, inverse and pseudo-inverse of matrix $\mathbf{X}$ are denote as $\mathbf{X}^T$, $\mathbf{X}^{-1}$ and $\mathbf{X}^{\dagger}$. 
For convenience, we shall introduce some definitions of tensor operations, involved in the paper are given as follows. 

% \begin{figure}[!t]
%   \centering
%   \includegraphics[width=3in]{FCTN2.eps}
%   \captionof{figure}{A graphical representation of FCTN decomposition}
%   \label{fig_1}
% \end{figure}

\noindent $\mathbf{Definition~1(Matricization)}$ For an N-order tensor $\mathcal{X} \in \mathbb{R}^{I_1\times I_2\times \cdots \times I_N}$, 
the mode-n matricization of $\mathcal{X}$ arranges the mode-n fibers into the columns of the unfold matrix, denote as 
$\mathbf{X}_{(n)} \in \mathbb{R}^{I_k \times \prod_{i\neq n}I_i}$ and its elements are
\begin{equation*}
  \mathbf{X}_{(n)}(i_n, \overline{i_1\cdots i_{n-1}i_{n+1}\cdots i_{N}}) = \mathcal{X}(i_1,i_2,\cdots,i_N).
\end{equation*}

\noindent $\mathbf{Definition~2(Mode-n ~Matrix ~Product)}$ The mode-n matrix product of an $N$th-order tensor $\mathcal{X} \in \mathbb{R}^{I_1 \times I_2 \times \cdots \times I_N}$ 
with a matrix $\mathbf{U} \in \mathbb{R}^{J_n \times I_n}$ is denoted by the symbol $\mathcal{X}\times_n\mathbf{U}$. This is 
a $I_1 \times I_2 \times \cdots \times I_{n-1} \times J_n \times I_{n+1} \times \cdots \times I_N$ tensor whose elements are defined as
\begin{equation*}
  (\mathcal{X}\times_n\mathbf{U})_{i_1\cdots i_{n-1}ji_{n+1}\cdots i_N} = \sum_{i_n=1}^{I_n} x_{i_1i_2\cdots i_N}a_{ji_n}
\end{equation*}
where $j=1,\cdots,J_n$; $i_k=1,\cdots,I_k$;$k=1,\cdots,N$. 

Thus, the mode-n matrix product of the tensor $\mathcal{X} \in \mathbb{R}^{I_1\times I_2\times \cdots \times I_N}$ and the 
matrix $\mathbf{U} \in \mathbb{R}^{J_n\times I_n}$ can be expressed in the form of tensor unfolding:
\begin{equation*}
  \mathcal{Y} = \mathcal{X} \times_n \mathbf{U} \Longleftrightarrow \mathbf{Y}_{(n)}=\mathbf{UX}_{(n)}
\end{equation*}
\begin{figure}[H]
  \centering
  \includegraphics[width=3in]{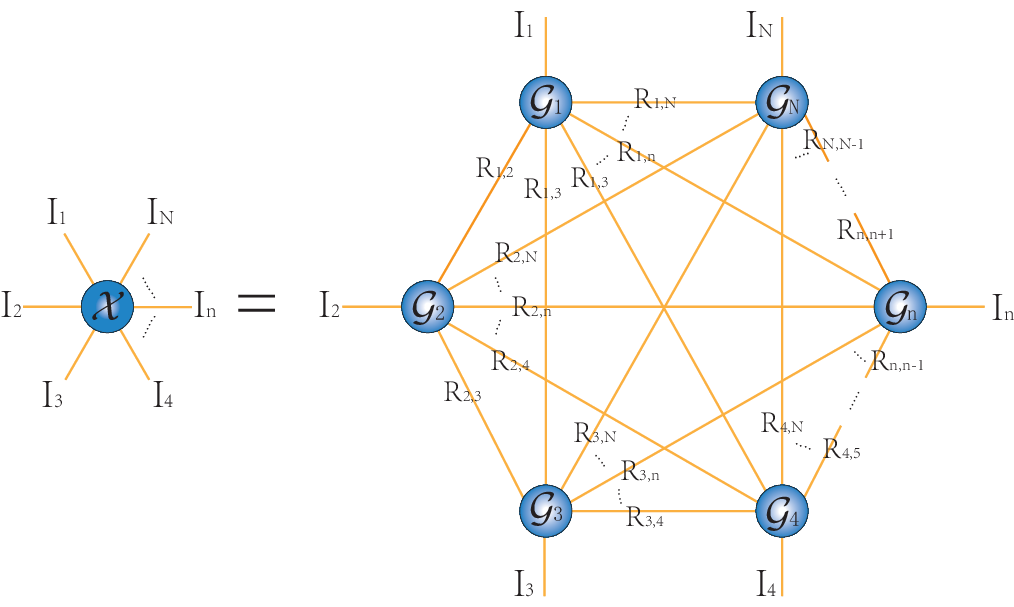}
  \caption{A graphical representation of FCTN decomposition}
  \label{fig_1}
\end{figure}
where $\mathbf{X}_{(n)}\in \mathbb{R}^{I_n\times I_1\cdots I_{n-1}I_{n+1}\cdots I_N}$ is the mode-n unfolding of $N$th-order 
tensor $\mathcal{X}$.

\noindent $\mathbf{Definition~3(Kronecker~Product)}$ The Kronecker product of a matrix $\mathbf{A} \in \mathbb{R}^{m\times n}$ 
and a matrix $\mathbf{B} \in \mathbb{R}^{p\times q}$ is a matrix of size $mp\times nq$, difined as
\begin{equation*}
  \begin{aligned}
  [\mathbf{A} \otimes \mathbf{B}] &= [\mathbf{a}_1\mathbf{B},\cdots,\mathbf{a}_n\mathbf{B}] \\ &= [a_{ij}\mathbf{B}]_{i=1,j=1}^{m,n} \\ &= 
  \begin{bmatrix}
    a_{11}\mathbf{B} & a_{12}\mathbf{B} & \cdots & a_{1n}\mathbf{B}\\ 
    a_{21}\mathbf{B} & a_{22}\mathbf{B} & \cdots & a_{2n}\mathbf{B}\\
    \vdots           & \vdots           & \ddots & \vdots\\
    a_{m1}\mathbf{B} & a_{m2}\mathbf{B} & \cdots & a_{mn}\mathbf{B}\\
  \end{bmatrix}
\end{aligned}
\end{equation*}

\noindent $\mathbf{Definition~4(Generalized ~Transposition)}$\cite{14} Suppose there is a set of arrangements $\mathbf{n}=[1,2,\cdots ,N]$, 
rearrange the sequence of $\mathbf{n}$ to obtain a new sequence $\overline{\mathbf{n}}=[n_1,n_2,\cdots ,n_N]$, $n_i \in \mathbf{n}$. 
After reordering the $N$th order tensor $\mathcal{X} \in \mathbb{R}^{I_1 \times I_2 \times \cdots \times I_N}$, we can obtain 
the new $N$th-order tensor $\overline{\mathcal{X}} \in \mathbb{R}^{I_{n_1} \times I_{n_2} \times \cdots \times I_{n_N}}$.

\noindent $\mathbf{Definition~5(Generalized ~Tensor ~Unfolding)}$\cite{14} Suppose we have a $N$th-order tensor $\mathcal{X} \in \mathbb{R}^{I_1 \times I_2 \times \cdots \times I_N}$ 
and a sequence $\overline{\mathbf{n}}$, then the generalized tensor unfolding of $\mathcal{X}$ is defined as a matrix
\begin{equation*}
  \mathbf{X}_{[n_{1:j};n_{j+1:N}]} = reshape(\overline{\mathcal{X}},\prod_{i=1}^j I_{n_i},\prod_{i=j+1}^N I_{n_i}).
\end{equation*}
where the reshape operation is a function provided by MATLAB. 

\noindent $\mathbf{Definition~6(Tensor~Contraction)}$\cite{14} Suppose we have tensor $\mathcal{A} \in \mathbb{R}^{I_1 \times I_2 \times \cdots \times I_N}$ 
and tensor $\mathcal{B} \in \mathbb{R}^{J_1 \times J_2 \times \cdots \times J_M}$. If they are both rearranged according to Definition 4, we will 
obtain tensor $\overline{\mathcal{A}} \in \mathbb{R}^{I_{n_1} \times I_{n_2} \times \cdots \times I_{n_N}}$ and tensor $\overline{\mathcal{B}} \in \mathbb{R}^{J_{m_1} \times J_{m_2} \times \cdots \times I_{m_M}}$, which 
satisfied $I_{n_i}=J_{m_i}$ with $i = 1,2,3,\cdots,k$. Their contractionis defined as
\begin{equation*}
  \begin{aligned}
  \mathcal{C} = \mathcal{A} \times_{n_{1:k}}^{m_{1:k}} \mathcal{B}~\in~\mathbb{R}^{I_{n_{k+1}} \times \cdots \times I_{n_{N}} \times J_{m_{k+1}} \times \cdots \times J_{m_{M}}}~\Longleftrightarrow\\
  \mathbf{C}_{[1:N-k;N-k+1:N+M-2k]} = \mathbf{A}_{[n_{k+1}:n_{N};n_1:n_k]} \mathbf{B}_{[m_1:m_k; m_{k+1:M}]}
  \end{aligned}
\end{equation*}
Especially, it requires $n_{k+1} < n_{k+2} < \cdots < n_{N}$ and $m_{k+1} < m_{k+2} < \cdots < m_{M}$ to guarantee 
the uniquencess.

\noindent $\mathbf{Definition~7(FCTN ~Decomposition)}$\cite{14} The FCTN decomposition model can decompose an $N$th-order tensor $\mathcal{X}\in \mathbb{R}^{I_1\times I_2 \times \cdots \times I_N}$ 
to a sequence of factors $\mathcal{G}_n \in \mathbb{R}^{R_{1,n}\times R_{2,n}\times \cdots \times R_{n-1,n} \times I_n \times R_{n,n+1} \times R_{n,n+2} \times \cdots \times R_{n,N}}$, $n = 1,2,\cdots,N$. 
And we define it as 
$$
\begin{aligned}
&\mathcal{X}\left(i_{1}, i_{2}, \cdots, i_{N}\right)= \\
&\sum_{r_{1,2}=1}^{R_{1,2}} \sum_{r_{1,3}=1}^{R_{1,3}} \cdots \sum_{r_{1, N}=1}^{R_{1, N}} \sum_{r_{2,3}=1}^{R_{2,3}} \cdots \sum_{r_{2, N}=1}^{R_{2, N}} \cdots \sum_{r_{N-1, N}=1}^{R_{N-1, N}} \\
&\left\{\mathcal{G}_{1}\left(i_{1}, r_{1,2}, r_{1,3}, \cdots, r_{1, N}\right)\right. \\
\quad &\mathcal{G}_{2}\left(r_{1,2}, i_{2}, r_{2,3}, \cdots, r_{2, N}\right) \cdots \\
\quad &\mathcal{G}_{n}\left(r_{1, n}, r_{2, n}, \cdots, r_{n-1, n}, i_{n}, r_{n, n+1}, \cdots, r_{n, N}\right) \cdots \\
&\left.\mathcal{G}_{N}\left(r_{1, N}, r_{2, N}, \cdots, r_{N-1, N}, i_{N}\right)\right\}
\end{aligned}
$$
For the sane of record, we use the equation $\mathcal{X} = FCTN(\{\mathcal{G}_n\}_{n=1}^N) = FCTN(\mathcal{G}_1,\mathcal{G}_2,\cdots,\mathcal{G}_N)$ to 
represent the FCTN decomposition. To show the FCTN decomposition more clearly, we use Fig. 1 to illustrate it.

\noindent $\mathbf{Definition~8(FCTN ~Composition)}$\cite{14} We denote the process of generating $\mathcal{X}$ by $FCTN$ factors $\mathcal{G}_n(n = 1,2,\cdots,N)$ 
as $FCTN(\{ \mathcal{G}_n\}_{n=1}^N)$ and we call it FCTN composition. In addition, if all the factors participate in the composition except $\mathcal{G}_n(n\in \{1,2,\cdots,N\})$, 
we denote it as $FCTN(\{\mathcal{G}_n\}_{n=1}^N,/\mathcal{G}_n)$.

\section{Learning Algorithm}\label{section3}
In this section, we will show several algorithms to learn the LMTN model. Since tensor decomposition model has different requirements in both 
accuracy and speed, we focused on the performance from developing different optimizations. 

\subsection{LMTN Decomposition} 
The LMTN model represents a higher-order tensor multilinear operation over a set of latent factors, which are consisted by the $N$th-order factors and 
matrices. In order to illustrate our decomposition method more visually, we will show it in Fig. 2. For the sake of better visibility, we use $\mathbf{R}$ to denote the 
rank of the LMTN model. Assuming an $N$th-order tensor is decomposed by the LMTN model, we can obtain $N$th-order low rank tensors $\mathcal{G}_i$ and matrices $\mathbf{M}_i$. 
Hence, the rank of LMTN model can be naturally expressed as $\mathbf{R} \in \mathbb{R}^{N\times N}$, which is
\begin{equation}
  \mathbf{R} = 
  \begin{pmatrix}
   R_{1,1} & R_{1,2} & \cdots & R_{1,N}\\
   R_{1,2} & R_{2,2} & \cdots & R_{2,N}\\
   \vdots  & \vdots  & \ddots & \vdots \\
   R_{1,N} & R_{2,N} & \cdots & R_{N,N}
  \end{pmatrix} 
\end{equation} 
where $R_{i,j}(i \neq j)$ denotes the dimension of the common mode between the factor $\mathcal{G}_i$ and the factor $\mathcal{G}_j$, and $R_{i,i}$ denotes 
the dimension of the common mode between the factor $\mathcal{G}_i$ and the matrix $\mathbf{M}_i$. Furthermore, $\mathbf{R}$ is a skew-symmetric matrix. 
\begin{figure}[H]
  \centering
  \includegraphics[width=3in]{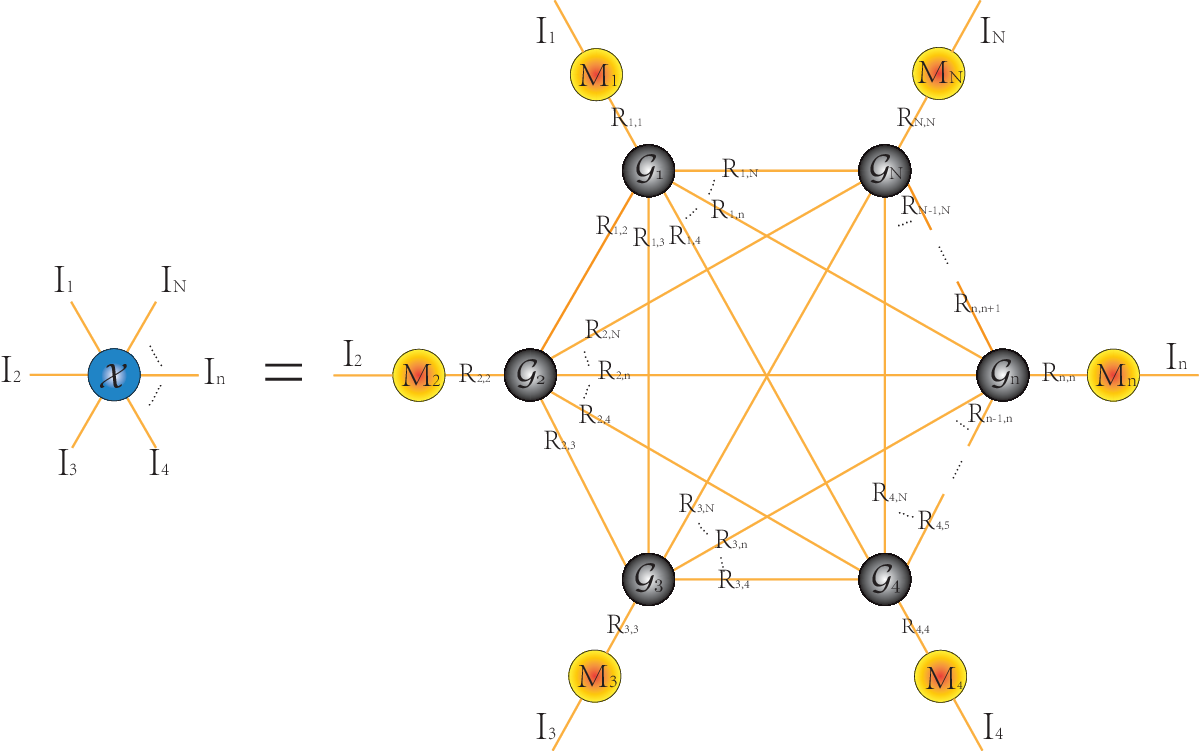}
  \caption{LMTN decomposition. The black balls represent the $N$th-order factor 
  $\mathcal{G}_{i} \in \mathbb{R}^{R_{1,i}\times R_{2,i} \times \cdots \times R_{i,i} \times R_{i,i+1} \times \cdots \times R{i,N}},~i\in [1,2,\cdots ,N]$ 
  and the yellow balls represent the matrix $\mathbf{M}_i \in \mathbb{R}^{I_i \times R_{i,i}}$ with $i \in [1,2,\cdots ,N]$}
  \label{fig_2}
\end{figure}

\noindent $\mathbf{Definition~9(LMTN~Decomposition)}$ The element-wise form of the LMTN decomposition can be expressed as 
\begin{equation}
  \begin{aligned}
  &\mathcal{X}(i_1,i_2,\dots,i_N)=LMTN(\{ \mathcal{G}_n\} _{n=1}^N ,\{ \mathbf{M}_n\}_{n=1}^N )\\
  &=\sum_{r_{1,1}=1}^{R_{1,1}} \sum_{r_{2,2}=1}^{R_{2,2}} \cdots \sum_{r_{N,N}=1}^{R_{N,N}}\\
  &[\sum_{r_{1,2}=1}^{R_{1,2}} \sum_{r_{1,3}=1}^{R_{1,3}} \cdots \sum_{r_{1, N}=1}^{R_{1, N}} \sum_{r_{2,3}=1}^{R_{2,3}} \cdots \sum_{r_{2, N}=1}^{R_{2, N}} \cdots \sum_{r_{N-1, N}=1}^{R_{N-1, N}} \\
  &\left\{\mathcal{G}_{1}\left(i_{1}, r_{1,2}, r_{1,3}, \cdots, r_{1, N}\right)\right. \\
  \quad &\mathcal{G}_{2}\left(r_{1,2}, i_{2}, r_{2,3}, \cdots, r_{2, N}\right) \cdots \\
  \quad &\mathcal{G}_{n}\left(r_{1, n}, r_{2, n}, \cdots, r_{n-1, n}, i_{n}, r_{n, n+1}, \cdots, r_{n, N}\right) \cdots \\
  &\left.\mathcal{G}_{N}\left(r_{1, N}, r_{2, N}, \cdots, r_{N-1, N}, i_{N}\right)\right\}] \\
  &\mathbf{M}_1(i_1,r_{1,1}) \mathbf{M}_2(i_2,r_{2,2}) \cdots \mathbf{M}_N(i_N,r_{N,N})
  \end{aligned}
\end{equation} 
where $\mathbf{M}_n$ is a semi-orthogonal matrix, i.e. $\mathbf{M}_n^T \mathbf{M}_n=\mathbf{E}$, $\mathbf{E}$ is the unit matrix, and $R_{n,n} \leq I_n$. 

\noindent $\mathbf{Theorem~1(Generalized~Relation)}$ There is a transformation relation for the $N$th-order tensor $\mathcal{X}$:
\begin{equation}
  \begin{aligned}
    &\mathcal{X}=FCTN(\{ \mathcal{G}_n\}_{n=1}^N) \times_1 \mathbf{M}_1 \times_2 \mathbf{M}_2 \cdots \times_N \mathbf{M}_N\\
    &\Rightarrow FCTN(\{ \mathcal{G}_n\}_{n=1}^N) = \mathcal{X} \times_1 \mathbf{M}_1^T \times_2 \mathbf{M}_2^T \cdots \times_N \mathbf{M}_N^T
  \end{aligned}
\end{equation}
where $FCTN(\{ \mathcal{G}_n\}_{n=1}^N) \in \mathbb{R}^{R_{1,1}\times \cdots R_{N,N}}$ and $\mathbf{M}_n \in \mathbb{R}^{I_n \times R_{n,n}}$, and $R_{n,n} \leq I_n$. And we provide its proof in the supplementary material. 

\noindent $\mathbf{Definition~10(LMTN~Composition)}$ For each $\mathcal{G}_i$ and $\mathbf{M}_i$, we define that $\mathcal{Y}_i = \mathcal{G}_i \times_i \mathbf{M}_i\in \mathbb{R}^{R_{1,i}\times R_{2,i} \cdots \times R_{i-1,i} \times I_i \times R_{i,i+1} \times R_{i,N}}$ 
with $i\in [N]$ and that we denote the process of generating $\mathcal{X}$ by $LMTN$ factors $\mathcal{G}_n(n = 1,2,\cdots,N)$ and $\mathbf{M}_n(n=1,2,\cdots,N)$ 
as $LMTN(\{ \mathcal{Y}_n\}_{n=1}^N)$ or $LMTN(\{ \mathcal{G}_n\}_{n=1}^N,\{ \mathbf{M}_n\}_{n=1}^N)$. Similarly, if all the factors participate in the composition except $\mathcal{Y}_k(k\in \{1,2,\cdots,N\})$, 
we denote it as $LMTN(\{\mathcal{Y}_n\}_{n=1}^N,/\mathcal{Y}_k)$.

\noindent $\mathbf{Definition~11(LMTN's~Matrix~Representation)}$ Supposing that $\mathcal{X} = LMTN(\{ \mathcal{G}_n\}_{n=1}^N,\{ \mathbf{M}_n\}_{n=1}^N)$ and $\mathcal{Y}_{\neq k}=LMTN(\{\mathcal{Y}_n\}_{n=1}^N,/\mathcal{Y}_k)$, 
we have that 
\begin{equation}
  \begin{aligned}
    \mathbf{X}_{(k)} &= (\mathbf{Y}_k)_{(k)} (\mathbf{Y}_{\neq k})_{[m_{1:N-1};n_{1:N-1}]}\\
    & = \mathbf{M}_k (\mathbf{G}_k)_{(k)} (\mathbf{Y}_{\neq k})_{[m_{1:N-1};n_{1:N-1}]}
  \end{aligned}
\end{equation}
where
$$
m_{i}=\left\{\begin{array}{ll}
2 i, & \text {if } i<k, \\
2 i-1, & \text {if } i \geq k,
\end{array} \text { and } n_{i}= \begin{cases}2 i-1, & \text {if } i<k \\
2 i, & \text {if } i \geq k\end{cases}\right.
$$

Definition 11 is very important for LMTN decomposition computations and we will use it frequently in the following optimization algorithms.

\subsection{LMTN-PAM algorithm}
In this subsection, we try to apply the proposed algorithm to the tensor completion task and build a model for solving it. 
The proposed of tensor completion is approximating the missing elements from partially observed elements by exploiting the high correlation between the elements. 
Given an incomplete observation $\mathcal{T}\in \mathbb{R}^{I_1\times I_2\times \cdots \times I_N}$
of the target tensor $\mathcal{X} \in \mathbb{R}^{I_1 \times I_2 \times \cdots \times I_N}$, our LMTN decomposition-based proximal alternating minimization 
(PAM) model can be
\begin{equation}
  \begin{aligned}
    \min_{\mathcal{X},\mathcal{G},\mathbf{M}} \frac{1}{2} \Vert \mathcal{X} - LMTN(\{ \mathcal{G}_n\}_{n=1}^N,\{ \mathbf{M}_n\}_{n=1}^N) \Vert_F^2 + l_{\mathbb{S}}(\mathcal{X})
  \end{aligned}
\end{equation}
where 
$$
\iota_{\mathbb{S}}(\mathcal{X}):=\left\{\begin{aligned}
0&, \text { if } \mathcal{X} \in \mathbb{S },\\
\infty&, \text { otherwise, }
\end{aligned}\text{with } \mathbb{S}:=\left\{\mathcal{X}: \mathcal{P}_{\Omega}(\mathcal{X}-\mathcal{T})=0\right\}\right.
$$

Here the $\Omega$ is the index set of observed entries and $\mathcal{P}_{\Omega}(\mathcal{T})$ denotes all the observed entries of tensor $\mathcal{T}$.
For the equation (5) we can unfold it into the form of a matrix as 
\begin{equation}
  \begin{aligned}
    \min_{\mathbf{X}_{(k)},(\mathbf{G}_{k})_{(k)},\mathbf{M}_{k}} \frac{1}{2} \Vert \mathbf{X}_{(k)} -  \mathbf{M}_{k} (\mathbf{G}_{k})_{(k)} \mathbf{Y}_{(\neq k)} \Vert_F^2 + l_{\mathbb{S}}(\mathcal{X})
  \end{aligned}
\end{equation}
where $\mathcal{Y}_{\neq k} = LMTN(\{\mathcal{Y}_n\}_{n=1}^N,/\mathcal{Y}_{k})$, and $(\mathbf{Y}_{\neq k})_{[m_{1:N-1};n_{1:N-1}]}$ is abbreviated to $\mathbf{Y}_{(\neq k)}$ for convenience. 
In addition, the sequence $\mathbf{m}$ and $\mathbf{n}$ are the same setting as Definition 11. 

We employ the framework of PAM\cite{24} to solve proplem (6), whose solution can be obtained by alternately updating 
\begin{equation}
  \left\{ 
  \begin{aligned} 
    \mathbf{M}_k^{s+1} = \mathop{\arg\min}_{\mathbf{M}_k}  &\{ f(\mathbf{M}_{1:k-1}^{(s+1)},\mathbf{M}_k,\mathbf{M}_{k+1:N}^{(s)},\mathcal{G}_{1:k-1}^{(s+1)},\mathcal{G}_{k:N}^{(s)} )\\ & + \frac{\rho}{2} \Vert \mathbf{M}_{k} - \mathbf{M}_{k}^{(s)} \Vert^2 \}\\
    \mathcal{G}_{k}^{(s+1)} = \mathop{\arg\min}_{\mathcal{G}_k} &\{ f(\mathbf{M}_{1:k}^{(s+1)},\mathbf{M}_{k+1:N}^{(s)},\mathcal{G}_{1:k-1}^{(s+1)} , \mathcal{G}_{k}, \mathcal{G}_{k+1:N}^{(s)} )\\ & + \frac{\rho}{2} \Vert \mathcal{G}_{k} - \mathcal{G}_{k}^{(s)} \Vert^2 \}
  \end{aligned}
  \right.
\end{equation}

1)Update $\mathbf{M}_k$: According Definition 11, the $\mathbf{M}_k$-subproblems can be rewritten as: 
\begin{equation}
  \begin{aligned}
    \mathbf{M}_{k} = \mathop{\arg\min}_{\mathbf{M}_{k}} &\frac{1}{2} \Vert \mathbf{X}_{(k)}^{(s)} -  \mathbf{M}_{k} (\mathbf{G}_{k})_{(k)}^{(s)} \mathbf{Y}_{(\neq k)}^{(s)} \Vert_F^2 \\
     &+ \frac{\rho}{2} \Vert \mathbf{M}_{k} - \mathbf{M}_{k}^{(s)} \Vert^2
  \end{aligned}
\end{equation}
where $\mathcal{Y}_{\neq k}^{(s)} = LMTN(\mathcal{Y}_{1:k-1}^{(s+1)},\mathcal{Y}_k,\mathcal{Y}_{k+1:N}^{(s)},/\mathcal{Y}_k)$, and the problem (8) 
can be directly solved as:
\begin{equation}
  \begin{aligned}
    \mathbf{M}_{k}^{(s+1)} = & [ \rho \mathbf{M}_k^{(s)} + \mathbf{X}_{(k)}^{(s)} \mathbf{Y}_{(\neq k)}^{(s)T} (\mathbf{G}_k)_{(k)}^{(s)T} ] \\ &( \rho \mathbf{I} + (\mathbf{G}_k)_{(k)}^{(s)} \mathbf{Y}_{(\neq k)}^{(s)} \mathbf{Y}_{(\neq k)}^{(s)T} (\mathbf{G}_k)_{(k)}^{(s)T} )^{-1} 
  \end{aligned}
\end{equation}
It should be noted that the latent matrices $\mathbf{M}_k$ are not necessary to be orthogonal in LMTN-PAM.

2)Update $\mathcal{G}_k$: According to Definition 11, the $\mathcal{G}_k$-subproblem can be rewritten as
\begin{equation}
  \begin{aligned}
    \mathbf{G}_{(k)}^{(s+1)} = &\mathop{\arg\min}_{\mathcal{G}_k} \frac{1}{2} \Vert \mathbf{X}_{(k)}^{(s)} -  \mathbf{M}_{k}^{(s+1)} (\mathbf{G}_{k})_{(k)} \mathbf{Y}_{(\neq k)}^{(s)} \Vert_F^2\\ & + \frac{\rho}{2} \Vert (\mathbf{G}_k)_{(k)} - (\mathbf{G}_k)_{(k)}^{(s)} \Vert^2
  \end{aligned}
\end{equation}
the problem (10) can be directly solved as
\begin{equation}
  \begin{aligned}
    & \rho (\mathbf{M}_k^{(s+1)T} \mathbf{M}_k^{(s+1)})^{-1} (\mathbf{G}_k)_{k} + (\mathbf{G}_k)_{k} \mathbf{Y}_{(\neq k)}^{(s)} \mathbf{Y}_{(\neq k)}^{(s)T} \\ = &( \mathbf{M}_k^{(s+1)T} \mathbf{M}_k^{(s+1)})^{-1} [\rho (\mathbf{G}_{k})_{(k)} + \mathbf{M}_k^{(s+1)T} \mathbf{X}_{(k)}^{(s)} \mathbf{Y}_{(\neq k)}^{(s)T}] 
  \end{aligned}
\end{equation}
In addition, equation (11) matches the form of the equation $\mathbf{AX} + \mathbf{XB} = \mathbf{C}$. The problem of solving this type of equation is the classical sylvester problem. 
By the properties of the Kronecker product, we have $vec(\mathbf{AXB}) = (\mathbf{B}^T\otimes \mathbf{A}) vec(X)$. Therefore, equation (11) can be rewritten as 
\begin{equation}
  \begin{aligned}
    (\mathbf{I}\otimes \rho \mathbf{M}_{k}^{*} + (\mathbf{Y}_{(\neq k)}^{(s)} \mathbf{Y}_{(\neq k)}^{(s)T})^T \otimes \mathbf{I} )vec((\mathbf{G}_k)_{(k)}) \\
    = vec( \mathbf{M}_{k}^{*} [\rho (\mathbf{G}_{k})_{(k)} + \mathbf{M}_k^{(s+1)T} \mathbf{X}_{(k)}^{(s)} \mathbf{Y}_{(\neq k)}^{(s)T}] )
  \end{aligned}
\end{equation}
Where $\mathbf{M}_{k}^{*} = (\mathbf{M}_k^{(s+1)T} \mathbf{M}_k^{(s+1)})^{-1}$. The solution for $(\mathbf{G}_k)_{(k)}$ was already obvious, but we can also use MATLAB function to solve for it:
\begin{equation}
  \begin{aligned}
    (\mathbf{G}_k)_{(k)}^{(s+1)} = sylvester(\rho \mathbf{M}_{k}^{*},~\mathbf{Y}_{(\neq k)}^{(s)} \mathbf{Y}_{(\neq k)}^{(s)T},\\~\mathbf{M}_{k}^{*} [\rho (\mathbf{G}_{k})_{(k)} + \mathbf{M}_k^{(s+1)T} \mathbf{X}_{(k)}^{(s)} \mathbf{Y}_{(\neq k)}^{(s)T}] )
  \end{aligned}
\end{equation}

3)Update $\mathcal{X}$: The problem of updating $\mathcal{X}$ is a least squares problem, and we can directly have the following closed-form solution:
\begin{equation}
  \begin{aligned}
    &\mathcal{X}^{(s+1)} =  \mathcal{P}_{\Omega} (\mathcal{T})+ \\ &\mathcal{P}_{ \overline{\Omega}} ( \frac{ LMTN(\{ \mathcal{G}_n^{(s+1)}\}_{n=1}^N,\{ \mathbf{M}_n^{(s+1)}\}_{n=1}^N)+\rho \mathcal{X}^{(s)} } {1+\rho})
  \end{aligned}
\end{equation}

Although Algorithm 1 is a model for tensor completion, when the observed tensor is complete, then our algorithm is a classical tensor decomposition algorithm. 
Furthermore, the factor matrices $\mathbf{M}_k$ are not strictly orthogonally constrained in this algorithm, so we consider other algorithms for the development of the model.

\begin{algorithm}[H]
  \setlength\baselineskip{12pt} 
  \caption{LMTN-PAM Algorithm} 
  \label{alg:Framwork} 
  \begin{algorithmic}[1] %这个1 表示每一行都显示数字
  \REQUIRE The incomplete tensor $\mathcal{T}\in \mathbb{R}^{I_1 \times \cdots \times I_N}$, the index $\Omega$, LMTN rank $\mathbf{R} \in \mathbb{R}^{N\times N}$, 
  maximum iterations $\mathbf{maxit}$, minimum error $\mathbf{tol}$ and hyperparameter $\rho$;\\ 
  \ENSURE The reconstructed tensor $\mathcal{X}$, factors $\{ \mathcal{G}_n \}_{n=1}^{N}$ and matrices $\{ \mathbf{M}_n \}_{n=1}^{N}$
  \STATE Randomly initialise $\{ \mathcal{G}_n^{(0)} \}_{n=1}^{N}$ and $\{ \mathbf{M}_n^{(0)} \}_{n=1}^{N}$, $s=1$.
  \label{ code:fram:extract }%对此行çš"标记,方便在文中引ç"¨ç®—法çš"æŸä¸ªæ­¥éª  \STATE $\mathbf{while}$ not converge and the $s<\mathbf{maxit}$, $\mathbf{do}$; 
  \label{code:fram:trainbase}
  \STATE ~~$\mathbf{for}$ k = 1 to N $\mathbf{do}$
  \label{code:fram:add}
  \STATE ~~~~Update $\mathbf{M}_k^{(s)}$ via (9);
  \label{code:fram:classify}
  \STATE ~~~~Update $\mathcal{G}_k^{(s)}$ via (13);
  \label{code:fram:classify}
  \STATE ~~$\mathbf{end~for}$
  \label{code:fram:classify}
  \STATE ~~Update $\mathcal{X}^{(s)}$ via (14);
  \label{code:fram:select}
  \STATE Check the convergence condition:
  ~~\begin{center}$\Vert \mathcal{X}^{(s)}- \mathcal{X}^{(s-1)} \Vert_F / \Vert \mathcal{X}^{(s-1)}\Vert_F < \mathbf{tol}$ \end{center}
  \label{code:fram:classify}
  \STATE ~~Let s = s+1;
  \label{code:fram:classify}
  \STATE $\mathbf{end~while}$
  \end{algorithmic}
\end{algorithm}

\subsection{LMTN-SVD algorithm}
We propose the second algorithm for cumputing the LMTN decomposition for tensor completion using $R_{k,k}$ sequential SVDs. This algorithm will 
be called LMTN algorithm. Given an incomplete observation $\mathcal{T}\in \mathbb{R}^{I_1\times I_2\times \cdots \times I_N}$
of the target tensor $\mathcal{X} \in \mathbb{R}^{I_1 \times I_2 \times \cdots \times I_N}$, our LMTN decomposition-based 
SVD model can be
\begin{equation}
  \begin{aligned}
    \min_{\mathcal{X},\mathcal{G},\mathbf{M}} & \frac{1}{2} \Vert \mathcal{X}\times_1 \mathbf{M}_1^T \cdots \times_N \mathbf{M}_N^T  - FCTN(\{ \mathcal{G}_n\}_{n=1}^N) \Vert_F^2 \\ & +  l_{\mathbb{S}}(\mathcal{X})
  \end{aligned}
\end{equation}
where 
$$
\iota_{\mathbb{S}}(\mathcal{X}):=\left\{\begin{aligned}
0&, \text { if } \mathcal{X} \in \mathbb{S },\\
\infty&, \text { otherwise, }
\end{aligned}\text{with } \mathbb{S}:=\left\{\mathcal{X}: \mathcal{P}_{\Omega}(\mathcal{X}-\mathcal{T})=0\right\}\right.
$$

For the equation (15) we can unfold it into the form of a matrix as 
\begin{equation}
  \begin{aligned}
    \min_{\mathbf{X}_{(k)},(\mathbf{G}_{k})_{(k)},\mathbf{M}_{k}} \frac{1}{2} \Vert \mathbf{M}_k \mathbf{X}_{(k)} \mathbf{M}_{\otimes k}^T - (\mathbf{G}_k)_{(k)}\mathbf{G}_{(\neq k)} \Vert_F^2 + l_{\mathbb{S}}(\mathcal{X})
  \end{aligned}
\end{equation}
where $\mathcal{G}_{\neq k} = FCTN(\{\mathcal{G}_{k=1}^N\},/\mathcal{G}_k)$, and $(\mathbf{G}_{\neq k})_{[m_{1:N-1};n_{1:N-1}]}$ is abbreviated to $\mathbf{G}_{(\neq k)}$ for convenience. 
In addition, $\mathbf{M}_{\otimes k} = \mathbf{M}_{k-1}^T \otimes \cdots \otimes \mathbf{M}_{1}^T \otimes \mathbf{M}_{N}^T \cdots \mathbf{M}_{k+1}^T$ and the operation $\otimes$ denotes 
the Kronecker product. 

1)Update $\mathbf{M}_k$: Suppose we have a tensor $\mathcal{B}_k$, which is obtained from the operation $\mathcal{X} \times_1 \mathbf{M}_{1}^T \cdots \times_{k-1} \mathbf{M}_{k-1}^T \times_{k+1} \mathbf{M}_{k+1}^T \cdots \times_{N} \mathbf{M}_{N}^T$. 
And we calculate the model-k unfolding of the tensor $\mathcal{B}_k$ to obtain $\mathbf{B}_{(k)}$. By using SVD, i.e., $\mathbf{B}_{(k)} = \mathbf{U} \mathbf{\Sigma} \mathbf{V}^T$, we can obtain the matrix $\mathbf{M}_k$ 
by appropriately reshaping $\mathbf{U}$, which is calculate by 
\begin{equation}
  \begin{aligned}
    \mathbf{M}_k^{(s+1)} = \mathbf{U}(:,1:R_{k,k})
  \end{aligned}
\end{equation}

2)Update $\mathcal{G}_k$: According to definition 11, the $\mathcal{G}_k$-subproblem can be rewritten as:
\begin{equation}
  \begin{aligned}
    \mathbf{G}_{(k)}^{(s+1)} = \mathop{\arg\min}_{(\mathbf{G}_{k})_{(k)}} \frac{1}{2} \Vert \mathbf{M}_k^{(s+1)} \mathbf{X}_{(k)}^{(s)} \mathbf{M}_{\otimes k}^{(s)T} - (\mathbf{G}_k)_{(k)}\mathbf{G}_{(\neq k)}^{(s)} \Vert_F^2
  \end{aligned}
\end{equation}
where $\mathcal{G}_{\neq k}^{(s)} = FCTN(\mathcal{G}_{1:k-1}^{(s+1)},\mathcal{G}_k,\mathcal{G}_{k+1:N}^{(s)},/\mathcal{G}_k)$ and $\mathbf{M}_{\otimes k}^{(s)T} = \mathbf{M}_{k-1}^{(s+1)T} \otimes \cdots \otimes \mathbf{M}_{1}^{(s+1)T} \otimes \mathbf{M}_{N}^{(s)T} \cdots \mathbf{M}_{k+1}^{(s)T}$. Hence the problem (18) 
can be solved as: 
\begin{equation}
  \begin{aligned}
    \mathbf{G}_{(k)}^{(s+1)} = \mathbf{M}_{k}^{(s+1)T} \mathbf{X}_{(k)}^{(s)} \mathbf{M}_{\otimes k}^{(s)T} \mathbf{G}_{(\neq k)}^{(s)T}[\mathbf{G}_{(\neq k)}^{(s)} \mathbf{G}_{(\neq k)}^{(s)T}]^{-1} 
  \end{aligned}
\end{equation}
And the eq.(19) also can be simplified as:
\begin{equation}
  \begin{aligned}
    \mathbf{G}_{(k)}^{(s+1)} = \mathbf{M}_{k}^{(s+1)T} \mathbf{X}_{(k)}^{(s)} \mathbf{M}_{\otimes k}^{(s)T} \mathbf{G}_{(\neq k)}^{(s)\dagger}
  \end{aligned}
\end{equation}

3)Update $\mathcal{X}$: The problem of updating $\mathcal{X}$ is a least squares problem, and we can easily have the solution:
\begin{equation}
  \begin{aligned}
    \mathcal{X}^{(s+1)} & = \mathcal{P}_{\overline{\Omega}} (LMTN(\{ \mathcal{G}_n^{(s+1)}\}_{n=1}^N,\{ \mathbf{M}_n^{(s+1)}\}_{n=1}^N)) \\
    & +\mathcal{P}_{\Omega} (\mathcal{T})
  \end{aligned}
\end{equation}

The iterations repeat until some combination of stopping conditions is satisfied. More details of the LMTN-SVD are given in Alg.2. 

\begin{algorithm}[H] 
  \caption{LMTN-SVD Algorithm} 
  \label{alg:Framwork} 
  \begin{algorithmic}[1] %这个1 表示每一行都显示数字
  \REQUIRE The incomplete tensor $\mathcal{T}\in \mathbb{R}^{I_1 \times \cdots \times I_N}$, the index $\Omega$, LMTN rank $\mathbf{R} \in \mathbb{R}^{N\times N}$, 
  maximum iterations $\mathbf{maxit}$, minimum error $\mathbf{tol}$;\\ 
  \ENSURE The reconstructed tensor $\mathcal{X}$, factors $\{ \mathcal{G}_n \}_{n=1}^{N}$ and matrices $\{ \mathbf{M}_n \}_{n=1}^{N}$
  \STATE Randomly initialise $\{ \mathcal{G}_n^{(0)} \}_{n=1}^{N}$ and $\{ \mathbf{M}_n^{(0)} \}_{n=1}^{N}$, $s=1$.
  \label{ code:fram:extract }%对此行çš"标记,方便在文中引ç"¨ç®—法çš"æŸä¸ªæ­¥éª  \STATE $\mathbf{while}$ not converge and the $s<\mathbf{maxit}$, $\mathbf{do}$
  \label{code:fram:trainbase}
  \STATE ~~$\mathbf{for}$ k = 1 to N $\mathbf{do}$
  \label{code:fram:add}
  \STATE ~~~~$\begin{aligned} &\text{Obtain}~\mathcal{B}_k^{(s)}~\text{via}~\mathcal{X} \times_1 \mathbf{M}_{1}^T \cdots \times_{k-1} \mathbf{M}_{k-1}^T \times_{k+1} \\ & \mathbf{M}_{k+1}^T \cdots \times_{N} \mathbf{M}_{N}^T \text{;}\end{aligned}$
  \label{code:fram:add}
  \STATE ~~~~Update $\mathbf{M}_k^{(s)}$ via (17);
  \label{code:fram:classify}
  \STATE ~~~~Update $\mathcal{G}_k^{(s)}$ via (19) or (20);
  \label{code:fram:classify}
  \STATE ~~$\mathbf{end~for}$
  \label{code:fram:classify}
  \STATE ~~Update $\mathcal{X}^{(s)}$ via (21);
  \label{code:fram:select}
  \STATE Check the convergence condition:
  ~~\begin{center}$\Vert \mathcal{X}^{(s)}- \mathcal{X}^{(s-1)} \Vert_F / \Vert \mathcal{X}^{(s-1)}\Vert_F < \mathbf{tol}$ \end{center}
  \label{code:fram:classify}
  \STATE ~~Let s = s+1;
  \label{code:fram:classify}
  \STATE $\mathbf{end~while}$
  \end{algorithmic}
\end{algorithm}

\subsection{LMTN-AR algorithm}
One important limitation of LMTN decomposition is that LMTN rank must be fixed, which may difficult to obtain a good performance. 
Although we can change the rank of the LMTN, the computation and time will increase. Therefore, we try to design an algorithm to auto increase 
our LMTN rank, which is called LMTN-AR algorithm. 

The tensor decomposition with auto ranks strategy\cite{10} suggests that the factors should start from rank-1 tensor. 
Therefore, the LMTN-AR algorithm need to initialize $R_{i,j}$ with $i,j\in \mathbb{R}^N$ to the small value, such as 1 or 2. And then we need to 
design the maximum rank $R_{i,j}^{'}$ with $i,j\in \mathbb{R}^N$ in $\mathbf{R}_{max} \in \mathbb{R}^{N\times N}$. For the optimization of each 
factor $\mathcal{G}_k$ and matrix $\mathbf{M}_k$, it was updated according to SVD algorithm firstly, getting the 
updated approximation error $\varepsilon$. After each iteration, $\varepsilon$ is computed and compared to the specific criteria. And 
if satisfied the criteria, we will increase the rank by $R_{i,j} = R_{i,j}+1$ with $i \neq j$. The specific criteria can be expressed by: 
\begin{equation}
  \begin{aligned}
    | \varepsilon^{(s+1)} - \varepsilon^{(s)} | > \tau | \varepsilon^{(s)} - \mathbf{tol} | 
  \end{aligned}
\end{equation}

The reasonable choices for $\tau$ value between 0.1 and 0.5. In addition, considering that $R_{i,i}$ may have a large range of choices, 
we build a gradient descent method to update it. The loss function is first simply set up as:
\begin{equation}
  \begin{aligned}
    L_i(R_{i,i}) = \frac{1}{2} (R_{i,i}^{'} - R_{i,i})^2
  \end{aligned}
\end{equation}
Where the $R_{i,i}^{'}$ is the expected maximum rank. And we calculate its gradient of $R_{i,i}$ to obtain $\bigtriangledown L_i(R_{i,i})$, and we update the $R_{i,i}$ value each time from:
\begin{equation}  
  \begin{aligned}
    R_{i,i}^{(s+1)} = R_{i,i}^{(s)} -\alpha \bigtriangledown L_i(R_{i,i}^{(s)})
  \end{aligned}
\end{equation}
Where the parameter $\alpha$ is the stride of the gradient descent, which we generally choose to be 0.05 to 0.2. After updating their rank each time, we also update the 
size of the factor $\mathcal{G}_k$ and the matrix $\mathbf{M}_k$. We keep the original data unchanged and the 
expanded parts are updated with additional random filling of the new data. We can see more details in Alg.3. 

This algorithm is convenient and is usually able to find a low-rank representation to approximate the original tensor automatically. 
We can have $\mathbf{R}_{max}$ as a limit on the largest size of factors even when we choose $\tau$ inappropriately, which ensures that the algorithm 
can achieve good recovery results in a reasonably small area. 

\begin{algorithm}[H] 
  \caption{LMTN-AR Algorithm.} 
  \label{alg:Framwork} 
  \begin{algorithmic}[1] %这个1 表示每一行都显示数字
  \REQUIRE The incomplete tensor $\mathcal{T}\in \mathbb{R}^{I_1 \times \cdots \times I_N}$, the index $\Omega$, maximum rank of LMTN $\mathbf{R}_{max} \in \mathbb{R}^{N\times N}$, threshold $\tau$, 
  maximum iterations $\mathbf{maxit}$ and minimum error $\mathbf{tol}$;\\ 
  \ENSURE The reconstructed tensor $\mathcal{X}$, factors $\{ \mathcal{G}_n \}_{n=1}^{N}$ and matrices $\{ \mathbf{M}_n \}_{n=1}^{N}$
  \STATE Initialise $R_{i,j}=2~\text{for}~i,j=1,2\cdots,N $. 
  \label{ code:fram:extract }%对此行çš"标记,方便在文中引ç"¨ç®—法çš"æŸä¸ªæ­¥éª  \STATE randomly initialise $\{ \mathcal{G}_n^{(0)} \}_{n=1}^{N}$ and $\{ \mathbf{M}_n^{(0)} \}_{n=1}^{N}$, $s=1$.
  \label{ code:fram:extract }%对此行çš"标记,方便在文中引ç"¨ç®—法çš"æŸä¸ªæ­¥éª  \STATE $\mathbf{while}$ not converge and the $s<\mathbf{maxit}$, $\mathbf{do}$
  \label{code:fram:trainbase}
  \STATE ~~$\mathbf{for}$ k = 1 to N $\mathbf{do}$
  \label{code:fram:add}
  \STATE ~~~~$\begin{aligned} &\text{Obtain}~\mathcal{B}_k^{(s)}~\text{via}~\mathcal{X} \times_1 \mathbf{M}_{1}^T \cdots \times_{k-1} \mathbf{M}_{k-1}^T \times_{k+1} \\ & \mathbf{M}_{k+1}^T \cdots \times_{N} \mathbf{M}_{N}^T \text{;}\end{aligned}$
  \label{code:fram:add}
  \STATE ~~~~Update $\mathbf{M}_k^{(s)}$ via (17);
  \label{code:fram:classify}
  \STATE ~~~~Update $\mathcal{G}_k^{(s)}$ via (19) or (20);
  \label{code:fram:classify}
  \STATE ~~$\mathbf{end~for}$
  \label{code:fram:classify}
  \STATE ~~Update $\mathcal{X}^{(s)}$ via (21);
  \label{code:fram:classify}
  \STATE ~~Evaluate relative error $\varepsilon^{(s)}$;
  \label{code:fram:classify}
  \STATE ~~$\mathbf{If}$ satisfied (22);
  \label{code:fram:classify}
  \STATE ~~~~$R_{i,j} = R_{i,j} + 1~\text{for}~i\neq j$;
  \label{code:fram:classify}
  \STATE ~~~~Update $R_{i,j}~\text{for}~i = j$ via (24);
  \label{code:fram:classify}
  \STATE ~~$\mathbf{end~If}$;
  \label{code:fram:classify}
  \STATE ~~$\begin{aligned}&\text{Increase~the~size~of~factors~}\mathcal{G}_k^{(s)}\text{~and~} \mathbf{M}_k^{(s)} \text{~by~random~}\\ &\text{sample;} \end{aligned}$
  \label{code:fram:select}
  \STATE ~~Check the convergence condition:
  ~~\begin{center}$\Vert \mathcal{X}^{(s)}- \mathcal{X}^{(s-1)} \Vert_F / \Vert \mathcal{X}^{(s-1)}\Vert_F < \mathbf{tol}$ \end{center}
  \label{code:fram:classify}
  \STATE ~~Let s = s+1;  
  \label{code:fram:classify}
  \STATE $\mathbf{end~while}$
  \end{algorithmic}
\end{algorithm}

\subsection{Convergence Analysis and Computational Complexity}
In this part, we first established the convergence of our algorithms. For the convergence analysis of the LMTN-PAM algorithm, we refer to the proofs carried 
out in the work of the literature\cite{25,26}. 

\noindent $\mathbf{Theorem~2(LMTN-PAM~convergence)}$ For the sequence \{$\mathcal{G}^{(s)}$, $\mathbf{M}^{(s)}$, $\mathcal{X}^{(s)}$\} global convergence to a critical point obtained by LMTN-PAM. 

In order to prove Theorem 2, we only need to prove the following conditions.

(a) If $\mathcal{G}_k^{(0)}(k=1,2,\cdots,N)$, $\mathbf{M}_k^{(0)}$ and $\mathcal{X}_k^{(0)}$ are bounded, \{$\mathcal{G}^{(s)}$\}, \{$\mathbf{M}^{(s)}$\} and \{$\mathcal{X}^{(s)}$\} are the bounded sequences;

(b) $f(\mathcal{G},\mathbf{M}, \mathcal{X})$ is a proper lower semi-continuous function;

(c) $f(\mathcal{G},\mathbf{M}, \mathcal{X})$ satisfies the Kurdyka–Łojasiewicz property\cite{27} at \{ $\mathcal{G}^{(s)}$, $\mathbf{M}^{(s)}$, $\mathcal{X}^{(s)}$\};

(d) \{$\mathcal{G}^{(s)}$, $\mathbf{M}^{(s)}$, $\mathcal{X}^{(s)}$\} satisfied lemmas 1 and 2. 

For brevity, our specific proof can be found in supplementary material. And for the LMTN-SVD algorithm, we refer to the theorems and proofs in the literature\cite{28} to give Theorem 3 below. Similarly, the proof can be found in supplementary material.

\noindent $\mathbf{Theorem~3(LMTN-SVD~convergence)}$ Let $\{ \mathbf{u}\}$ be a sequence of $\{ \mathcal{G}_k, \mathbf{M}_k, \mathcal{X} \}$ generated by LMTN-SVD algorithm. 
For any $s$, if $\mathbf{u}^{(s)}$ is not a stationary point of $f(\mathbf{u}^{(s)})$, i.e., $\nabla f(\mathbf{u}^{(s)})\neq 0$, then $f(\mathbf{u}^{(s+1)}) \leq f(\mathbf{u}^{(s)})$.

Since the LMTN-AR algorithm's convergence like the LMTN-SVD algorithm after the appropriate rank is chosen. Therefore, we will not repeat the convergence analysis of the LMTN-AR algorithm. 

Furthermore, we analyse the computation complexity of our algorithm. We use $R_1$ to denote $(R_{i,j})_{i\neq j}$ and $R_2$ to denote $(R_{i,j})_{i=j}$ for analysis convenience, and the rank of the other algorithms is uniformly denoted by $R$. 
Assuming the $N$th-order incomplete tensor $\mathcal{X}\in \mathbb{R}^{I_1\times I_2 \times I_3 \cdots \times I_N}$ with $I = I_1 = I_2 = I_3 \cdots = I_N$. For the LMTN-SVD algorithm, the computation cost mainly includes two parts: 
(1)the contraction of $\mathcal{G}_k(k\in N)$ and $\mathcal{B}_k$. (2)Updating of $\mathcal{G}_k$ and $\mathbf{M}_k$. 
For the contraction of $\mathcal{G}_k$ and $\mathcal{B}_k$, our computational complexity is $O(N\sum_{k=2}^{N}I^kR_1^{k(N-k)+k-1})$ and $O(N\sum_{k=2}^{N-1}R_2^{k-1}I^{N+2-k})$ respectively. 
Similarly, the complexity of updating $\mathcal{G}_k$ and $\mathbf{M}_k$ are $O(NIR_2^N+NR_1^{2(N-1)}R_2^{N-1}+NR_1^{3(N-1)}R_2)$ and $O(NI^3)$. 
More over, the computational complexity of the final $\mathcal{X}$ contrction is negligible in the overall algorithm. 
Therefore, the computational complexity of the LMTN-SVD algorithm is $O(N\sum_{k=2}^{N}I^kR_1^{k(N-k)+k-1} + N\sum_{k=2}^{N-1}R_2^{k-1}I^{N+2-k} + NIR_2^N+NR_1^{2(N-1)}R_2^{N-1}+NR_1^{3(N-1)}R_2 + NI^3)$. 
Using the same method, we can also calculate the computational complexity of our LMTN-PAM algorithm, which can be seen in Table \uppercase\expandafter{\romannumeral1}. 
\begin{table}[]
  \caption{The computational complexity of different methods on an-order tensor of size $I\times \cdots \times I$. }
  \resizebox{1\columnwidth}{!}{
  \begin{tabular}{c|c}
  \hline
  Method   & Computational complexity                            \\ \hline
  TR-ALS   & $O(NPR^4I^N + NR^6) $                                   \\ \hline
  HaLRTC   & $O(NI^{N+1})$                                            \\ \hline
  TR-WOPT  & $O(NR^2I^N + NR^4I^{N-1})$                                  \\ \hline
  TRLRF    & $O(NR^2I^N + NR^6)$                                      \\ \hline
  FCTN-PAM & $O(N\sum_{k=2}^{N} I^kR^{k(N-k)+k-1} + NI^{N-1}R^{2(N-1)} + NR^{3(N-1)})$ \\ \hline
  LMTN-SVD & $\begin{aligned} O(N\sum_{k=2}^{N}I^kR_1^{k(N-k)+k-1} + N\sum_{k=2}^{N-1}R_2^{k-1}I^{N+2-k} \\+ NIR_2^N+NR_1^{2(N-1)}R_2^{N-1}+NR_1^{3(N-1)}R_2 + NI^3)\end{aligned}$            \\ \hline
  LMTN-PAM & $\begin{aligned}O(N\sum_{k=2}^NI^kR_2^{k(N-k)+k-1}+NI^{N-1}R_1^{N-1}R_2+NI^NR_2\\+NR_2^3+NI^{N-1}R_1^{2(N-1)}+NR_2^2R_1^{N-1})\end{aligned}$                                                    \\ \hline
  \end{tabular}}
\end{table}

Observing the table above we have no difficulty in finding that the computational complexity of our algorithm is the largest. 
But relative to the FCTN-PAM algorithm, our algorithm can give good experimental results much faster for suitable R1 and R2, details of which can be found in Table \uppercase\expandafter{\romannumeral4}. 
The reason is that we are better able to find latent matrices with the smaller ranks of the chosen tensor. 
At the same time, we analyse the mainstream tensor decomposition algorithm of how many parameters they require to represent the original tensor, which can be seen in table \uppercase\expandafter{\romannumeral2} in detail. 

\begin{table}[]
  \centering
  \caption{The number of parameters required of different tensor decomposition algorithms.}
  \begin{tabular}{c|c}
  \hline
  Method   & Computational complexity                            \\ \hline
  CP   & $O(NIR) $                                   \\ \hline
  Tucker   & $O(NIR+R^N)$                                            \\ \hline
  TT  & $O((N-2)IR^2+2IR)$                                  \\ \hline
  TR    & $O(NIR^2)$                                      \\ \hline
  FCTN & $O(NIR^{N-1})$ \\ \hline
  LMTN & $O(NR_2R_1^{N-1}+NIR_2)$            \\ \hline
  \end{tabular}
\end{table}

In fact, the size of our algorithm $R_2$ is always smaller than $I$, generally half the size of $I$. In particular, when dealing with higher-order tensors, 
we are able to reduce the number of parameters in the FCTN decomposition's size of the exponential several times. 

\section{Numerical Experiments}
In this section, we evaluate the proposed methods in different situations to show the superior performance of the proposed methods. 
We simply set the tensor of the same value of rank $\{\mathbf{R}_{i,j}\}_{i\neq j}$ inside the experiment and the value of rank $\{\mathbf{R}_{i,j}\}_{i=j}$ 
will be set as the special number. Then, more specific settings will be given in the subsection. 

\subsection{Impact of hyperparameter $\tau$}
We test the variation of the psnr of the LMTN-AR algorithm for different values of the hyperparameter $\tau$ at different missing rates. 
The first experiment is run on the HSV\footnote{The data is available at http://openremotesensing.net/kb/data/.} dataset, where we set the maximum rank $\mathbf{R}_{i,j}$ of the tensor $\mathcal{X}$, the maximum number of iterations to 50 and the minimum error to $10^{-6}$. 
Our general approach to designing the maximum rank $\{\mathbf{R}_{i,j}\}_{i=j}$ is 20 and the maximum rank $\{\mathbf{R}\}_{i\neq j}$ is typically 5. 
In the second experiment, we tested on the container video and all the parameters remained the same value except for the hyperparameter $\tau$. 
We use psnr and runtime as indicators and the details can be seen in Fig. 3. 

\begin{figure*}
  \centering
  \subfigure[HSV]{
    \begin{minipage}{0.26\linewidth}
      \includegraphics[height = 1.7in,width=2in]{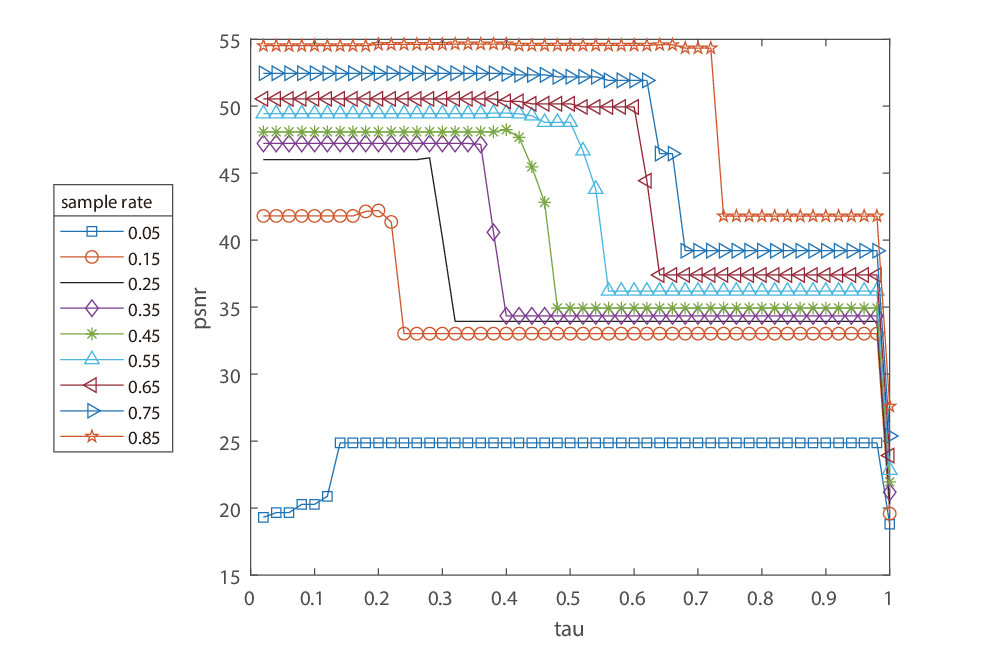}
    \end{minipage}
    \begin{minipage}{0.23\linewidth}
      \includegraphics[height = 1.7in,width=1.7in]{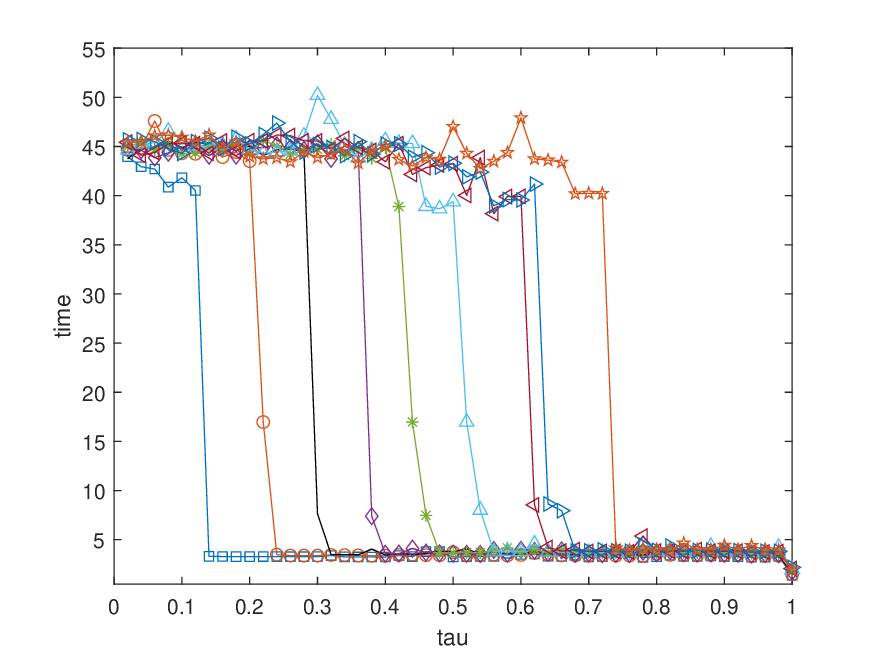}
    \end{minipage}
  }
  \subfigure[contianer]{
    \begin{minipage}{0.23\linewidth}
      \includegraphics[height = 1.7in,width=1.7in]{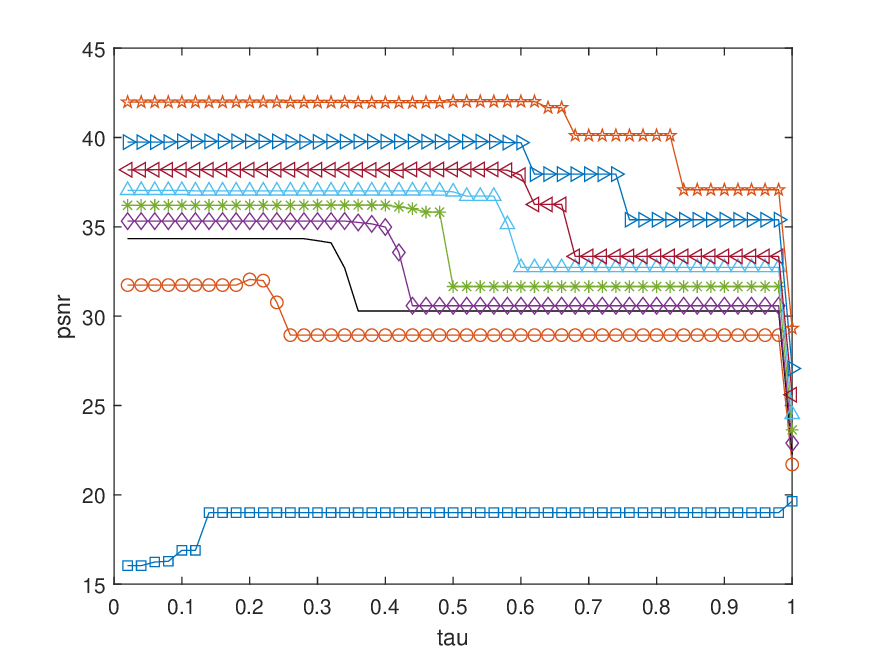}
    \end{minipage}
    \begin{minipage}{0.23\linewidth}
      \includegraphics[height = 1.7in,width=1.7in]{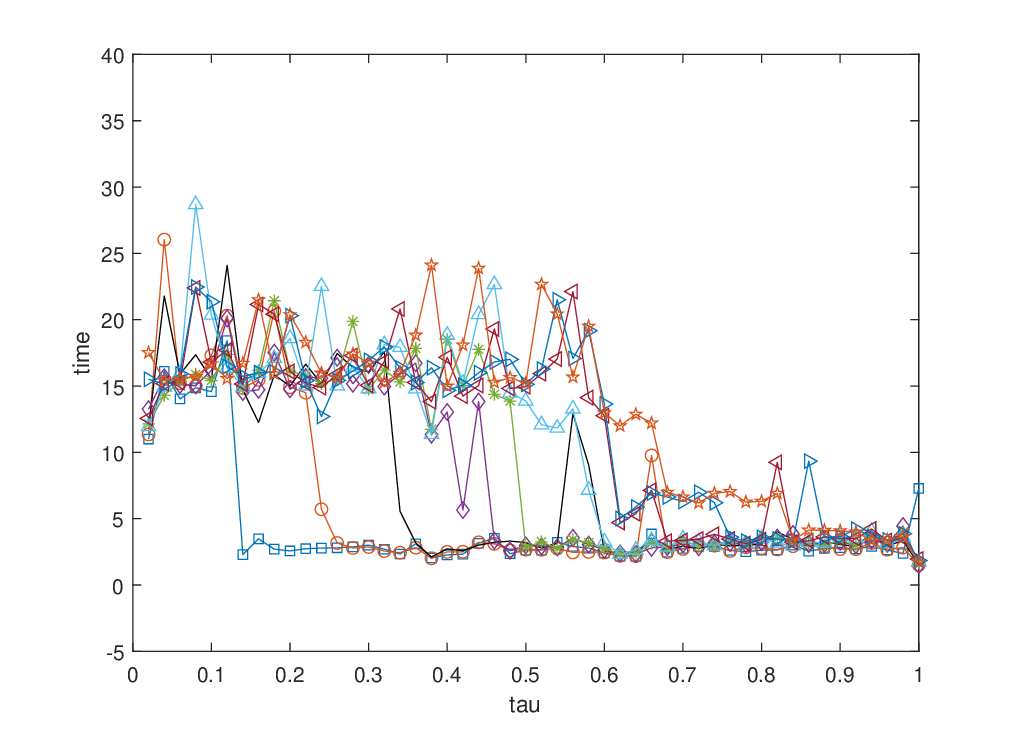}
    \end{minipage}
  }
      \caption{The first picture within groups (a) and (b) presents psnr as the vertical coordinate and $\tau$ as the horizontal coordinate, indicating the value of the algorithm's psnr at different $\tau$ values. 
      In addition, the different coloured curves indicate the experiments performed at different missing rates. The second figure in groups (a) and (b) plotted as time in the vertical coordinate and $\tau$ in 
      the horizontal coordinate, indicating the running time of the algorithm at different $\tau$ values.}
  \label{fig_3}
\end{figure*}

Observe that, when the value of hyperparameter $\tau$ between 0.2 to 0.8, the result will become unstable. Good results will be run when $\tau$ is less than around 0.2. When the hyperparameter $\tau$ chosen  
is greater than around 0.8, the normal result is also run, but the running time is also fast because the algorithm chooses a tiny size to the factors and the amount 
of data can be compressed obviously. In summary, the experiments show that we can pick our own $\tau$ value between 0 and 1, depending on our needs. 

\subsection{Impact of Tensor Transposition}
The fully-connected tensor network decomposition has the good performance on the correlation characterization and transposition invariance. 
In this subsection, we try to verify the transposition invariance of our LMTN algorithm experimentally and the accuracy of our algorithm is the best. 
We tested these algorithms on video container\footnote{The data is available at http://trace.eas.asu.edu/yuv/.} and hyperspectral images HSV, both of which have a missing rate of 80\%. 
All the data are arranged in a different order to obtain the new data, and then performed a tensor-completion experiment on them. 
We compared our proposed method to the TMac\cite{20}, TR-WOPT\cite{21}, HaLRTC\cite{22}, and TRLRF\cite{23} algorithm, where the maximum iterations is 50 and the minimum error is set to $10^{-5}$. 
In the following two groups of experiments, the rank of our LMTN algorithms in hyperspectral images of HSV (size 60 × 60 × 20 × 20) was $\mathbf{R}_{1,1} = \mathbf{R}_{2,2} = 70$, $\mathbf{R}_{3,3}=3$, $\mathbf{R}_{4,4}=10$ and $\{\mathbf{R}_{i,j}\}_{i\neq j} = 5$, and the 
rank of the other algorithm, such as TR-WOPT and TRALS were chosen to be 10. In addition, TMac and HaLRTC do not need not be chosen the rank since 
they can use tensor nuclear norm to approximate the rank of the tensor directly. In particular, we set the TMac algorithm number of iteration to 1000 in order 
to obtain better results. We measure the final completed data with PSNR and the results are shown in Fig. 4. 

In the figure we can clearly observe that all of our proposed algorithms (LMTN-ALS, LMTN-SVD, LMTN-AR) have good translational invariance 
and our proposed algorithm has the best completion results of these algorithms. This is because the factors decomposed by our 
proposed decomposition algorithm can be linearly conjoined with each other, whereas the factors decomposed by other algorithms do not have 
this particular structure. However, we can also find that the HaLRTC algorithm also have excellent robustness, due to the fact 
that their algorithm is solved directly for the individual mode unfolding of the tensor with tensor nuclear norm, without using the 
decomposition method for the calculation. 
\begin{figure}[H]
  \centering
  \subfigure[container]{
    \begin{minipage}{0.47\linewidth}
      \includegraphics[height = 1.7in,width=1.7in]{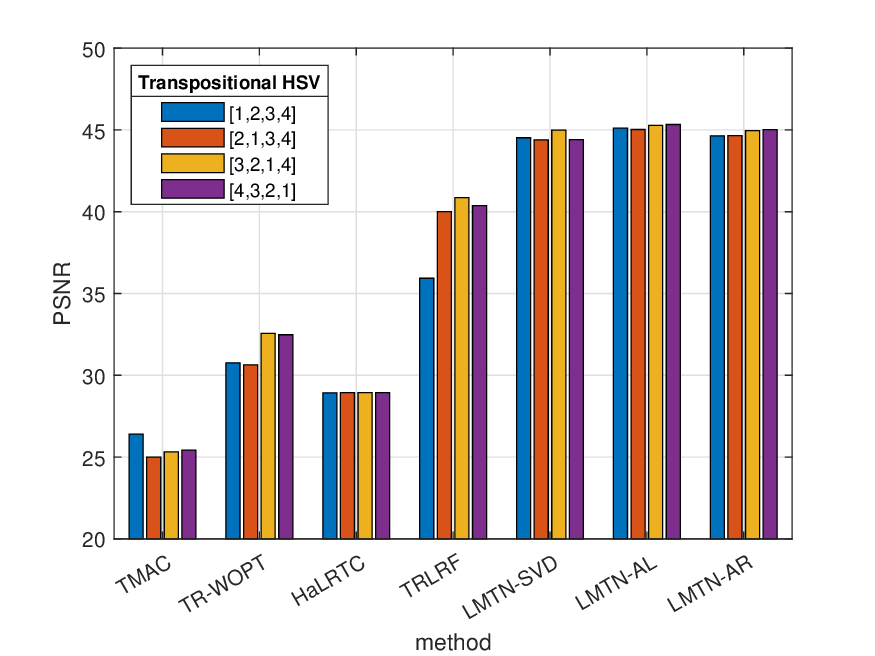}
    \end{minipage} 
  }
    \subfigure[HSV]{
      \begin{minipage}{0.47\linewidth}
        \includegraphics[height = 1.7in,width=1.7in]{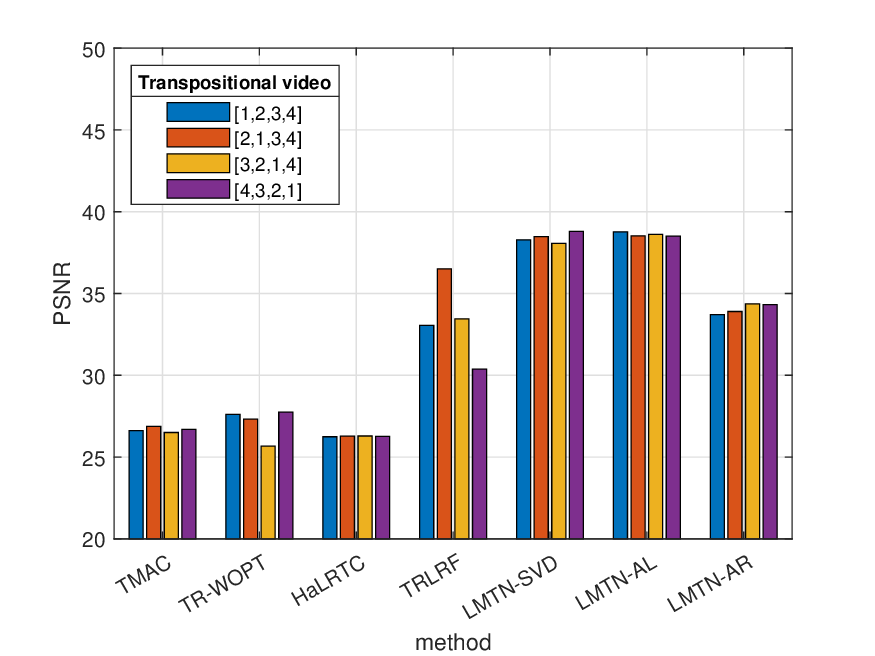}
      \end{minipage}
  }
      \caption{Reconstructed result on the different datasets with different permutations.}
  \label{fig_4}
\end{figure}
\subsection{Deep Learning Dataset Compression}
In this subsection, our goal was to compress the deep learning dataset and test the compression rate, 
recovery error and running time. The two deep learning datasets we used were CIFAR10\cite{41} (32$\times$32$\times$3$\times$10000) with 1.5$\times~10^8$ elements, 
and COIL100\cite{42} (32$\times$32$\times$3$\times$72$\times$100) with 2.2$\times~10^7$ elements. We then propose 
a measurement of the compression ratio (CR) for our algorithm, which is calculated as $CR = N_f/N_x$, where 
$N_f$ denotes the sum of the number of model internal elements and $N_x$ denotes the sum of the 
number of original data set elements. 
Different compression performance occurs with different ranks chosen. To make it easier to observe the compressive 
and recovery performance of our algorithm, we fixed the rank of the FCTN-PAM algorithm and then changed the 
other ranks of our algorithm. 

In Table \uppercase\expandafter{\romannumeral3}, we can observe that different errors and running times appear for our algorithm at different 
compression ratios. In CIFAR10, the RSE of our LMTN-PAM algorithm is already smaller than FCTN-PAM 
when the compression ratio is half of FCTN-PAM. And when the compression ratio is close to 
FCTN-PAM, the RSE of our proposed algorithms is smaller except for the LMTN-AR algorithm. 
We also find that sometimes LMTN-AR is not quite stable, this is because the different selection of $\tau$ on different datasets 
can lead to different results. But LMTN-AR are able to find a second-best decomposition result automatically without the need to set the tensor 
rank artificially. In coil100, we found that our LMTN-SVD algorithm is 6 to 7 times faster than the FCTN-PAM algorithm with a 0.42 points RSE raise when our CR metric is smaller. 
The decomposition error of our algorithm can have better results than the FCTN-PAM algorithm because the potential ranks 
inherent in the original data is smaller than the ranks of the FCTN-PAM algorithm. However, the FCTN decomposition has a large span of parameters 
for each rank selection on higher-order data, which is likely to miss the best representation of the tensor. On the contrary, the LMTN algorithm 
can mine this potential information, which is the main reason for its small decomposition error.

\begin{table}[H]
  \centering
  \caption{Comparison of data compression performance}
  \resizebox{.95\columnwidth}{!}{
  \begin{tabular}{c|cccccc}
  \hline
  \hline
           & \multicolumn{6}{c}{Cifar10}                                                     \\ \hline
           & CR      & RSE     & \multicolumn{1}{c|}{time}    & CR       & RSE     & time    \\ \hline
  FCTN-PAM & 0.0088 & 0.2912 & \multicolumn{1}{c|}{15.763} & 0.0088  & 0.2912 & 15.981 \\
  LMTN-SVD & 0.0049 & 0.3072 & \multicolumn{1}{c|}{10.195} & 0.0082  & 0.2865 & 14.163 \\
  LMTN-PAM & 0.0049 & 0.2854 & \multicolumn{1}{c|}{18.201} & 0.0082  & 0.2755 & 21.644 \\
  LMTN-AR  & 0.0046 & 0.3557 & \multicolumn{1}{c|}{73.829} & 0.0087  & 0.2875 & 24.545 \\ \hline
           & \multicolumn{6}{c}{Coil100}                                                     \\ \hline
           & CR      & RSE     & \multicolumn{1}{c|}{time}    & CR       & RSE     & time    \\ \hline
  FCTN-PAM & 1.74e-4 & 0.4258 & \multicolumn{1}{c|}{6.8708} & 1.74e-4  & 0.4258 & 7.1729  \\
  LMTN-SVD & 1.12e-4 & 0.4272 & \multicolumn{1}{c|}{1.1133} & 9.66e-5 & 0.4224 & 1.1498 \\
  LMTN-PAM & 1.12e-4 & 0.4234 & \multicolumn{1}{c|}{8.9041} & 9.66e-5 & 0.4128 & 9.0000 \\
  LMTN-AR  & 1.04e-5 & 0.4588 & \multicolumn{1}{c|}{2.4121} & 8.30e-5 & 0.4608 & 2.4549 \\
  \hline 
  \hline
  \end{tabular}}
\end{table}

\subsection{Hyperspecral Image Recovery}
We use two hyperspectral images HSI (200$\times$200$\times$80) and HSV (60$\times$60$\times$20$\times$20) at different missing rate for tensor completion experiments to see the recovery effect of them respectively. 
We also test several completion algorithms, and the experimental demonstrations are taken to show their best results. 
In our experiments we evaluate all methods using the PSNR and SSIM indexes, and in Fig. 5 we show the PSNR in a box line plot, and in Fig. 6 we show the PSNR and time in a line graph. 
Hyperspectral images are tested with the missing rate set inside the set \{95\%, 85\%, 75\%, 65\%, 55\%, 45\%, 35\%, 25\%, 15\%, 5\%\}. 
For the TMac algorithm we set its maximum iterations to 1000 and for all other algorithms the maximum iterations are 300. 
In this case, the rank $(\mathbf{R}_{i,j})_{i\neq j}$ of our proposed algorithm and the rank of the FCTN-PAM algorithm are both simply set the same, and the $\tau$ of the LMTN-AR algorithm is also set to 0.1. 

In Fig. 5, we can clearly observe that our proposed algorithm is close to and sometimes even better than the FCTN-PAM algorithm above the recovery results for different missing rates, and is also 
significantly better than the other algorithms. At the same time, the mean value of PSNR is close to the FCTN-PAM algorithm. And all our PSNR values are more concentrated and have better robustness 
like FCTN-PAM algorithm. These experimental results also demonstrate the good performance of our proposed algorithm for the completion task of hyperspectral images. 
Although the TR-WOPT algorithm also has concentrated values of PSNR for different missing rates, their algorithm is less accurate and takes much more time(see in the Fig. 6). 

In Fig. 6, we can observe the variation in accuracy and time of the various algorithms for the two hyperspectral image data at different sampling rates. 
We have chosen 20*log(1+time) as the vertical coordinate when plotting the line chart for running time, because the algorithms like TR-WOPT and TRLRF have long 
running times and the degree of variation of each algorithm cannot be clearly observed in a normal line chart. 
In particular, although the TMac and HaLRTC algorithms are faster in running time, the accuracy of their algorithms are less accurate than the others. 
This phenomenon is more evident for the fourth-order and higher-order tensor. Especially, when the HSV data is at a high missing rate, our algorithm 
is also 5 dB better than other algorithms for the PSNR value except for the FCTN-PAM algorithm. 

Although the accuracy of the FCTN-PAM algorithm will be higher, the running time of our algorithm is faster in most cases. This is because the number of parameters is already compressed and 
the running time required is also reduced. However, running the LMTN-PAM algorithm takes longer occasionally because it does not constrain the orthogonality of the latent matrices internally, which 
leads to the non-uniqueness of the latent matrices.

\begin{figure}[H]
  \centering
  \subfigure[HSV]{
    \begin{minipage}{0.45\linewidth}
      \includegraphics[height = 1.3in,width=1.55in]{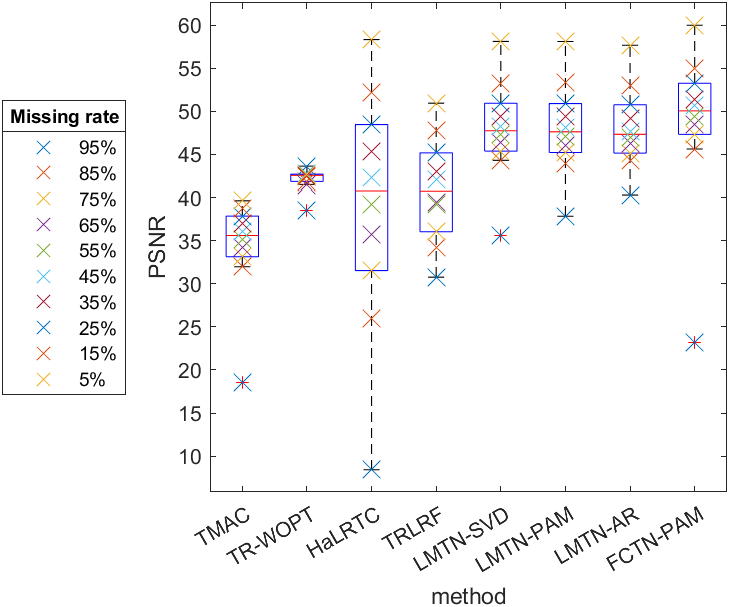}
    \end{minipage} 
  }
    \subfigure[HSI]{
      \begin{minipage}{0.45\linewidth}
        \includegraphics[height = 1.3in,width=1.25in]{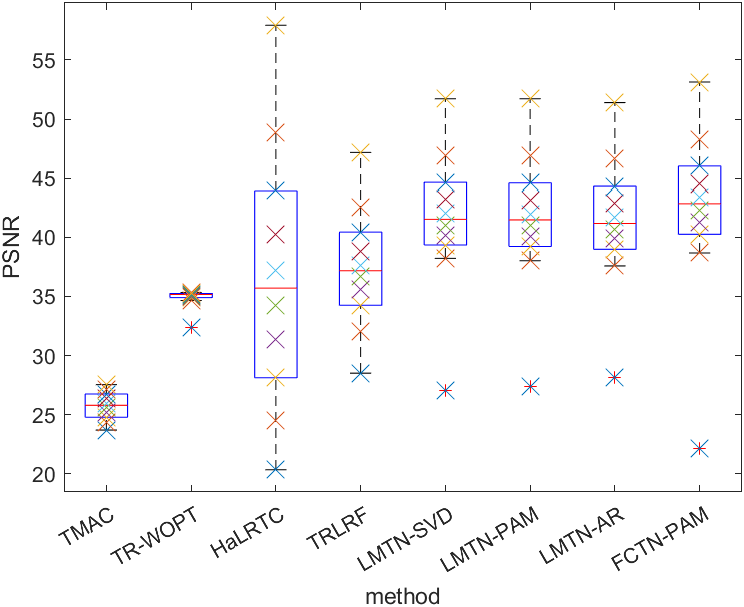}
      \end{minipage}
  }
      \caption{Box line plot of experimental results on two different hyperspectral image datasets. The points of the red crosses indicate the outliers determined and the red lines indicate the average of all results. 
      The points of the red crosses indicate outliers and the red lines indicate the average of all results.}
  \label{fig_5}
\end{figure}

\begin{figure*}
  \centering
  \subfigure[HSV]{
    \begin{minipage}{0.23\linewidth}
      \includegraphics[height = 1.7in,width=1.7in]{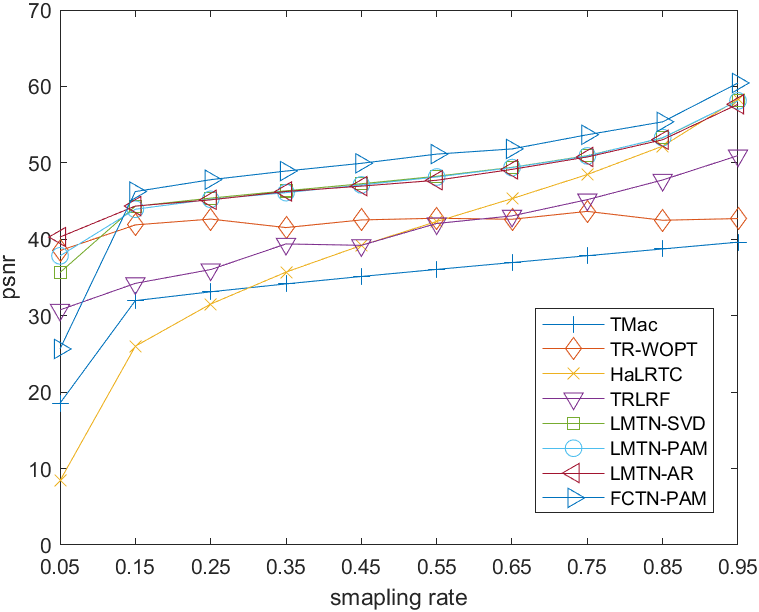}
    \end{minipage}
    \begin{minipage}{0.23\linewidth}
      \includegraphics[height = 1.7in,width=1.7in]{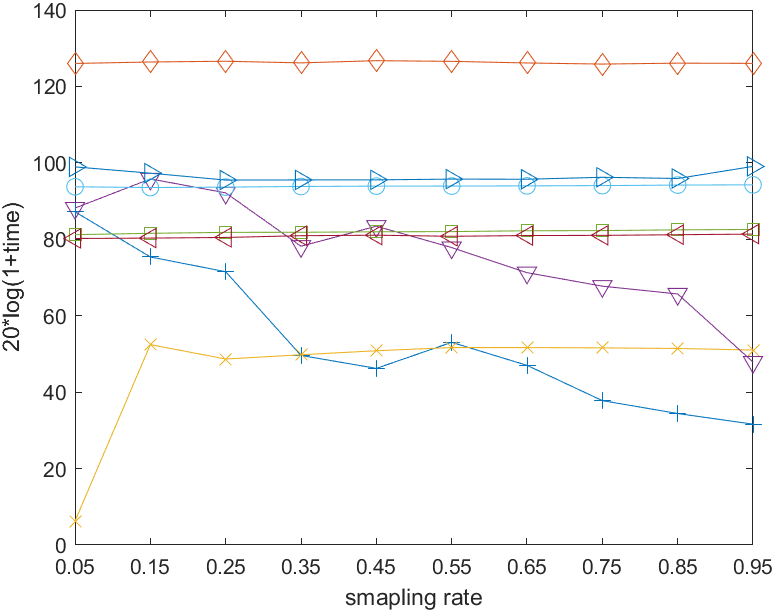}
    \end{minipage}
  }
  \subfigure[HSI]{
    \begin{minipage}{0.23\linewidth}
      \includegraphics[height = 1.7in,width=1.7in]{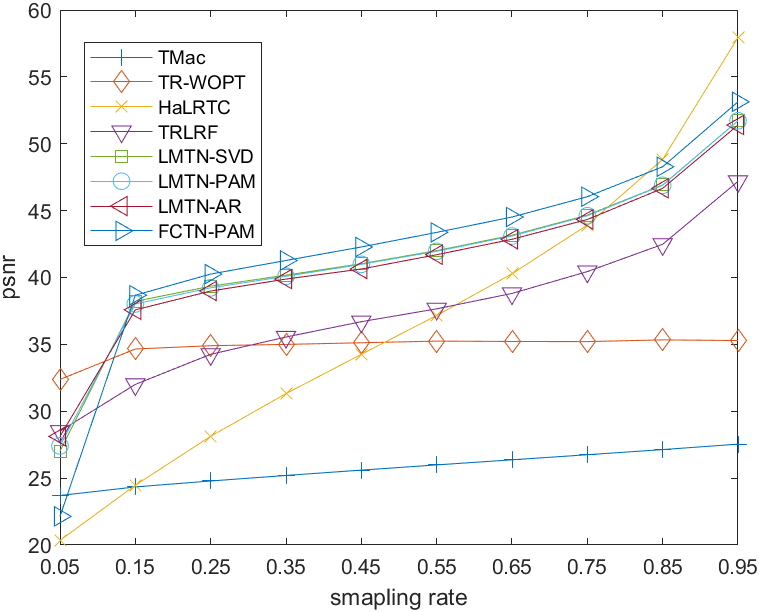}
    \end{minipage}
    \begin{minipage}{0.23\linewidth}
      \includegraphics[height = 1.7in,width=1.7in]{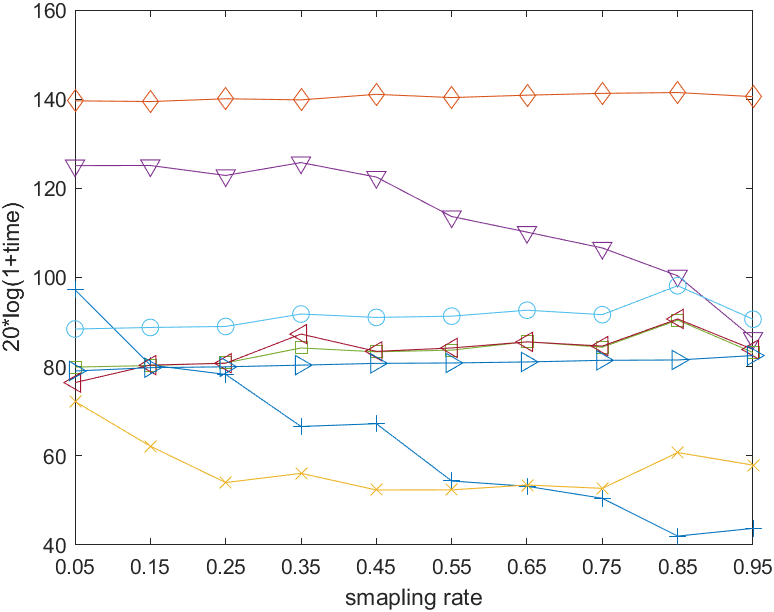}
    \end{minipage}
  }
      \caption{We plot the trend of psnr and time with increasing sampling rate for different algorithms in the form of a line chart. Where we use 20*log(1+time) to mark the 
      vertical coordinates on the line chart of the algorithm's running time.}
  \label{fig_6}
\end{figure*}

\subsection{Video Recovery}
The test video\footnote{The data is available at http://openremotesensing.net/kb/data/.} dataset consisted of four color videos of size 144$\times$176$\times$3$\times$50 (spatial height $\times$ spatial width $\times$ color channel $\times$ frame). 
For each data, we test four miss rate(MR): 95\%, 90\%, 85\%, 75\% and employ PSNR, SSIM and running time as the quantitative metric. 
The maximum number of iterations and the minimum error for all algorithms in the experiments we set to 300 and $10^{-6}$ respectively. 
Among them, the tensor rank of the TR-WOPT and TRLRF algorithms we choose are all 10, and the tensor rank of our proposed algorithms are $\{\mathbf{R}_{i,j}\}_{i\neq j} = 5$, $\mathbf{R}_{1,1} = \mathbf{R}_{2,2} = 70$, $\mathbf{R}_{3,3} = 3$ and $\mathbf{R}_{4,4} = 10$. 
Especially, our proposed LMTN-AR algorithm has the maximum rank of $\{\mathbf{\hat{R}}_{i,j}\}_{i\neq j} = 6$, $\mathbf{\hat{R}}_{1,1} = \mathbf{\hat{R}}_{2,2} = 80$, $\mathbf{\hat{R}}_{3,3} = 3$ and $\mathbf{\hat{R}}_{4,4} = 20$. 
After several experimental tests, we recommend in these videos of size 144 × 176 × 3 × 50 that the rank of the tensor $\{\mathbf{R}_{i,j}\}_{i\neq j}$, 
$\mathbf{R}_{1,1} = \mathbf{R}_{2,2}$, $\mathbf{R}_{3,3}$ and $\mathbf{R}_{4,4}$ can be chosen in the sets \{4, 5, 6\}, \{60, 70, 80\}, \{2, 3\} and \{10, 15\} respectively. 
For the choice of $\{\mathbf{\hat{R}}_{i,j}\}_{i\neq j}$, $\mathbf{\hat{R}}_{1,1} = \mathbf{\hat{R}}_{2,2}$, $\mathbf{\hat{R}}_{3,3}$ and $\mathbf{\hat{R}}_{4,4}$ can be chosen from the sets \{5, 6, 7\}, \{70, 80, 90\}, \{3\}, \{15, 20\} respectively. 
These choices of tensor rank are not considered being the best, but we can obtain some valuable results in a limited time. 

Table \uppercase\expandafter{\romannumeral4} show LMTN algorithms are faster than FCTN-PAM algorithm, and LMTN-SVD algorithm is 3-4 times faster than FCTN-PAM with only a 1.8 points accuracy drop. 
We can also see that although the HaLRTC algorithm is fast, the accuracy of tensor completion is the worst one usually. 
The running time of our algorithm is the fastest on the completion based on the tensor decomposition method. 
The reason is that the dimensionality reduction of the latent matrices, where we can effectively reduce the amount of data for the factors and speed up our algorithm.

\begin{table*}[]
  \centering
  \caption{Performance of each algorithm for various datasets with different missing rates}
  \resizebox{1.95\columnwidth}{!}{
  \begin{tabular}{c|c|ccc|ccc|ccc|ccc}
    \hline
    \hline
                              & Missing Rate & \multicolumn{3}{c|}{95\%}                                                               & \multicolumn{3}{c|}{90\%}                                                               & \multicolumn{3}{c|}{85\%}                                                               & \multicolumn{3}{c}{75\%}                                                               \\ \hline
    Dataset                   & Methods      & \multicolumn{1}{c|}{PSNR}   & \multicolumn{1}{c|}{SSIM}   & TIME                        & \multicolumn{1}{c|}{PSNR}   & \multicolumn{1}{c|}{SSIM}   & TIME                        & \multicolumn{1}{c|}{PSNR}   & \multicolumn{1}{c|}{SSIM}   & TIME                        & \multicolumn{1}{c|}{PSNR}   & \multicolumn{1}{c|}{SSIM}   & TIME                       \\ \hline
    \multirow{8}{*}{container} & TMac         & \multicolumn{1}{c|}{23.879} & \multicolumn{1}{c|}{0.7808} & 258.62                      & \multicolumn{1}{c|}{24.929} & \multicolumn{1}{c|}{0.8058} & 226.33                      & \multicolumn{1}{c|}{25.460} & \multicolumn{1}{c|}{0.8218} & 244.46                      & \multicolumn{1}{c|}{26.196} & \multicolumn{1}{c|}{0.8443} & 191.716                     \\ \cline{2-14} 
                              & TR-WOPT      & \multicolumn{1}{c|}{27.361} & \multicolumn{1}{c|}{0.8324} & 2387.0                      & \multicolumn{1}{c|}{29.694} & \multicolumn{1}{c|}{0.8742} & 2277.5                      & \multicolumn{1}{c|}{30.751} & \multicolumn{1}{c|}{0.8951} & 2033.2                      & \multicolumn{1}{c|}{30.343} & \multicolumn{1}{c|}{0.8884} & 2136.6                     \\ \cline{2-14} 
                              & HaLRTC       & \multicolumn{1}{c|}{18.570} & \multicolumn{1}{c|}{0.6099} & $\mathbf{68.764}$                      & \multicolumn{1}{c|}{21.571} & \multicolumn{1}{c|}{0.7249} & $\mathbf{49.568}$                      & \multicolumn{1}{c|}{23.560} & \multicolumn{1}{c|}{0.7958} & $\mathbf{40.350}$                      & \multicolumn{1}{c|}{26.643} & \multicolumn{1}{c|}{0.8772} & $\mathbf{40.015}$                     \\ \cline{2-14} 
                              & TRLRF        & \multicolumn{1}{c|}{27.816} & \multicolumn{1}{c|}{0.8571} & 1108.3                      & \multicolumn{1}{c|}{29.977} & \multicolumn{1}{c|}{0.8962} & 1213.4                      & \multicolumn{1}{c|}{30.831} & \multicolumn{1}{c|}{0.9066} & 1157.0                      & \multicolumn{1}{c|}{31.803} & \multicolumn{1}{c|}{0.9245} & 1137.6                     \\ \cline{2-14} 
                              & FCTN-PAM     & \multicolumn{1}{c|}{20.456} & \multicolumn{1}{c|}{0.6904} & \multicolumn{1}{c|}{653.65} & \multicolumn{1}{c|}{\underline{33.163}} & \multicolumn{1}{c|}{\underline{0.9251}} & \multicolumn{1}{c|}{609.65} & \multicolumn{1}{c|}{$\mathbf{34.857}$} & \multicolumn{1}{c|}{$\mathbf{0.9401}$} & \multicolumn{1}{c|}{608.56} & \multicolumn{1}{c|}{$\mathbf{35.779}$} & \multicolumn{1}{c|}{$\mathbf{0.9528}$} & \multicolumn{1}{c}{630.13} \\ \cline{2-14} 
                              & LMTN-SVD     & \multicolumn{1}{c|}{\underline{27.918}} & \multicolumn{1}{c|}{\underline{0.8481}} & 141.15                      & \multicolumn{1}{c|}{33.091} & \multicolumn{1}{c|}{0.9237} & 151.97                      & \multicolumn{1}{c|}{34.032} & \multicolumn{1}{c|}{0.9328} & 170.96                      & \multicolumn{1}{c|}{35.123} & \multicolumn{1}{c|}{0.9438} & 176.10                     \\ \cline{2-14} 
                              & LMTN-PAM     & \multicolumn{1}{c|}{$\mathbf{28.633}$} & \multicolumn{1}{c|}{$\mathbf{0.8618}$} & 519.29                      & \multicolumn{1}{c|}{$\mathbf{33.314}$} & \multicolumn{1}{c|}{$\mathbf{0.9270}$} & 519.16                      & \multicolumn{1}{c|}{\underline{34.134}} & \multicolumn{1}{c|}{\underline{0.9347}} & 536.44                      & \multicolumn{1}{c|}{\underline{35.202}} & \multicolumn{1}{c|}{\underline{0.9453}} & 541.32                     \\ \cline{2-14} 
                              & LMTN-AR      & \multicolumn{1}{c|}{26.545} & \multicolumn{1}{c|}{0.8267} & \underline{119.02}                      & \multicolumn{1}{c|}{30.447} & \multicolumn{1}{c|}{0.9033} & \underline{138.02}                      & \multicolumn{1}{c|}{31.064} & \multicolumn{1}{c|}{0.9069} & \underline{126.06}                      & \multicolumn{1}{c|}{32.155} & \multicolumn{1}{c|}{0.9141} & \underline{163.63}                     \\ \hline
    \multirow{8}{*}{calphone} & TMac         & \multicolumn{1}{c|}{24.928} & \multicolumn{1}{c|}{0.7193} & 279.66                      & \multicolumn{1}{c|}{25.514} & \multicolumn{1}{c|}{0.7456} & 215.08                      & \multicolumn{1}{c|}{25.943} & \multicolumn{1}{c|}{0.7655} & 140.29                      & \multicolumn{1}{c|}{26.686} & \multicolumn{1}{c|}{0.7964} & 165.65                     \\ \cline{2-14} 
                              & TR-WOPT      & \multicolumn{1}{c|}{27.674} & \multicolumn{1}{c|}{0.7757} & 2113.8                      & \multicolumn{1}{c|}{29.064} & \multicolumn{1}{c|}{0.8297} & 2179.4                      & \multicolumn{1}{c|}{28.923} & \multicolumn{1}{c|}{0.8294} & 2317.8                      & \multicolumn{1}{c|}{29.364} & \multicolumn{1}{c|}{0.8429} & 2345.2                     \\ \cline{2-14} 
                              & HaLRTC       & \multicolumn{1}{c|}{17.813} & \multicolumn{1}{c|}{0.5303} & $\mathbf{70.762}$                      & \multicolumn{1}{c|}{21.620} & \multicolumn{1}{c|}{0.6604} & $\mathbf{44.372}$                      & \multicolumn{1}{c|}{23.900} & \multicolumn{1}{c|}{0.7449} & $\mathbf{36.502}$                      & \multicolumn{1}{c|}{27.183} & \multicolumn{1}{c|}{0.8476} & $\mathbf{36.131}$                     \\ \cline{2-14} 
                              & TRLRF        & \multicolumn{1}{c|}{\underline{27.711}} & \multicolumn{1}{c|}{\underline{0.8012}} & 1143.5                      & \multicolumn{1}{c|}{28.546} & \multicolumn{1}{c|}{0.8324} & 1215.4                      & \multicolumn{1}{c|}{29.653} & \multicolumn{1}{c|}{0.8612} & 1263.2                      & \multicolumn{1}{c|}{30.920} & \multicolumn{1}{c|}{0.8887} & 1218.0                     \\ \cline{2-14} 
                              & FCTN-PAM     & \multicolumn{1}{c|}{21.985} & \multicolumn{1}{c|}{0.5874} & \multicolumn{1}{c|}{681.56} & \multicolumn{1}{c|}{29.521} & \multicolumn{1}{c|}{0.8678} & \multicolumn{1}{c|}{633.75} & \multicolumn{1}{c|}{$\mathbf{31.210}$} & \multicolumn{1}{c|}{$\mathbf{0.8953}$} & \multicolumn{1}{c|}{662.25} & \multicolumn{1}{c|}{$\mathbf{32.064}$} & \multicolumn{1}{c|}{$\mathbf{0.9141}$} & \multicolumn{1}{c}{711.65} \\ \cline{2-14} 
                              & LMTN-SVD     & \multicolumn{1}{c|}{27.427} & \multicolumn{1}{c|}{0.7808} & 152.18                      & \multicolumn{1}{c|}{\underline{30.018}} & \multicolumn{1}{c|}{\underline{0.8726}} & 158.36                      & \multicolumn{1}{c|}{30.772} & \multicolumn{1}{c|}{0.8928} & 159.79                      & \multicolumn{1}{c|}{31.595} & \multicolumn{1}{c|}{0.9106} & 180.69                     \\ \cline{2-14} 
                              & LMTN-PAM     & \multicolumn{1}{c|}{$\mathbf{27.875}$} & \multicolumn{1}{c|}{$\mathbf{0.8031}$} & 518.55                      & \multicolumn{1}{c|}{$\mathbf{30.111}$} & \multicolumn{1}{c|}{$\mathbf{0.8775}$} & 529.26                      & \multicolumn{1}{c|}{\underline{30.773}} & \multicolumn{1}{c|}{\underline{0.8930}} & 519.51                      & \multicolumn{1}{c|}{\underline{31.621}} & \multicolumn{1}{c|}{\underline{0.9134}} & 540.11                     \\ \cline{2-14} 
                              & LMTN-AR      & \multicolumn{1}{c|}{15.703} & \multicolumn{1}{c|}{0.3660} & \underline{135.80}                      & \multicolumn{1}{c|}{29.085} & \multicolumn{1}{c|}{0.8553} & \underline{136.22}                      & \multicolumn{1}{c|}{29.471} & \multicolumn{1}{c|}{0.8670} & \underline{137.87}                      & \multicolumn{1}{c|}{30.608} & \multicolumn{1}{c|}{0.8991} & \underline{150.26}                     \\ \hline
    \multirow{8}{*}{news}     & TMac         & \multicolumn{1}{c|}{23.446} & \multicolumn{1}{c|}{0.6977} & 255.01                      & \multicolumn{1}{c|}{24.263} & \multicolumn{1}{c|}{0.7308} & 222.52                      & \multicolumn{1}{c|}{24.644} & \multicolumn{1}{c|}{0.7509} & 195.63                      & \multicolumn{1}{c|}{25.459} & \multicolumn{1}{c|}{0.7834} & 99.998                     \\ \cline{2-14} 
                              & TR-WOPT      & \multicolumn{1}{c|}{27.267} & \multicolumn{1}{c|}{0.7961} & 1587.2                      & \multicolumn{1}{c|}{30.230} & \multicolumn{1}{c|}{0.8814} & 2220.3                      & \multicolumn{1}{c|}{29.933} & \multicolumn{1}{c|}{0.8767} & 1631.9                      & \multicolumn{1}{c|}{29.568} & \multicolumn{1}{c|}{0.8706} & 1407.8                     \\ \cline{2-14} 
                              & HaLRTC       & \multicolumn{1}{c|}{16.214} & \multicolumn{1}{c|}{0.5224} & $\mathbf{85.261}$                      & \multicolumn{1}{c|}{20.178} & \multicolumn{1}{c|}{0.6685} & $\mathbf{54.067}$                      & \multicolumn{1}{c|}{22.513} & \multicolumn{1}{c|}{0.7553} & $\mathbf{53.004}$                      & \multicolumn{1}{c|}{25.982} & \multicolumn{1}{c|}{0.8586} & $\mathbf{45.549}$                     \\ \cline{2-14} 
                              & TRLRF        & \multicolumn{1}{c|}{26.925} & \multicolumn{1}{c|}{0.7972} & 905.74                      & \multicolumn{1}{c|}{29.106} & \multicolumn{1}{c|}{0.8619} & 1249.2                      & \multicolumn{1}{c|}{30.585} & \multicolumn{1}{c|}{0.8949} & 883.66                      & \multicolumn{1}{c|}{31.834} & \multicolumn{1}{c|}{0.9168} & 817.24                     \\ \cline{2-14} 
                              & FCTN-PAM     & \multicolumn{1}{c|}{23.543} & \multicolumn{1}{c|}{0.6709} & \multicolumn{1}{c|}{650.95} & \multicolumn{1}{c|}{\underline{31.972}} & \multicolumn{1}{c|}{\underline{0.9158}} & \multicolumn{1}{c|}{739.85} & \multicolumn{1}{c|}{$\mathbf{33.041}$} & \multicolumn{1}{c|}{$\mathbf{0.9353}$} & \multicolumn{1}{c|}{753.84} & \multicolumn{1}{c|}{$\mathbf{34.129}$} & \multicolumn{1}{c|}{$\mathbf{0.9519}$} & \multicolumn{1}{c}{627.35} \\ \cline{2-14} 
                              & LMTN-SVD     & \multicolumn{1}{c|}{27.259} & \multicolumn{1}{c|}{0.8060} & 200.75                      & \multicolumn{1}{c|}{31.680} & \multicolumn{1}{c|}{0.9109} & 183.52                      & \multicolumn{1}{c|}{32.314} & \multicolumn{1}{c|}{0.9235} & 203.17                      & \multicolumn{1}{c|}{33.248} & \multicolumn{1}{c|}{0.9376} & 232.97                     \\ \cline{2-14} 
                              & LMTN-PAM     & \multicolumn{1}{c|}{\underline{27.842}} & \multicolumn{1}{c|}{\underline{0.8269}} & 522.59                      & \multicolumn{1}{c|}{$\mathbf{31.973}$} & \multicolumn{1}{c|}{$\mathbf{0.9169}$} & 508.04                      & \multicolumn{1}{c|}{\underline{32.518}} & \multicolumn{1}{c|}{\underline{0.9268}} & 540.03                      & \multicolumn{1}{c|}{\underline{33.499}} & \multicolumn{1}{c|}{\underline{0.9403}} & 519.03                     \\ \cline{2-14} 
                              & LMTN-AR      & \multicolumn{1}{c|}{$\mathbf{27.958}$} & \multicolumn{1}{c|}{$\mathbf{0.8489}$} & \underline{158.23}                      & \multicolumn{1}{c|}{30.723} & \multicolumn{1}{c|}{0.9099} & \underline{172.79}                      & \multicolumn{1}{c|}{30.799} & \multicolumn{1}{c|}{0.9158} & \underline{191.77}                      & \multicolumn{1}{c|}{31.604} & \multicolumn{1}{c|}{0.9275} & 141.68                     \\ \hline
    \multirow{8}{*}{mobile}   & TMac         & \multicolumn{1}{c|}{16.780} & \multicolumn{1}{c|}{0.3355} & 220.37                      & \multicolumn{1}{c|}{17.428} & \multicolumn{1}{c|}{0.4023} & 174.88                      & \multicolumn{1}{c|}{17.887} & \multicolumn{1}{c|}{0.4512} & \underline{118.83}                      & \multicolumn{1}{c|}{18.722} & \multicolumn{1}{c|}{0.5327} & \underline{74.163}                     \\ \cline{2-14} 
                              & TR-WOPT      & \multicolumn{1}{c|}{17.421} & \multicolumn{1}{c|}{0.4001} & 2199.7                      & \multicolumn{1}{c|}{18.510} & \multicolumn{1}{c|}{0.4716} & 2244.1                      & \multicolumn{1}{c|}{18.781} & \multicolumn{1}{c|}{0.4932} & 2258.0                      & \multicolumn{1}{c|}{18.842} & \multicolumn{1}{c|}{0.4976} & 2218.0                     \\ \cline{2-14} 
                              & HaLRTC       & \multicolumn{1}{c|}{12.405} & \multicolumn{1}{c|}{0.1703} & $\mathbf{85.345}$                      & \multicolumn{1}{c|}{14.385} & \multicolumn{1}{c|}{0.2664} & $\mathbf{50.564}$                      & \multicolumn{1}{c|}{15.757} & \multicolumn{1}{c|}{0.3632} & $\mathbf{37.892}$                      & \multicolumn{1}{c|}{17.991} & \multicolumn{1}{c|}{0.5343} & $\mathbf{24.898}$                     \\ \cline{2-14} 
                              & TRLRF        & \multicolumn{1}{c|}{17.598} & \multicolumn{1}{c|}{0.4282} & 1414.9                      & \multicolumn{1}{c|}{18.662} & \multicolumn{1}{c|}{0.5073} & 1261.8                      & \multicolumn{1}{c|}{19.433} & \multicolumn{1}{c|}{0.5698} & 1257.0                      & \multicolumn{1}{c|}{20.365} & \multicolumn{1}{c|}{0.6421} & 1272.3                     \\ \cline{2-14} 
                              & FCTN-PAM     & \multicolumn{1}{c|}{16.222} & \multicolumn{1}{c|}{0.3173} & \multicolumn{1}{c|}{617.47} & \multicolumn{1}{c|}{\underline{19.414}} & \multicolumn{1}{c|}{$\mathbf{0.5606}$} & \multicolumn{1}{c|}{632.86} & \multicolumn{1}{c|}{$\mathbf{20.191}$} & \multicolumn{1}{c|}{$\mathbf{0.6109}$} & \multicolumn{1}{c|}{647.73} & \multicolumn{1}{c|}{$\mathbf{21.016}$} & \multicolumn{1}{c|}{$\mathbf{0.6769}$} & \multicolumn{1}{c}{654.23} \\ \cline{2-14} 
                              & LMTN-SVD     & \multicolumn{1}{c|}{$\mathbf{17.895}$} & \multicolumn{1}{c|}{$\mathbf{0.4367}$} & 179.23                      & \multicolumn{1}{c|}{$\mathbf{19.456}$} & \multicolumn{1}{c|}{\underline{0.5576}} & 185.77                      & \multicolumn{1}{c|}{\underline{20.069}} & \multicolumn{1}{c|}{\underline{0.6079}} & 185.47                      & \multicolumn{1}{c|}{\underline{20.881}} & \multicolumn{1}{c|}{\underline{0.6702}} & 188.72                     \\ \cline{2-14} 
                              & LMTN-PAM     & \multicolumn{1}{c|}{\underline{17.780}} & \multicolumn{1}{c|}{\underline{0.4356}} & 512.03                      & \multicolumn{1}{c|}{19.390} & \multicolumn{1}{c|}{0.5562} & 560.95                      & \multicolumn{1}{c|}{19.997} & \multicolumn{1}{c|}{0.6051} & 560.67                      & \multicolumn{1}{c|}{20.803} & \multicolumn{1}{c|}{0.6682} & 573.62                     \\ \cline{2-14} 
                              & LMTN-AR      & \multicolumn{1}{c|}{16.482} & \multicolumn{1}{c|}{0.4078} & \underline{164.38}                      & \multicolumn{1}{c|}{17.637} & \multicolumn{1}{c|}{0.5032} & \underline{169.05}                      & \multicolumn{1}{c|}{18.252} & \multicolumn{1}{c|}{0.5550} & 168.97                      & \multicolumn{1}{c|}{18.989} & \multicolumn{1}{c|}{0.6184} & 173.82                     \\ \hline \hline
    \end{tabular}}
\end{table*}

Further, we show the results of our video-completion experiments in Fig. 7, showing the first frame and the residual images of the container, news and bridge videos after video-completion. 
All the algorithm settings are the same as the settings on beginning of this subsection , and the video missing rate is at 85\%. 
In addition, we also tested a video-completion experiment using a bus video (144$\times$176$\times$3$\times$50) with a missing rate at 60\%. 
The experimental video has a large range of motion, but there is still a lot of redundant information in each frame of the video, which we 
also tried to complete. All settings of the algorithm are the same as in the video experiments above, and the results can be seen in Fig. 8. 
We show the images at frames 10, 20, 30 and 40 respectively in Fig. 8 after the video has been completed. 
\begin{figure*}
  \centering
        \subfigure{
        \begin{minipage}{0.085\linewidth}
          \centering{\footnotesize Observed} 
        \includegraphics[height = 1.5cm,width=1.7cm]{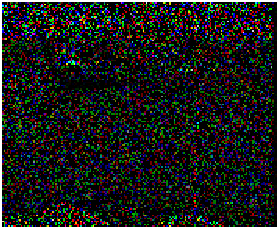}
        \end{minipage}
        }
        \subfigure{
        \begin{minipage}{0.085\linewidth}
          \centering{\footnotesize TMac}
        \includegraphics[height = 1.5cm,width=1.7cm]{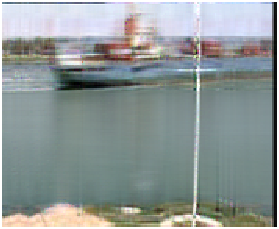}
        \end{minipage}    
        }
        \subfigure{
        \begin{minipage}{0.085\linewidth}
          \centering{\footnotesize TR-WOPT}
        \includegraphics[height = 1.5cm,width=1.7cm]{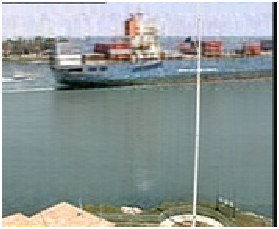}
        \end{minipage}    
        }
        \subfigure{
        \begin{minipage}{0.085\linewidth}
          \centering{\footnotesize HaLRTC}
        \includegraphics[height = 1.5cm,width=1.7cm]{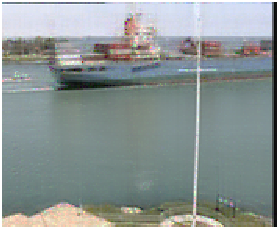}
        \end{minipage}    
        }
        \subfigure{
        \begin{minipage}{0.085\linewidth}
          \centering{\footnotesize TRLRF}
        \includegraphics[height = 1.5cm,width=1.7cm]{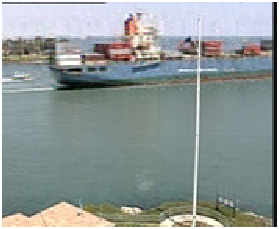}
        \end{minipage}    
        }
        \subfigure{
        \begin{minipage}{0.085\linewidth}
          \centering{\footnotesize FCTN-PAM}
        \includegraphics[height = 1.5cm,width=1.7cm]{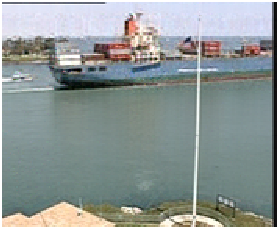}
        \end{minipage}    
        }
        \subfigure{
        \begin{minipage}{0.085\linewidth}
          \centering{\footnotesize LMTN-AR}
        \includegraphics[height = 1.5cm,width=1.7cm]{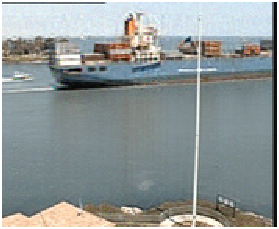}
        \end{minipage}    
        }
        \subfigure{
        \begin{minipage}{0.085\linewidth}
          \centering{\footnotesize LMTN-PAM}
        \includegraphics[height = 1.5cm,width=1.7cm]{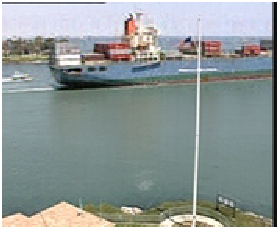}
        \end{minipage}    
        }
        \subfigure{
        \begin{minipage}{0.085\linewidth}
          \centering{\footnotesize LMTN-SVD}
        \includegraphics[height = 1.5cm,width=1.7cm]{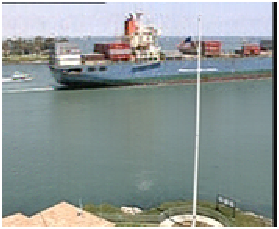}
        \end{minipage}    
        }
        \subfigure{
        \begin{minipage}{0.085\linewidth}
          \centering{\footnotesize Ground truth}
        \includegraphics[height = 1.5cm,width=1.7cm]{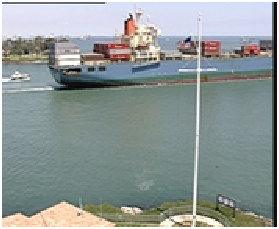}
        \end{minipage}    
        }

        \subfigure{
        \begin{minipage}{0.085\linewidth}
        \includegraphics[height = 1.5cm,width=1.7cm]{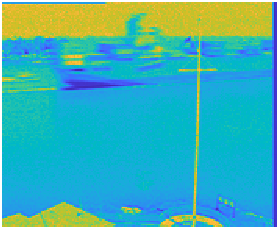}
        \end{minipage}
        }
        \subfigure{
        \begin{minipage}{0.085\linewidth}
        \includegraphics[height = 1.5cm,width=1.7cm]{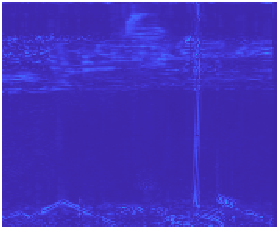}
        \end{minipage}    
        }
        \subfigure{
        \begin{minipage}{0.085\linewidth}
        \includegraphics[height = 1.5cm,width=1.7cm]{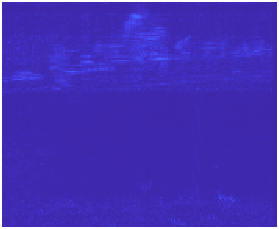}
        \end{minipage}    
        }
        \subfigure{
        \begin{minipage}{0.085\linewidth}
        \includegraphics[height = 1.5cm,width=1.7cm]{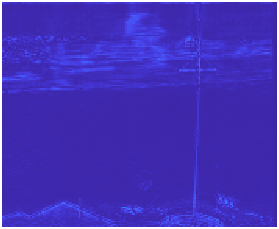}
        \end{minipage}    
        }
        \subfigure{
        \begin{minipage}{0.085\linewidth}
        \includegraphics[height = 1.5cm,width=1.7cm]{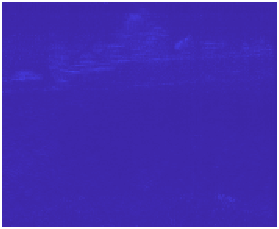}
        \end{minipage}    
        }
        \subfigure{
        \begin{minipage}{0.085\linewidth}
        \includegraphics[height = 1.5cm,width=1.7cm]{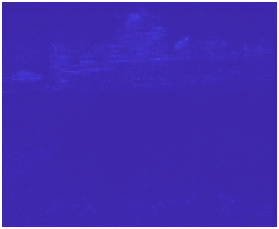}
        \end{minipage}    
        }
        \subfigure{
        \begin{minipage}{0.085\linewidth}
        \includegraphics[height = 1.5cm,width=1.7cm]{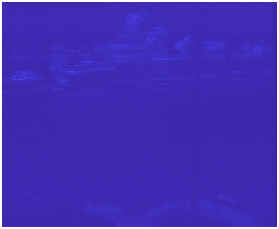}
        \end{minipage}    
        }
        \subfigure{
        \begin{minipage}{0.085\linewidth}
        \includegraphics[height = 1.5cm,width=1.7cm]{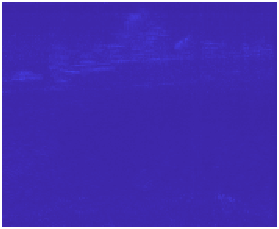}
        \end{minipage}    
        }
        \subfigure{
        \begin{minipage}{0.085\linewidth}
        \includegraphics[height = 1.5cm,width=1.7cm]{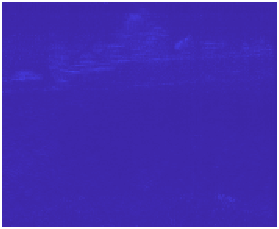}
        \end{minipage}    
        }
        \subfigure{
        \begin{minipage}{0.085\linewidth}
        \includegraphics[height = 1.5cm,width=1.7cm]{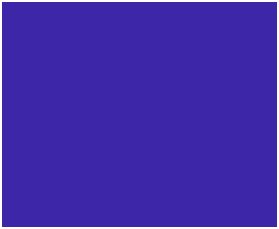}
        \end{minipage}    
        }

        \subfigure{
        \begin{minipage}{0.085\linewidth}
        \includegraphics[height = 1.5cm,width=1.7cm]{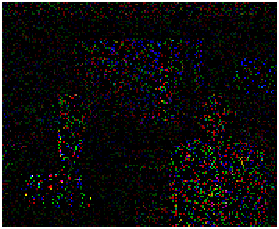}
        \end{minipage}
        }
        \subfigure{
        \begin{minipage}{0.085\linewidth}
        \includegraphics[height = 1.5cm,width=1.7cm]{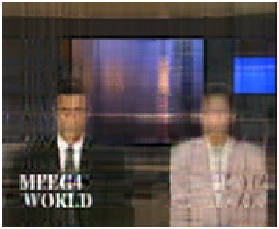}
        \end{minipage}    
        }
        \subfigure{
        \begin{minipage}{0.085\linewidth}
        \includegraphics[height = 1.5cm,width=1.7cm]{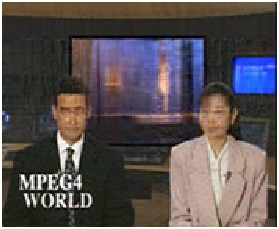}
        \end{minipage}    
        }
        \subfigure{
        \begin{minipage}{0.085\linewidth}
        \includegraphics[height = 1.5cm,width=1.7cm]{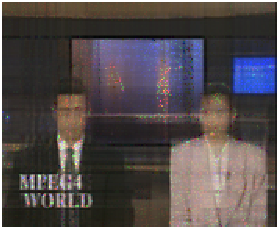}
        \end{minipage}    
        }
        \subfigure{
        \begin{minipage}{0.085\linewidth}
        \includegraphics[height = 1.5cm,width=1.7cm]{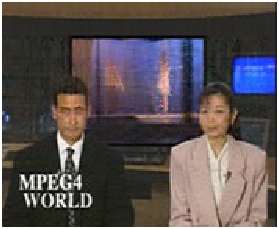}
        \end{minipage}    
        }
        \subfigure{
        \begin{minipage}{0.085\linewidth}
        \includegraphics[height = 1.5cm,width=1.7cm]{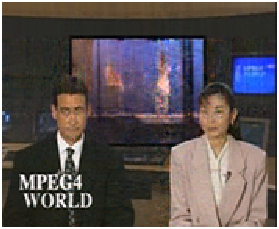}
        \end{minipage}    
        }
        \subfigure{
        \begin{minipage}{0.085\linewidth}
        \includegraphics[height = 1.5cm,width=1.7cm]{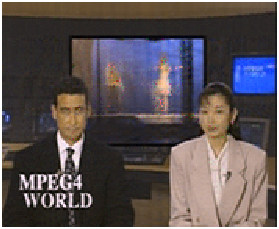}
        \end{minipage}    
        }
        \subfigure{
        \begin{minipage}{0.085\linewidth}
        \includegraphics[height = 1.5cm,width=1.7cm]{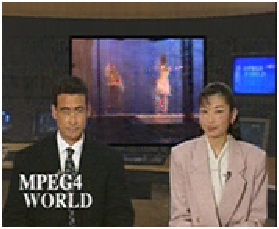}
        \end{minipage}    
        }
        \subfigure{
        \begin{minipage}{0.085\linewidth}
        \includegraphics[height = 1.5cm,width=1.7cm]{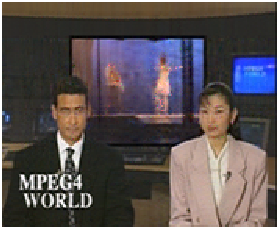}
        \end{minipage}    
        }
        \subfigure{
        \begin{minipage}{0.085\linewidth}
        \includegraphics[height = 1.5cm,width=1.7cm]{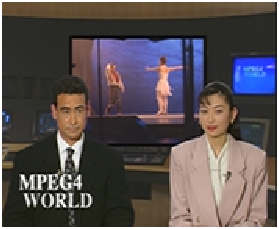}
        \end{minipage}    
        }

        \subfigure{
        \begin{minipage}{0.085\linewidth}
        \includegraphics[height = 1.5cm,width=1.7cm]{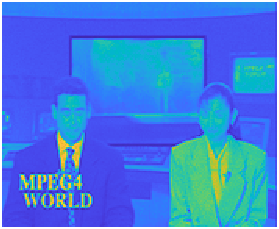}
        \end{minipage}
        }
        \subfigure{
        \begin{minipage}{0.085\linewidth}
        \includegraphics[height = 1.5cm,width=1.7cm]{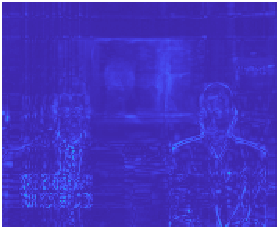}
        \end{minipage}    
        }
        \subfigure{
        \begin{minipage}{0.085\linewidth}
        \includegraphics[height = 1.5cm,width=1.7cm]{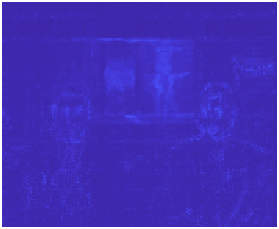}
        \end{minipage}    
        }
        \subfigure{
        \begin{minipage}{0.085\linewidth}
        \includegraphics[height = 1.5cm,width=1.7cm]{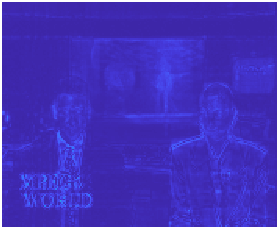}
        \end{minipage}    
        }
        \subfigure{
        \begin{minipage}{0.085\linewidth}
        \includegraphics[height = 1.5cm,width=1.7cm]{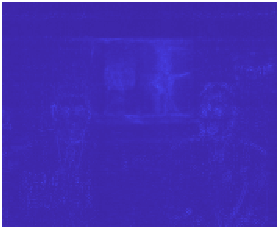}
        \end{minipage}    
        }
        \subfigure{
        \begin{minipage}{0.085\linewidth}
        \includegraphics[height = 1.5cm,width=1.7cm]{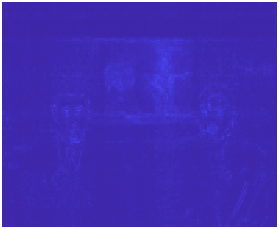}
        \end{minipage}    
        }
        \subfigure{
        \begin{minipage}{0.085\linewidth}
        \includegraphics[height = 1.5cm,width=1.7cm]{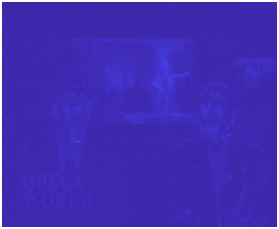}
        \end{minipage}    
        }
        \subfigure{
        \begin{minipage}{0.085\linewidth}
        \includegraphics[height = 1.5cm,width=1.7cm]{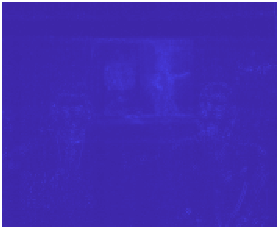}
        \end{minipage}    
        }
        \subfigure{
        \begin{minipage}{0.085\linewidth}
        \includegraphics[height = 1.5cm,width=1.7cm]{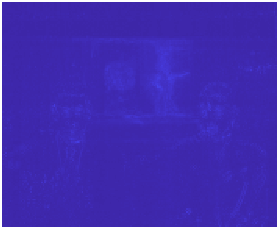}
        \end{minipage}    
        }
        \subfigure{
        \begin{minipage}{0.085\linewidth}
        \includegraphics[height = 1.5cm,width=1.7cm]{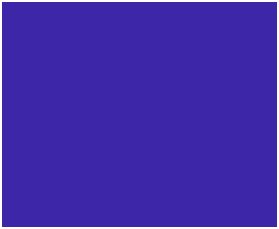}
        \end{minipage}    
        }

        \subfigure{
        \begin{minipage}{0.085\linewidth}
        \includegraphics[height = 1.5cm,width=1.7cm]{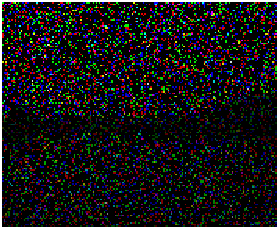}
        \end{minipage}
        }
        \subfigure{
        \begin{minipage}{0.085\linewidth}
        \includegraphics[height = 1.5cm,width=1.7cm]{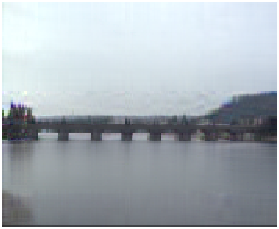}
        \end{minipage}    
        }
        \subfigure{
        \begin{minipage}{0.085\linewidth}
        \includegraphics[height = 1.5cm,width=1.7cm]{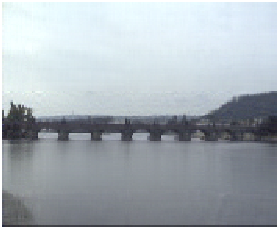}
        \end{minipage}    
        }
        \subfigure{
        \begin{minipage}{0.085\linewidth}
        \includegraphics[height = 1.5cm,width=1.7cm]{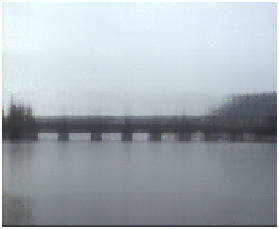}
        \end{minipage}    
        }
        \subfigure{
        \begin{minipage}{0.085\linewidth}
        \includegraphics[height = 1.5cm,width=1.7cm]{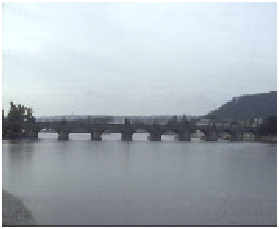}
        \end{minipage}    
        }
        \subfigure{
        \begin{minipage}{0.085\linewidth}
        \includegraphics[height = 1.5cm,width=1.7cm]{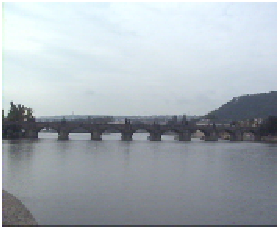}
        \end{minipage}    
        }
        \subfigure{
        \begin{minipage}{0.085\linewidth}
        \includegraphics[height = 1.5cm,width=1.7cm]{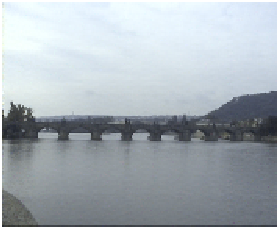}
        \end{minipage}    
        }
        \subfigure{
        \begin{minipage}{0.085\linewidth}
        \includegraphics[height = 1.5cm,width=1.7cm]{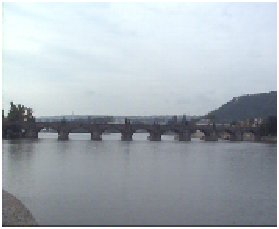}
        \end{minipage}    
        }
        \subfigure{
        \begin{minipage}{0.085\linewidth}
        \includegraphics[height = 1.5cm,width=1.7cm]{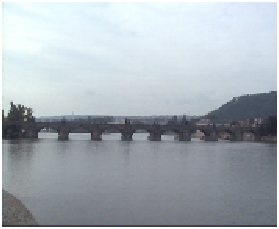}
        \end{minipage}    
        }
        \subfigure{
        \begin{minipage}{0.085\linewidth}
        \includegraphics[height = 1.5cm,width=1.7cm]{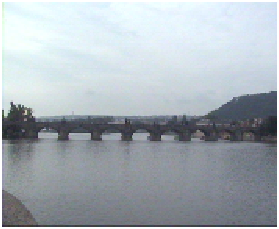}
        \end{minipage}    
        }

        \setcounter{subfigure}{0}
        \subfigure{
        \begin{minipage}{0.085\linewidth}
        \includegraphics[height = 1.5cm,width=1.7cm]{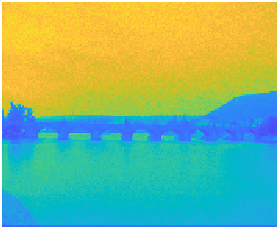}
        \end{minipage}
        }
        \subfigure{
        \begin{minipage}{0.085\linewidth}
        \includegraphics[height = 1.5cm,width=1.7cm]{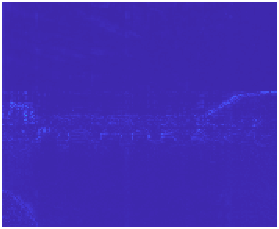}
        \end{minipage}    
        }
        \subfigure{
        \begin{minipage}{0.085\linewidth}
        \includegraphics[height = 1.5cm,width=1.7cm]{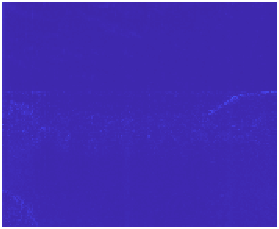}
        \end{minipage}    
        }
        \subfigure{
        \begin{minipage}{0.085\linewidth}
        \includegraphics[height = 1.5cm,width=1.7cm]{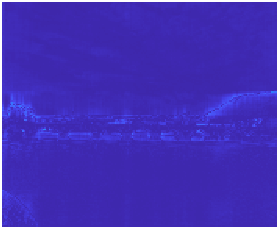}
        \end{minipage}    
        }
        \subfigure{
        \begin{minipage}{0.085\linewidth}
        \includegraphics[height = 1.5cm,width=1.7cm]{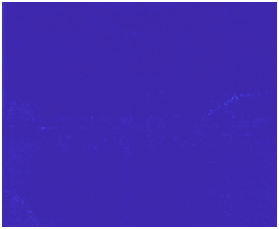}
        \end{minipage}    
        }
        \subfigure{
        \begin{minipage}{0.085\linewidth}
        \includegraphics[height = 1.5cm,width=1.7cm]{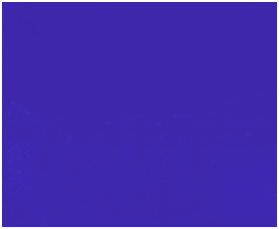}
        \end{minipage}    
        }
        \subfigure{
        \begin{minipage}{0.085\linewidth}
        \includegraphics[height = 1.5cm,width=1.7cm]{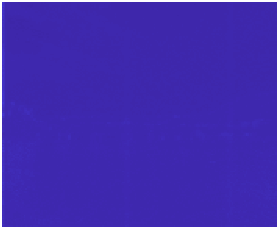}
        \end{minipage}    
        }
        \subfigure{
        \begin{minipage}{0.085\linewidth}
        \includegraphics[height = 1.5cm,width=1.7cm]{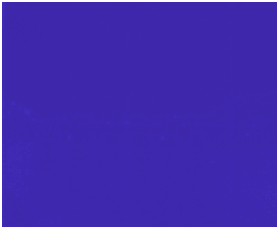}
        \end{minipage}    
        }
        \subfigure{
        \begin{minipage}{0.085\linewidth}
        \includegraphics[height = 1.5cm,width=1.7cm]{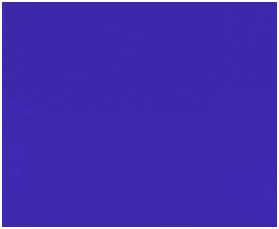}
        \end{minipage}    
        }
        \subfigure{
        \begin{minipage}{0.085\linewidth}
        \includegraphics[height = 1.5cm,width=1.7cm]{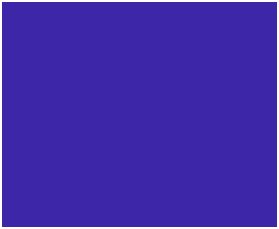}
        \end{minipage}    
        }

        \subfigure{
        \begin{minipage}{1\linewidth}
        \includegraphics[width=18cm]{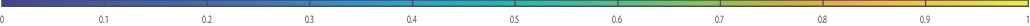}
        \end{minipage}    
        }
      \caption{Reconstructed results on three testing videos with 85\% missing rate. The odd-numbered rows are the visual results of the first frame of video container and video news, respectively; 
      the even-numbered rows are the residual images of the corresponding average values on their three colour channels. Where the colour closer to yellow indicates a larger residual value and 
      the colour closer to blue indicates a smaller residual value.}
  \label{fig_7}
\end{figure*}

\begin{figure*}
  \centering
        \subfigure{
        \begin{minipage}{0.085\linewidth}
          \centering{\footnotesize Observed}
        \includegraphics[height = 1.5cm,width=1.7cm]{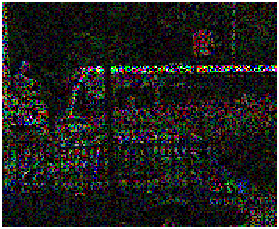}
        \end{minipage}
        }
        \subfigure{
        \begin{minipage}{0.085\linewidth}
          \centering{\footnotesize TMac}
        \includegraphics[height = 1.5cm,width=1.7cm]{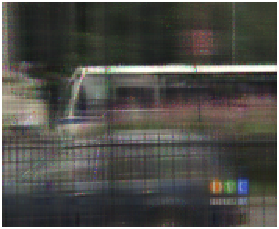}
        \end{minipage}    
        }
        \subfigure{
        \begin{minipage}{0.085\linewidth}
          \centering{\footnotesize TR-WOPT}
        \includegraphics[height = 1.5cm,width=1.7cm]{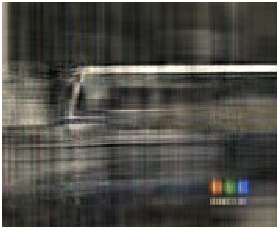}
        \end{minipage}    
        }
        \subfigure{
        \begin{minipage}{0.085\linewidth}
          \centering{\footnotesize HaLRTC}
        \includegraphics[height = 1.5cm,width=1.7cm]{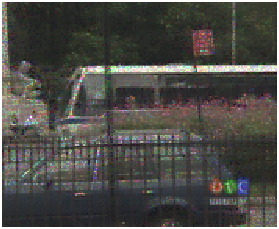}
        \end{minipage}    
        }
        \subfigure{
        \begin{minipage}{0.085\linewidth}
          \centering{\footnotesize TRLRF}
        \includegraphics[height = 1.5cm,width=1.7cm]{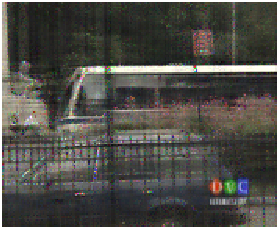}
        \end{minipage}    
        }
        \subfigure{
        \begin{minipage}{0.085\linewidth}
          \centering{\footnotesize FCTN-PAM}
        \includegraphics[height = 1.5cm,width=1.7cm]{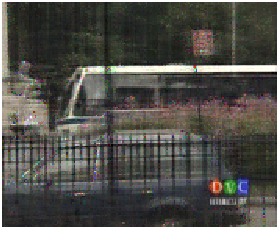}
        \end{minipage}    
        }
        \subfigure{
        \begin{minipage}{0.085\linewidth}
          \centering{\footnotesize LMTN-SVD}
        \includegraphics[height = 1.5cm,width=1.7cm]{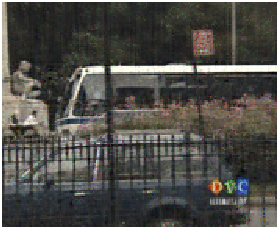}
        \end{minipage}    
        }
        \subfigure{
        \begin{minipage}{0.085\linewidth}
          \centering{\footnotesize LMTN-PAM}
        \includegraphics[height = 1.5cm,width=1.7cm]{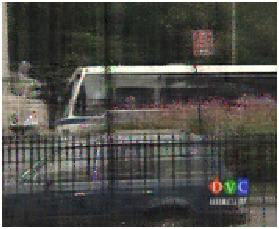}
        \end{minipage}    
        }
        \subfigure{
        \begin{minipage}{0.085\linewidth}
          \centering{\footnotesize LMTN-AR}
        \includegraphics[height = 1.5cm,width=1.7cm]{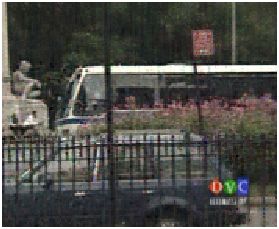}
        \end{minipage}    
        }
        \subfigure{
        \begin{minipage}{0.085\linewidth}
          \centering{\footnotesize Ground truth}
        \includegraphics[height = 1.5cm,width=1.7cm]{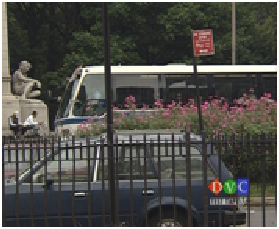}
        \end{minipage}    
        }

        \subfigure{
        \begin{minipage}{0.085\linewidth}
        \includegraphics[height = 1.5cm,width=1.7cm]{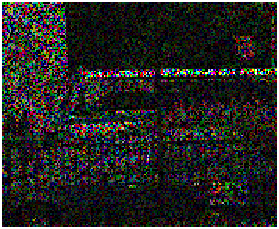}
        \end{minipage}
        }
        \subfigure{
        \begin{minipage}{0.085\linewidth}
        \includegraphics[height = 1.5cm,width=1.7cm]{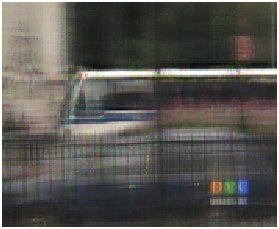}
        \end{minipage}    
        }
        \subfigure{
        \begin{minipage}{0.085\linewidth}
        \includegraphics[height = 1.5cm,width=1.7cm]{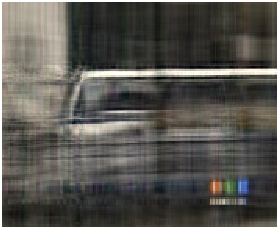}
        \end{minipage}    
        }
        \subfigure{
        \begin{minipage}{0.085\linewidth}
        \includegraphics[height = 1.5cm,width=1.7cm]{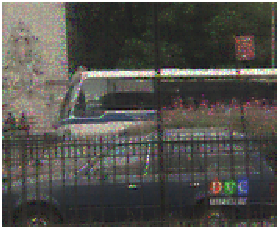}
        \end{minipage}    
        }
        \subfigure{
        \begin{minipage}{0.085\linewidth}
        \includegraphics[height = 1.5cm,width=1.7cm]{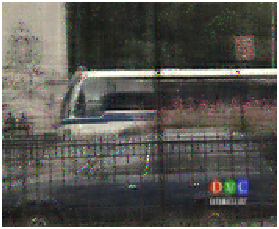}
        \end{minipage}    
        }
        \subfigure{
        \begin{minipage}{0.085\linewidth}
        \includegraphics[height = 1.5cm,width=1.7cm]{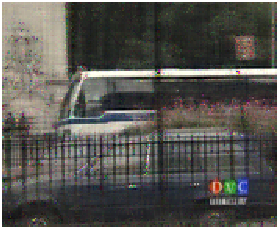}
        \end{minipage}    
        }
        \subfigure{
        \begin{minipage}{0.085\linewidth}
        \includegraphics[height = 1.5cm,width=1.7cm]{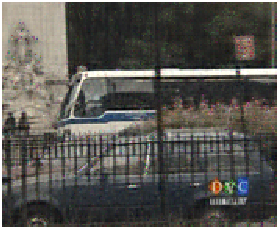}
        \end{minipage}    
        }
        \subfigure{
        \begin{minipage}{0.085\linewidth}
        \includegraphics[height = 1.5cm,width=1.7cm]{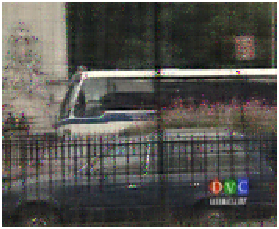}
        \end{minipage}    
        }
        \subfigure{
        \begin{minipage}{0.085\linewidth}
        \includegraphics[height = 1.5cm,width=1.7cm]{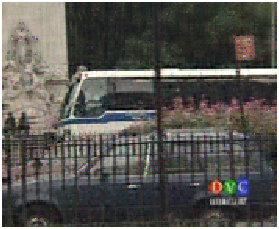}
        \end{minipage}    
        }
        \subfigure{
        \begin{minipage}{0.085\linewidth}
        \includegraphics[height = 1.5cm,width=1.7cm]{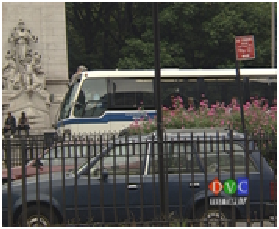}
        \end{minipage}    
        }

        \subfigure{
        \begin{minipage}{0.085\linewidth}
        \includegraphics[height = 1.5cm,width=1.7cm]{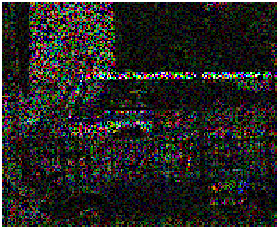}
        \end{minipage}
        }
        \subfigure{
        \begin{minipage}{0.085\linewidth}
        \includegraphics[height = 1.5cm,width=1.7cm]{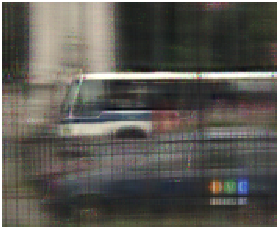}
        \end{minipage}    
        }
        \subfigure{
        \begin{minipage}{0.085\linewidth}
        \includegraphics[height = 1.5cm,width=1.7cm]{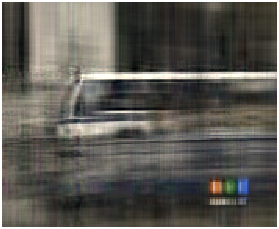}
        \end{minipage}    
        }
        \subfigure{
        \begin{minipage}{0.085\linewidth}
        \includegraphics[height = 1.5cm,width=1.7cm]{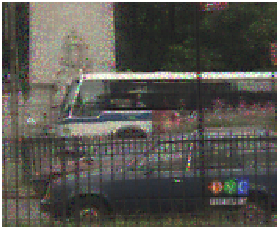}
        \end{minipage}    
        }
        \subfigure{
        \begin{minipage}{0.085\linewidth}
        \includegraphics[height = 1.5cm,width=1.7cm]{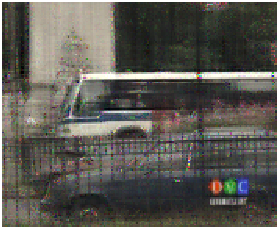}
        \end{minipage}    
        }
        \subfigure{
        \begin{minipage}{0.085\linewidth}
        \includegraphics[height = 1.5cm,width=1.7cm]{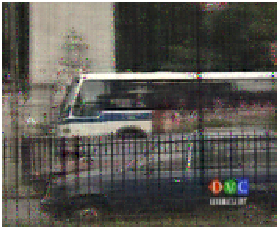}
        \end{minipage}    
        }
        \subfigure{
        \begin{minipage}{0.085\linewidth}
        \includegraphics[height = 1.5cm,width=1.7cm]{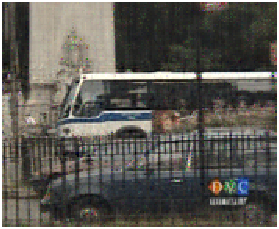}
        \end{minipage}    
        }
        \subfigure{
        \begin{minipage}{0.085\linewidth}
        \includegraphics[height = 1.5cm,width=1.7cm]{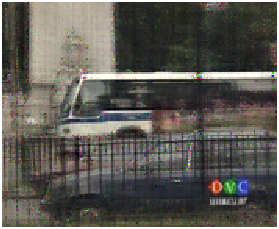}
        \end{minipage}    
        }
        \subfigure{
        \begin{minipage}{0.085\linewidth}
        \includegraphics[height = 1.5cm,width=1.7cm]{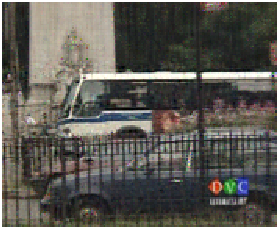}
        \end{minipage}    
        }
        \subfigure{
        \begin{minipage}{0.085\linewidth}
        \includegraphics[height = 1.5cm,width=1.7cm]{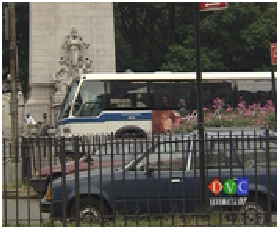}
        \end{minipage}    
        }

        \subfigure{
        \begin{minipage}{0.085\linewidth}
        \includegraphics[height = 1.5cm,width=1.7cm]{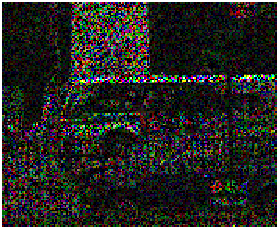}
        \end{minipage}
        }
        \subfigure{
        \begin{minipage}{0.085\linewidth}
        \includegraphics[height = 1.5cm,width=1.7cm]{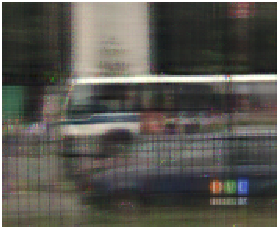}
        \end{minipage}    
        }
        \subfigure{
        \begin{minipage}{0.085\linewidth}
        \includegraphics[height = 1.5cm,width=1.7cm]{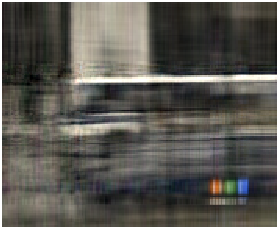}
        \end{minipage}    
        }
        \subfigure{
        \begin{minipage}{0.085\linewidth}
        \includegraphics[height = 1.5cm,width=1.7cm]{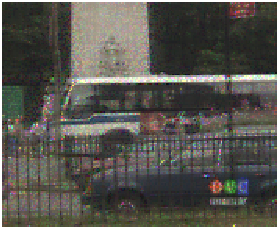}
        \end{minipage}    
        }
        \subfigure{
        \begin{minipage}{0.085\linewidth}
        \includegraphics[height = 1.5cm,width=1.7cm]{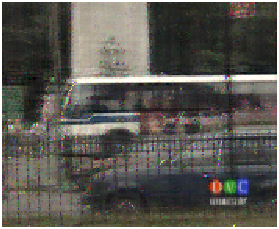}
        \end{minipage}    
        }
        \subfigure{
        \begin{minipage}{0.085\linewidth}
        \includegraphics[height = 1.5cm,width=1.7cm]{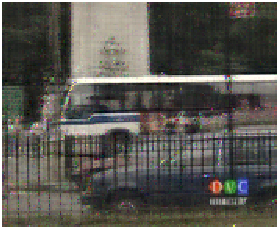}
        \end{minipage}    
        }
        \subfigure{
        \begin{minipage}{0.085\linewidth}
        \includegraphics[height = 1.5cm,width=1.7cm]{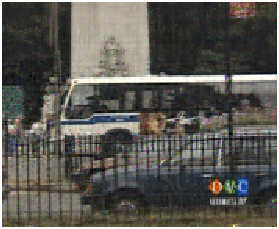}
        \end{minipage}    
        }
        \subfigure{
        \begin{minipage}{0.085\linewidth}
        \includegraphics[height = 1.5cm,width=1.7cm]{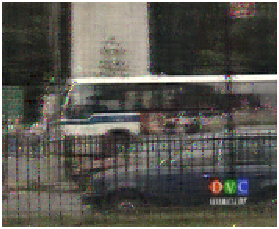}
        \end{minipage}    
        }
        \subfigure{
        \begin{minipage}{0.085\linewidth}
        \includegraphics[height = 1.5cm,width=1.7cm]{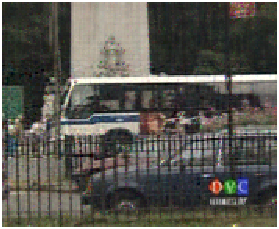}
        \end{minipage}    
        }
        \subfigure{
        \begin{minipage}{0.085\linewidth}
        \includegraphics[height = 1.5cm,width=1.7cm]{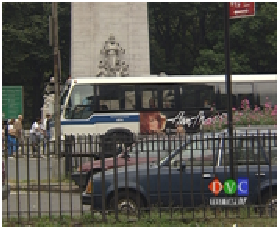}
        \end{minipage}    
        }
        
      \caption{The tested video bus size is 144$\times$176$\times$3$\times$50 and the figure shows a comparison of the results of the different algorithms after video completion. 
      From top to bottom, frames 10, 20, 30 and 40 of the video are shown respectively.}
  \label{fig_8}
\end{figure*}

Observing from the experimental results, the videos recovered by TMac, TR-WOPT, HaLRTC and TRLRF algorithms are blurred, 
while FCTN-PAM and our proposed LMTN series of algorithms are clearer. However, we can clearly see in our bus video experiments that our algorithms 
recover better than FCTN-PAM in terms of detail (statues on stone pillars). This is because our proposed LMTN algorithm 
is able to find the latent information inside FCTN-PAM in different situations, and it is effective in accelerating the algorithm operation. 

\subsection{Traffic data Recovery}
This experiment uses the traffic flow dataset\footnote{https://data.mendeley.com/datasets/tf483zkcmb/3} provided by Grenoble Traffic Lab, which collects traffic flow data from 
46 different road segments over a period of 244 days, measures 15 seconds each hour. We chose 30 days of data for the 
experiment and fused them into a new dataset consisting of 30 days of hourly, minute-by-minute measurements for 46 road sections. 
The size of the dataset was 60$\times$24$\times$30$\times$46 (minutes $\times$ hours $\times$ days $\times$ number of road sections). Then we randomly lost 
some data for tensor completion experiments. The first set had a missing rate of 95\% and the second set had a missing 
rate of 80\%. The rank of the FCTN-PAM algorithm is simply set to 4 and the rank of the LMTN algorithm is $(\mathbf{R}_{i,j})_{i\neq j}$ = 4 and $(\mathbf{G}_{i,j})_{i=j}$ = 20. 

Fig. 9 shows the difference between the original and reconstructed datasets divided by 1380 (30$\times$46). 
It can be observed that both our algorithm and the FCTN-PAM algorithm have good recovery results, which confirms the 
effectiveness of our algorithm in tensor completion.

\begin{figure*}
  \centering
        \subfigure{
        \begin{minipage}{0.095\linewidth}
        \includegraphics[height = 1.5cm,width=1.7cm]{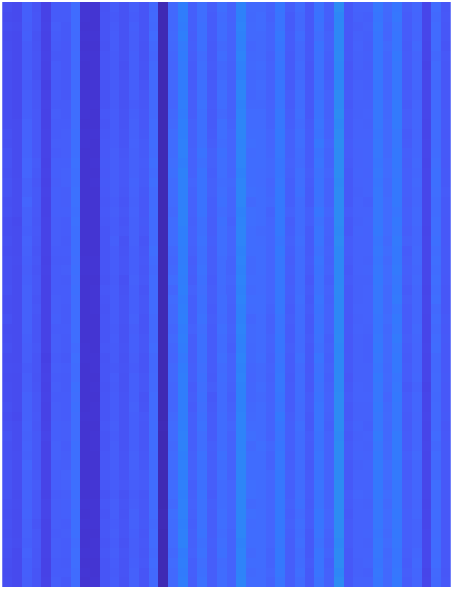}
        \end{minipage}
        }
        \subfigure{
        \begin{minipage}{0.095\linewidth}
        \includegraphics[height = 1.5cm,width=1.7cm]{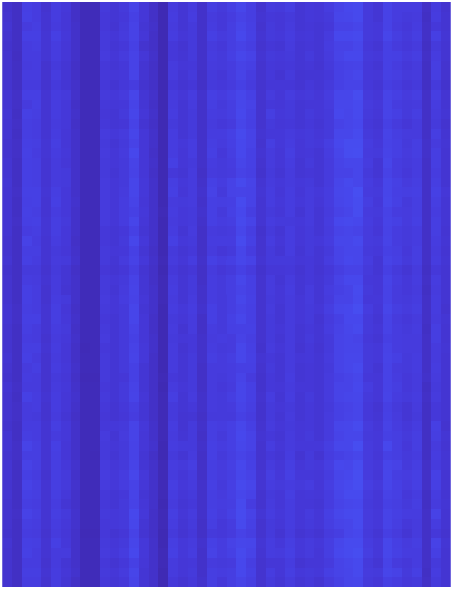}
        \end{minipage}    
        }
        \subfigure{
        \begin{minipage}{0.095\linewidth}
        \includegraphics[height = 1.5cm,width=1.7cm]{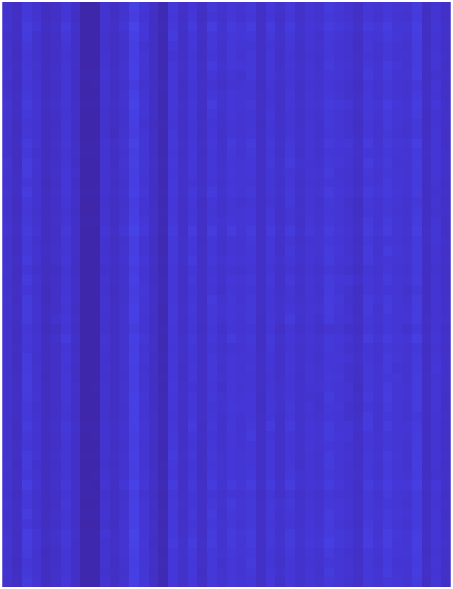}
        \end{minipage}    
        }
        \subfigure{
        \begin{minipage}{0.095\linewidth}
        \includegraphics[height = 1.5cm,width=1.7cm]{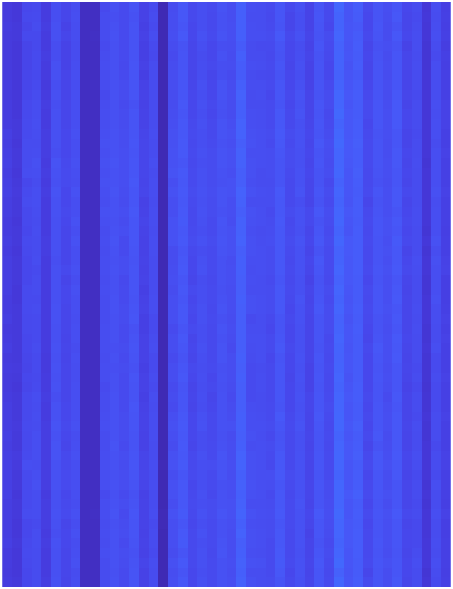}
        \end{minipage}    
        }
        \subfigure{
        \begin{minipage}{0.095\linewidth}
        \includegraphics[height = 1.5cm,width=1.7cm]{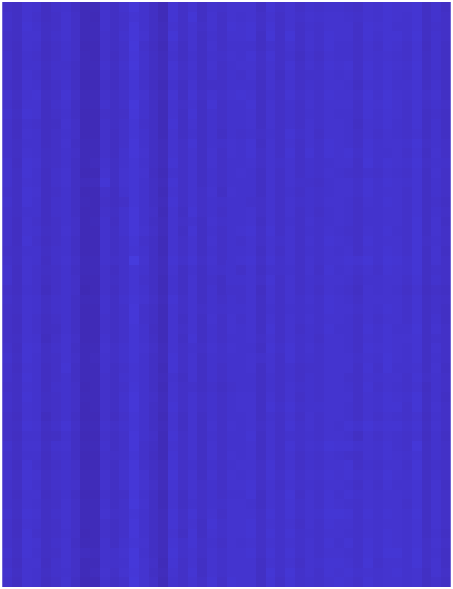}
        \end{minipage}    
        }
        \subfigure{
        \begin{minipage}{0.095\linewidth}
        \includegraphics[height = 1.5cm,width=1.7cm]{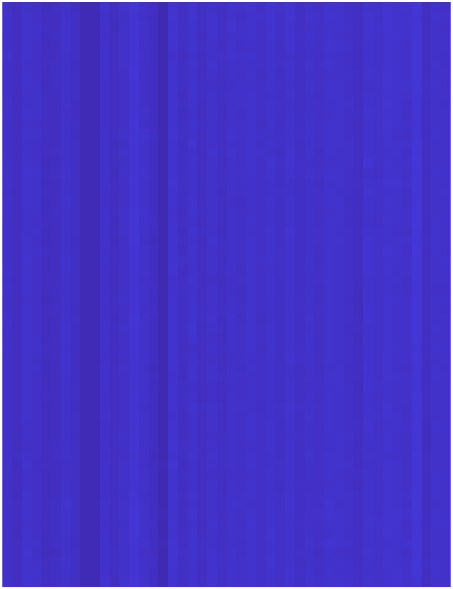}
        \end{minipage}    
        }
        \subfigure{
        \begin{minipage}{0.095\linewidth}
        \includegraphics[height = 1.5cm,width=1.7cm]{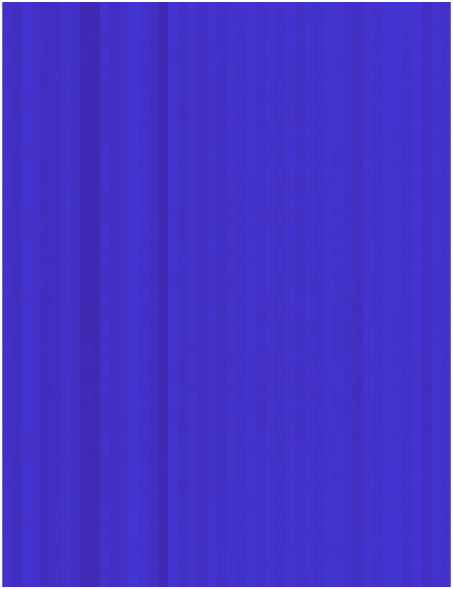}
        \end{minipage}    
        }
        \subfigure{
        \begin{minipage}{0.095\linewidth}
        \includegraphics[height = 1.5cm,width=1.7cm]{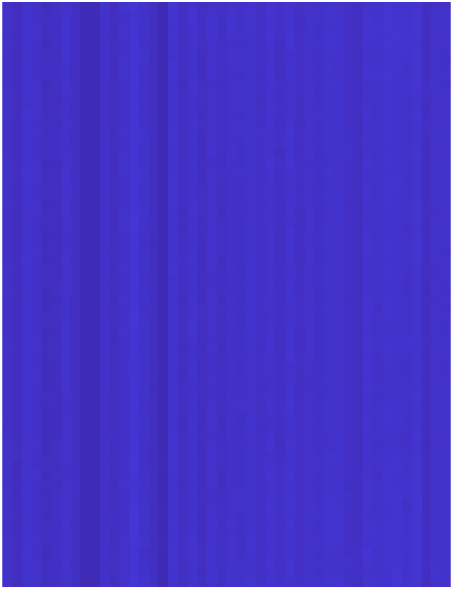}
        \end{minipage}    
        }
        \subfigure{
        \begin{minipage}{0.095\linewidth}
        \includegraphics[height = 1.5cm,width=1.7cm]{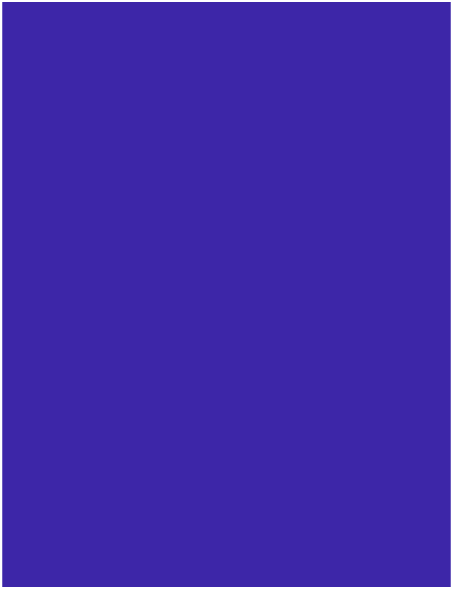}
        \end{minipage}    
        }

        \subfigure{
        \begin{minipage}{0.095\linewidth}
        \includegraphics[height = 1.5cm,width=1.7cm]{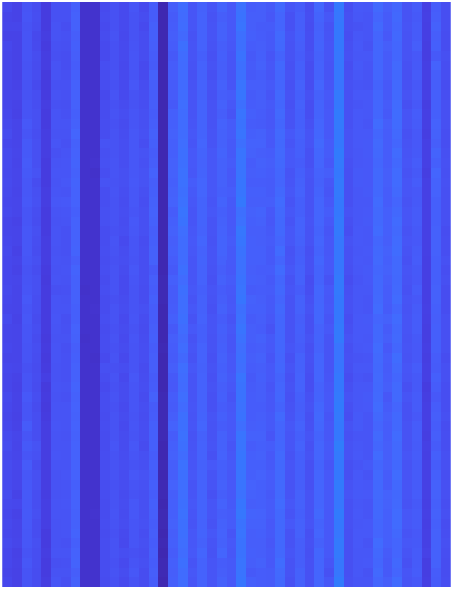}
        \end{minipage}
        }
        \subfigure{
        \begin{minipage}{0.095\linewidth}
        \includegraphics[height = 1.5cm,width=1.7cm]{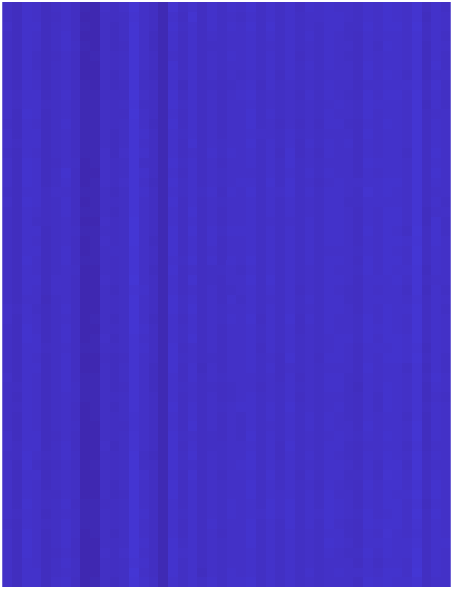}
        \end{minipage}    
        }
        \subfigure{
        \begin{minipage}{0.095\linewidth}
        \includegraphics[height = 1.5cm,width=1.7cm]{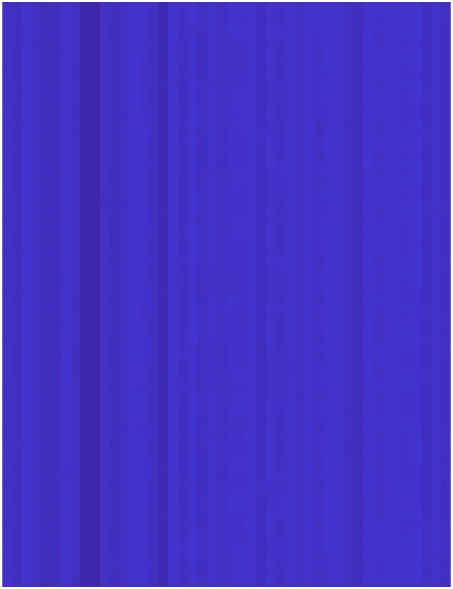}
        \end{minipage}    
        }
        \subfigure{
        \begin{minipage}{0.095\linewidth}
        \includegraphics[height = 1.5cm,width=1.7cm]{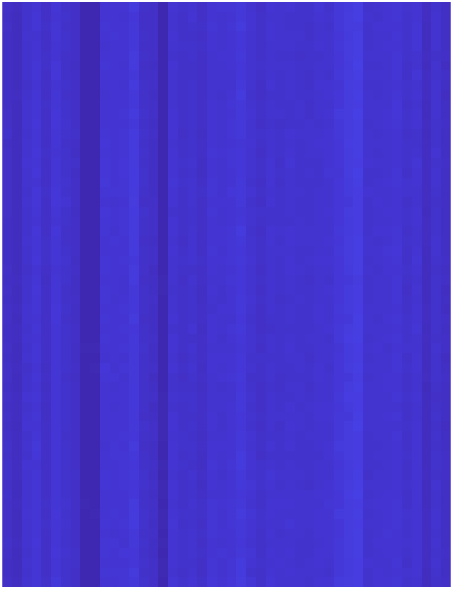}
        \end{minipage}    
        }
        \subfigure{
        \begin{minipage}{0.095\linewidth}
        \includegraphics[height = 1.5cm,width=1.7cm]{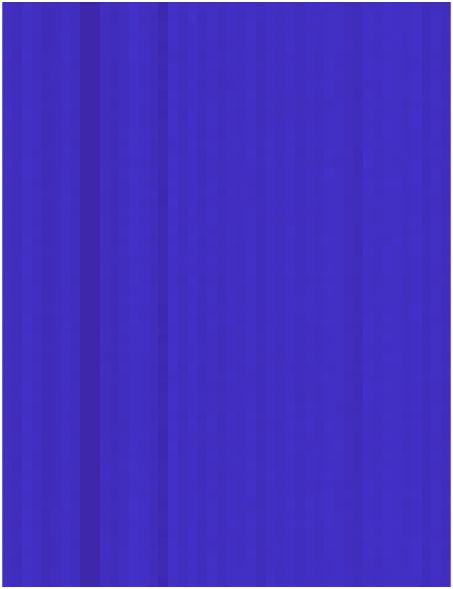}
        \end{minipage}    
        }
        \subfigure{
        \begin{minipage}{0.095\linewidth}
        \includegraphics[height = 1.5cm,width=1.7cm]{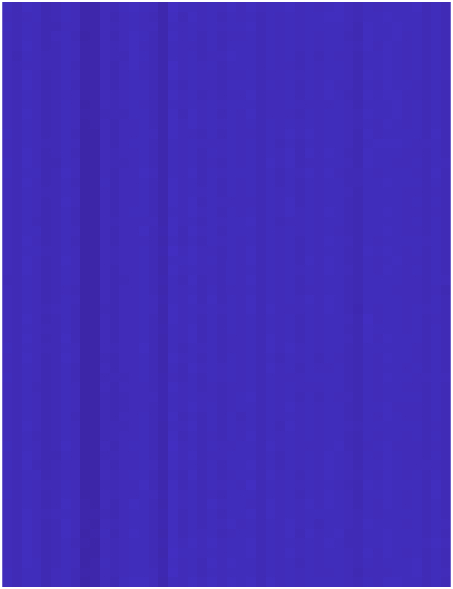}
        \end{minipage}    
        }
        \subfigure{
        \begin{minipage}{0.095\linewidth}
        \includegraphics[height = 1.5cm,width=1.7cm]{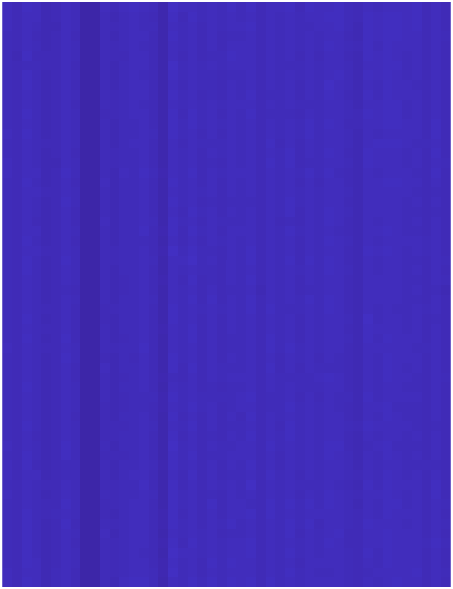}
        \end{minipage}    
        }
        \subfigure{
        \begin{minipage}{0.095\linewidth}
        \includegraphics[height = 1.5cm,width=1.7cm]{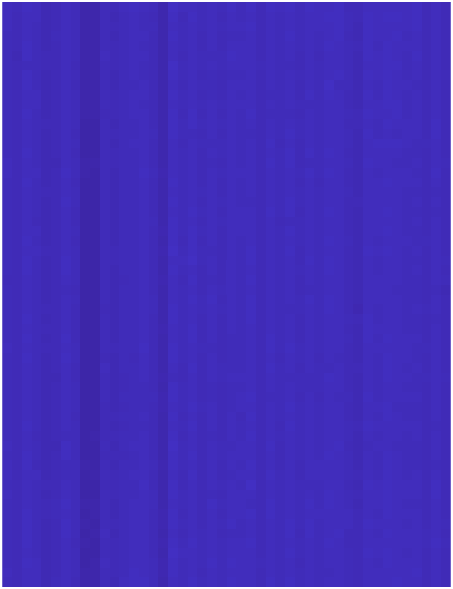}
        \end{minipage}    
        }
        \subfigure{
        \begin{minipage}{0.095\linewidth}
        \includegraphics[height = 1.5cm,width=1.7cm]{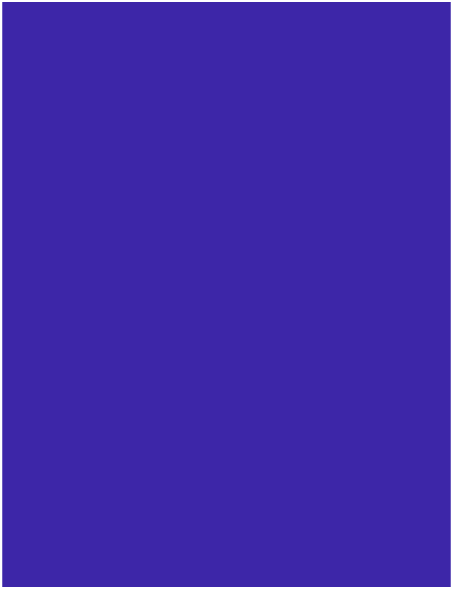}
        \end{minipage}    
        }

        \subfigure{
        \begin{minipage}{1\linewidth}
        \includegraphics[width=18cm]{rest_column2.eps}
        \end{minipage}    
        }
        
      \caption{The figure shows the average of the traffic flow residuals for all days, the vertical coordinates are indicated hour and 
      the horizontal coordinates are indicated minute, thus the size of each residual image is 24$\times$60. The first row shows the experimental 
      results of different algorithms for 95\% missing traffic flow data, and the second row shows the experimental results of different 
      algorithms for 80\% missing traffic flow data. }
  \label{fig_9}
\end{figure*}

\section{CONCLUSIONS}
In this paper, we consider a decomposition model of the potential matrix based on the FCTN model and apply it to the tensor completion task. 
We develop three different solution methods and they all have superior performance in tensor completion and data compression. 
In particular, LMTN-SVD is a non-recursive algorithm that is stable and efficient. LMTN-PAM requires us to give the rank of the tensor manually. 
LMTN-AR can find the rank of a tensor automatically, but sometimes it may encounter time-consuming situations. 
They are also shown that our model keeps the transpositional invariance of the FCTN model. 
To sum up, the overall performance of the LMTN algorithm is superior to the comparison methods.

\InterestConflict{The authors declare that they have no conflict of interest.}

%%%%%%%%%%%%%%%%%%%%%%%%%%%%%%%%%%%%%%%%%%%%%%%%%%%%%%%
%%% Supplements. ????????, ????
%%%%%%%%%%%%%%%%%%%%%%%%%%%%%%%%%%%%%%%%%%%%%%%%%%%%%%%
%\Supplements{}

%%%%%%%%%%%%%%%%%%%%%%%%%%%%%%%%%%%%%%%%%%%%%%%%%%%%%%%
%%% Reference section. ?¦Ï?????
%%% citation in the content using "some words~\cite{1,2}".
%%% ~ is needed to make the reference number is on the same line with the word before it.
%%%%%%%%%%%%%%%%%%%%%%%%%%%%%%%%%%%%%%%%%%%%%%%%%%%%%%%

%%%%%%%%%%%%%%%%%%%%%%%%%%%%%%%%%%%%%%%%%%%%%%%%%%%%%%%
%%% Appendix sections. ??????, ????
%%%%%%%%%%%%%%%%%%%%%%%%%%%%%%%%%%%%%%%%%%%%%%%%%%%%%%%
\begin{appendix}
\section{Proofs of Theorems}

\noindent $\mathbf{Theorem~1(Generalized~Relation)}$ There is a transformation relation for the $N$th-order tensor $\mathcal{X}$: 
\begin{equation*}
  \begin{aligned}
    &\mathcal{X}=FCTN(\{ \mathcal{G}_n\}_{n=1}^N) \times_1 \mathbf{M}_1 \times_2 \mathbf{M}_2 \cdots \times_N \mathbf{M}_N\\
    \Rightarrow &FCTN(\{ \mathcal{G}_n\}_{n=1}^N) = \mathcal{X} \times_1 \mathbf{M}_1^T \times_2 \mathbf{M}_2^T \cdots \times_N \mathbf{M}_N^T
  \end{aligned}
\end{equation*}
where $FCTN(\{ \mathcal{G}_n\}_{n=1}^N) \in \mathbb{R}^{R_{1,1}\times \cdots R_{N,N}}$ and $\mathbf{M}_n \in \mathbb{R}^{I_n \times R_{n,n}}$, and $R_{n,n} \leq I_n$. And its proof is provided in the Appendix A. 

\noindent $\mathit{Proof}$. Using the property $\mathcal{X} \times_n \mathbf{A} \times_n \mathbf{B}=\mathcal{X} \times (\mathbf{BA})$ of the mode product and noticing that $\mathbf{M}_n^T\mathbf{M}_n=\mathbf{I}$, we therefore have 
\begin{equation*}
  \begin{aligned}
    &\mathcal{X}\times_N \mathbf{M}_N^T  \\
    =&FCTN(\{ \mathcal{G}_n\}_{n=1}^N) \times_1 \mathbf{M}_1 \times_2 \mathbf{M}_2 \cdots \times_N \mathbf{M}_N \times_N \mathbf{M}_N^T\\
    =&FCTN(\{ \mathcal{G}_n\}_{n=1}^N) \times_1 \mathbf{M}_1 \times_2 \mathbf{M}_2 \cdots \times_N (\mathbf{M}_N^T \times \mathbf{M}_N)\\
    =&FCTN(\{ \mathcal{G}_n\}_{n=1}^N) \times_1 \mathbf{M}_1 \times_2 \mathbf{M}_2 \cdots \times_{N-1} \mathbf{M}_{N-1}
  \end{aligned}
\end{equation*}
Following this, we have $\mathcal{X}\times_N \mathbf{M}_N^T \times_{N-1} \mathbf{M}_{N-1}^T = FCTN(\{ \mathcal{G}_n\}_{n=1}^N) \times_1 \mathbf{M}_1 \times_2 \mathbf{M}_2 \cdots \times_{N-2} \mathbf{M}_{N-2}$. 
By extension, we can easily prove $FCTN(\{ \mathcal{G}_n\}_{n=1}^N) = \mathcal{X} \times_1 \mathbf{M}_1^T \times_2 \mathbf{M}_2^T \cdots \times_N \mathbf{M}_N^T$.

\noindent $\mathbf{Theorem~2(LMTN-PAM~convergence)}$ For the sequence \{$\mathcal{G}^{(s)}$, $\mathbf{M}^{(s)}$, $\mathcal{X}^{(s)}$\} global convergence to a critical point obtained by LMTN-PAM. 

 $\mathit{Proof}$. In order to prove Theorem 2, we refer to the literature \cite{43} and \cite{44} and only need to prove the following conditions.

 (a) If $\mathcal{G}_k^{(0)}(k=1,2,\cdots,N)$, $\mathbf{M}_k^{(0)}$ and $\mathcal{X}_k^{(0)}$ are bounded, \{$\mathcal{G}^{(s)}$\}, \{$\mathbf{M}^{(s)}$\} and \{$\mathcal{X}^{(s)}$\} are the bounded sequences;
 
 (b) $f(\mathcal{G},\mathbf{M}, \mathcal{X})$ is a proper lower semi-continuous function;
 
 (c) $f(\mathcal{G},\mathbf{M}, \mathcal{X})$ satisfies the Kurdyka–Łojasiewicz property\cite{45} at \{ $\mathcal{G}^{(s)}$, $\mathbf{M}^{(s)}$, $\mathcal{X}^{(s)}$\};
 
 (d) \{$\mathcal{G}^{(s)}$, $\mathbf{M}^{(s)}$, $\mathcal{X}^{(s)}$\} satisfied Lemmas 1 and 2.
 
 \noindent $Lemma~1$ Assuming that the sequences \{$\mathcal{G}^{(s)}$, $\mathbf{M}^{(s)}$, $\mathcal{X}^{(s)}$\} are all obtained by LMTN-PAM, then we have
 \begin{gather*}
 f(\mathcal{G}_{1:k-1}^{(s+1)}, \mathcal{G}_{k:N}^{(s)}, \mathbf{M}_{1:k}^{(s+1)}, \mathbf{M}_{k+1:N}^{(s)}, \mathcal{X}^{(s)}) \\ +  \frac{\rho}{2} \Vert \mathbf{M}_k^{(s+1)} - \mathbf{M}_k^{(s)} \Vert _F^2 \\
 \leq f(\mathcal{G}_{1:k-1}^{(s+1)}, \mathcal{G}_{k:N}^{(s)}, \mathbf{M}_{1:k-1}^{(s+1)}, \mathbf{M}_{k:N}^{(s)}, \mathcal{X}^{(s)}), k = 1,2,\cdots,N; \\
 f(\mathcal{G}_{1:k}^{(s+1)}, \mathcal{G}_{k+1:N}^{(s)}, \mathbf{M}_{1:k}^{(s+1)}, \mathbf{M}_{k+1:N}^{(s)}, \mathcal{X}^{(s)}) \\ + \frac{\rho}{2} \Vert \mathcal{G}_k^{(s+1)} - \mathcal{G}_k^{(s)} \Vert _F^2 \\
 \leq f(\mathcal{G}_{1:k-1}^{(s+1)}, \mathcal{G}_{k:N}^{(s)}, \mathbf{M}_{1:k}^{(s+1)}, \mathbf{M}_{k+1:N}^{(s)}, \mathcal{X}^{(s)}); \\
 f(\mathcal{G}^{(s+1)}, \mathbf{M}^{(s+1)}, \mathcal{X}^{(s+1)}) + \frac{\rho}{2} \Vert \mathcal{X}_k^{(s+1)} - \mathbf{X}_k^{(s)} \Vert _F^2 \\
 \leq f(\mathcal{G}^{(s+1)}, \mathbf{M}^{(s+1)}, \mathcal{X}^{(s)}).
 \end{gather*} 
 
 \noindent $Lemma~2$ Assuming that the sequences \{$\mathcal{G}^{(s)}$, $\mathbf{M}^{(s)}$, $\mathcal{X}^{(s)}$\} are all obtained by LMTN-PAM, then there exists $\mathcal{A}_k^{(s+1)} \in 0$, $\mathcal{B}_k^{(s+1)}$ and $\mathcal{C}^{(s+1)} \in \partial_{\mathcal{X}}l_{\mathbb{S}}(\mathcal{X}^{(s+1)})$ satisfies
 \begin{gather*}
 \Vert \mathcal{A}_k^{(s+1)} + \nabla_{\mathbf{M}_k}h(\mathcal{G}_{1:k-1}^{(s+1)}, \mathcal{G}_{k:N}^{(s)}, \mathbf{M}_{1:k}^{(s+1)}, \mathbf{M}_{k+1:N}^{(s)}), \mathcal{X}^{(s)} \Vert_F\\
 \leq \rho \Vert \mathbf{M}_k^{(s+1)} - \mathbf{M}_K^{(s)} \Vert_F;\\
 \Vert \mathcal{B}_k^{(s+1)} + \nabla_{\mathcal{G}_k}h(\mathcal{G}_{1:k}^{(s+1)}, \mathcal{G}_{k+1:N}^{(s)}, \mathbf{M}_{1:k}^{(s+1)}, \mathbf{M}_{k+1:N}^{(s)}), \mathcal{X}^{(s)} \Vert_F\\
 \leq \rho \Vert \mathcal{G}_k^{(s+1)} - \mathcal{G}_K^{(s)} \Vert_F;\\
 \Vert \mathcal{C}^{(s+1)} + \nabla_{\mathcal{X}_k}h(\mathcal{G}^{(s+1)}, M^{(s+1)}, \mathcal{X}^{(s+1)}) \Vert_F \\ \leq \rho \Vert \mathcal{X}_k^{(s+1)} - \mathcal{X}_K^{(s)} \Vert_F.
 \end{gather*} 
 where $h(\mathcal{G},\mathbf{M},\mathcal{X}) = \frac{1}{2}\Vert \mathcal{X}-LMTN(\{ \mathcal{G}_n\}_{n=1}^N,\{ \mathbf{M}_n\}_{n=1}^N) \Vert_F^2$ and $f(\mathcal{G},\mathbf{M},\mathcal{X}) = \frac{1}{2}\Vert \mathcal{X}-LMTN(\{ \mathcal{G}_n\}_{n=1}^N,\{ \mathbf{M}_n\}_{n=1}^N) \Vert_F^2 + l_{\mathbb{S}}(\mathcal{X})$
 
 First, initialize $\mathcal{G}_k^{(0)}(k=1,2,\cdots,N)$, $\mathbf{M}_k^{(0)}$ and $\mathcal{X}_k^{(0)}$ in a range of values that are bounded. Thus we only need to prove that $\mathcal{G}_k^{(s+1)}(k=1,2,\cdots,N)$, $\mathbf{M}_k^{(s+1)}$ and $\mathcal{X}_k^{(s+1)}$ are bounded when $\mathcal{G}_k^{(s)}(k=1,2,\cdots,N)$, $\mathbf{M}_k^{(s)}$ and $\mathcal{X}_k^{(s)}$ are bounded. 
 Suppose $\Vert \mathcal{G}_k^{(s)}\Vert_F\leq d $, $\Vert \mathbf{M}_k^{(s)}\Vert_F\leq c $, $\Vert \mathcal{X}^{(s)} \Vert_F\leq e$, according to equation (9) we have
 $$
 \begin{aligned}
 \Vert \mathbf{M}_{1}^{(s+1)} \Vert_F \leq &(\rho \Vert \mathbf{M}_{1}^{(s)} \Vert_F + \Vert \mathbf{X}_{(1)}^{(s)} \mathbf{Y}_{(\neq 1)}^{(s)} (\mathbf{G}_{1})_{(1)}^{(s)T} \Vert_F)\\
 &\Vert (\rho \mathbf{I} + \mathbf{Q}_1^{(s)} \mathbf{Q}_1^{(s)T})^{-1} \Vert_F\\
 \leq &[\rho c+ed^{n}c^{n-1}] \sqrt{\sum_{i=1}^{j}\frac{1}{(\rho+\theta_i)^2}}\\
 \leq &[\rho c+ed^{n}c^{n-1}] \frac{\sqrt{j}}{\rho}
 \end{aligned}
 $$
 where $\theta_i(i\in N)$ are denote to the eigenvalues of $Q_1^{(s)}$, and $Q_1^{(s)} = (\mathbf{G}_1)_{(1)}^{(s)}\mathbf{Y}_{(\neq 1)}^{(s)}$. 
 Therefore $\mathbf{M}_1^{(s+1)}$ is bounded, and in the same way, it can be deduced that $\mathbf{M}_2^{(s+1)},\mathbf{M}_3^{(s+1)},\cdot,\mathbf{M}_N^{(s+1)}$ are also bounded. 
 And supposing $\Vert \mathbf{M}_k^{(s+1)} \Vert_F \leq c_2$, we have 
 $$
 \begin{aligned}
   &\Vert \mathcal{G}_1^{(s+1)} \Vert_F\\
   \leq& [\rho \sum_{i=1}^{N} \mathbf{R}_{1,i}\Vert (\mathbf{M}_1^{(s+1)T}\mathbf{M}_1^{(s+1)})^{-1} \Vert_F + \mathbf{R}_{1,1}\Vert \mathbf{Y}_{(\neq 1)}^{(s)} \mathbf{Y}_{(\neq 1)}^{(s)T}\Vert_F]\\
   &\Vert (\mathbf{M}_1^{(s+1)T} \mathbf{M}_1^{(s+1)})^{-1}[\rho(\mathbf{G}_1)_{(1)}^{(s)} + \mathbf{M}_1^{(s+1)T} \mathbf{X}_{(1)}^{(s)T} \mathbf{Y}_{(\neq 1)}^{(s)T} ]\Vert_F \\
   \leq& [\rho \sum_{i=1,i\neq 1}^{N}\mathbf{R}_{1,i}\sqrt{\sum_{i=1}^{j}\frac{1}{\delta_i^2}} + \mathbf{R}_{1,1}ec_2^2d^{2(n-1)}c^{2(n-2)}] \\
   &[( \sum_{i=1,i\neq 1}^{N}\mathbf{R}_{1,i}\sqrt{\sum_{i=1}^{j}\frac{1}{\delta_i^2}})(\rho d+ec_2d^{n-1}c^{n-2})] \\
 \end{aligned}  
 $$
 where $\delta_i(i\in N)$ are denote to the eigenvalues of $M_1^{(s+1)T}M_1^{(s+1)}$, and $j$ denote the number of eigenvalues of $M_1^{(s+1)T}M_1^{(s+1)}$. 
 Thus, we proof the $\mathcal{G}_1^{(s+1)}$ is bounded. And it can be deduced that $\mathcal{G}_2^{(s+1)}, \mathcal{G}_3^{(s+1)},\cdots,\mathcal{G}_N^{(s+1)}$ are bounded. 
 Supposing that $\Vert \mathbf{M}_k^{(s+1)} \Vert_F\leq d_2$, we have
 $$
 \begin{aligned}
 \Vert \mathcal{X}^{(s+1)} \Vert_F \leq \frac{d_2^Nc_2^N+\rho e}{1+\rho} + \Vert \mathcal{T} \Vert_F
 \end{aligned}  
 $$
 Therefore, the proof of condition (a) is complete.
 
 Second, function $f(\mathcal{G},\mathbf{M},\mathcal{X})$ is composed of $h(\mathcal{G},\mathbf{M},\mathcal{X})$ and $l_{\mathbb{S}}(\mathcal{X})$. 
 Furthermore, $h(\mathcal{G},\mathbf{M},\mathcal{X})$ is a $C_1$ function whose gradient is Lipschitz continuous, and $l_{\mathbb{S}}(\mathcal{X})$ is 
 a proper lower semi-continuous function. Hence $f(\mathcal{G},\mathbf{M},\mathcal{X})$ is also a proper lower semi-continuous function. 
 
 Third, since $h(\mathcal{G},\mathbf{M},\mathcal{X})$ and $l_{\mathbb{S}}(\mathcal{X})$ are both semi-algebraic functions, $f(\mathcal{G},\mathbf{M},\mathcal{X})$ is also a semi-algebraic function. 
 And since the semi-algebraic functions satisfy Kurdyka–Łojasiewicz property, the proof of condition (c) is complete.
 
 Fourth, we will prove the lemma 1 and lemma 2. For the lemma 1, since the $\mathbf{M}_k^{(s+1)}$ is the optional solution of $\mathbf{M}_k^{(s)}$-subproblem, 
 we obtain that 
 
 \begin{gather*}
   f(\mathcal{G}_{1:k-1}^{(s+1)}, \mathcal{G}_{k:N}^{(s)}, \mathbf{M}_{1:k}^{(s+1)}, \mathbf{M}_{k+1:N}^{(s)}, \mathcal{X}^{(s)}) \\  + \frac{\rho}{2} \Vert \mathbf{M}_k^{(s+1)}- \mathbf{M}_k^{(s)} \Vert _F^2 \\
   \leq f(\mathcal{G}_{1:k-1}^{(s+1)}, \mathcal{G}_{k:N}^{(s)}, \mathbf{M}_{1:k-1}^{(s+1)}, \mathbf{M}_{k:N}^{(s)}, \mathcal{X}^{(s)}) \\  + \frac{\rho}{2} \Vert \mathbf{M}_k^{(s)} - \mathbf{M}_k^{(s)} \Vert _F^2 \\
   \leq f(\mathcal{G}_{1:k-1}^{(s+1)}, \mathcal{G}_{k:N}^{(s)}, \mathbf{M}_{1:k-1}^{(s+1)}, \mathbf{M}_{k:N}^{(s)}, \mathcal{X}^{(s)})
 \end{gather*}

 And $\mathcal{G}_k^{(s+1)}$ is the optional solution of $\mathcal{G}_k^{(s)}$-subproblem, we have 
 
 \begin{gather*}
   f(\mathcal{G}_{1:k}^{(s+1)}, \mathcal{G}_{k+1:N}^{(s)}, \mathbf{M}_{1:k}^{(s+1)}, \mathbf{M}_{k+1:N}^{(s)}, \mathcal{X}^{(s)}) \\  + \frac{\rho}{2} \Vert \mathcal{G}_k^{(s+1)} - \mathcal{G}_k^{(s)} \Vert _F^2 \\
   \leq f(\mathcal{G}_{1:k-1}^{(s+1)}, \mathcal{G}_{k:N}^{(s)}, \mathbf{M}_{1:k}^{(s+1)}, \mathbf{M}_{k+1:N}^{(s)}, \mathcal{X}^{(s)}) \\  + \frac{\rho}{2} \Vert \mathcal{G}_k^{(s)} - \mathcal{G}_k^{(s)} \Vert _F^2 \\
   \leq f(\mathcal{G}_{1:k-1}^{(s+1)}, \mathcal{G}_{k:N}^{(s)}, \mathbf{M}_{1:k}^{(s+1)}, \mathbf{M}_{k+1:N}^{(s)}, \mathcal{X}^{(s)}); 
 \end{gather*}

 Similarly, the $\mathcal{X}^{(s+1)}$ is the optional solution problem of $\mathcal{X}^{(s)}$-subproblem, we also have
 $$
 \begin{aligned}
   &f(\mathcal{G}^{(s+1)}, \mathbf{M}^{(s+1)}, \mathcal{X}^{(s+1)}) + \frac{\rho}{2} \Vert \mathcal{X}_k^{(s+1)} - \mathbf{X}_k^{(s)} \Vert _F^2 \\
   &\leq f(\mathcal{G}^{(s+1)}, \mathbf{M}^{(s+1)}, \mathcal{X}^{(s)})+ \frac{\rho}{2} \Vert \mathcal{X}_k^{(s)} - \mathbf{X}_k^{(s)} \Vert _F^2 \\
   &\leq f(\mathcal{G}^{(s+1)}, \mathbf{M}^{(s+1)}, \mathcal{X}^{(s)})\\
 \end{aligned}
 $$
 Therefore, we proof lemma 1 successfully. In the lemma 2, for each subproblem, we have 
 $$
 \begin{aligned}
   0 \in &\nabla_{\mathbf{M}_k}h(\mathcal{G}_{1:k-1}^{(s+1)}, \mathcal{G}_{k:N}^{(s)}, \mathbf{M}_{1:k}^{(s+1)}, \mathbf{M}_{k+1:N}^{(s)}) + \rho (\mathbf{M}_k - \mathbf{M}_k^{(s)}) \\
   0 \in &\nabla_{\mathcal{G}_k}h(\mathcal{G}_{1:k}^{(s+1)}, \mathcal{G}_{k+1:N}^{(s)}, \mathbf{M}_{1:k}^{(s+1)}, \mathbf{M}_{k+1:N}^{(s)}) + \rho (\mathcal{G}_k - \mathcal{G}_k^{(s)}) \\
   0 \in &\nabla_{\mathcal{X}}h(\mathcal{G}^{(s+1)}, M^{(s+1)}, \mathcal{X}^{(s+1)}) + \rho (\mathcal{X}_k - \mathcal{X}_k^{(s)}) \\ & + \partial_\mathcal{X} l_{\mathbb{S}}(\mathcal{X}^{(s+1)}) 
 \end{aligned} 
 $$
 Assuming that 
 $$
 \begin{aligned}
   \mathcal{A}_k^{(s+1)} = &-\nabla_{\mathbf{M}_k}h(\mathcal{G}_{1:k-1}^{(s+1)}, \mathcal{G}_{k:N}^{(s)}, \mathbf{M}_{1:k}^{(s+1)}, \mathbf{M}_{k+1:N}^{(s)})\\ & - \rho (\mathbf{M}_k - \mathbf{M}_k^{(s)}) \\
   \mathcal{B}_k^{(s+1)} =  &-\nabla_{\mathcal{G}_k}h(\mathcal{G}_{1:k}^{(s+1)}, \mathcal{G}_{k+1:N}^{(s)}, \mathbf{M}_{1:k}^{(s+1)}, \mathbf{M}_{k+1:N}^{(s)})\\ & - \rho (\mathcal{G}_k - \mathcal{G}_k^{(s)}) \\
   \mathcal{C}^{(s+1)} = &-\nabla_{\mathcal{X}}h(\mathcal{G}^{(s+1)}, M^{(s+1)}, \mathcal{X}^{(s+1)}) - \rho (\mathcal{X}_k - \mathcal{X}_k^{(s)})\\
   & - \partial_\mathcal{X} l_{\mathbb{S}}(\mathcal{X}^{(s+1)}) 
 \end{aligned} 
 $$
 Therefore, the condition (d) holds. And we proved LMTN-PAM can globally converge to a critical minima successfully. 
 
 \noindent $\mathbf{Theorem~3(LMTN-SVD~convergence)}$ Let $\{ \mathbf{u}\}$ be a sequence of $\{ \mathcal{G}_k, \mathbf{M}_k, \mathcal{X} \}$ generated by LMTN-SVD algorithm. 
 For any $s$, if $\mathbf{u}^{(s)}$ is not a stationary point of $f(\mathbf{u}^{(s)})$, i.e., $\nabla f(\mathbf{u}^{(s)})\neq 0$, then $f(\mathbf{u}^{(s+1)}) \leq f(\mathbf{u}^{(s)})$.
 
 \noindent $\mathit{Proof}$. In the LMTN-SVD algorithm, the update of the latent matrix $\mathbf{M}$ is a non-iterative update, while $\mathcal{G}$ and $\mathcal{X}$ are updated using least squares. 
 Thus the essence of the LMTN-SVD optimization algorithm framework is the alternating least squares(ALS) framework. 
 Since $\nabla f(\mathbf{u}^{(s)})\neq 0$, there exists a set of indicates $\mathcal{I}$ such that $\nabla_{\mathbf{u}_i^{(s)}} f(\mathbf{u}^{(s)})\neq 0$ for $i\in \mathcal{I}$. 
 For all subsequent steps $s$-th to $N$-th, the objective value is not increased due to the nature of the least squares solution. 
 Because the $i$-th step solves a least squares problem with a non-zero gradient, the objective value is strictly reduced. 
 Therefore we deduce that $f(\mathbf{u}^{(s+1)}) \leq f(\mathbf{u}^{(s)})$ if $\mathbf{u}^{(s)}$ is not a stationary point.

% \renewcommand{\thesection}{Appendix}%²»¼ÓABC

% \section{}

\end{appendix}

\end{multicols}
% \bibliographystyle{IEEEtran}
% \bibliography{IEEEexample}
\end{document}